\documentclass[journal]{IEEEtai}
\usepackage[T1]{fontenc}
\usepackage{amsmath,amsfonts}
\usepackage{algorithmic}
\usepackage{algorithm}
\usepackage{array}
\usepackage{booktabs}
\usepackage{longtable}
\usepackage{multirow}
\usepackage{tabularx}
\usepackage[caption=false,font=normalsize,labelfont=sf,textfont=sf]{subfig}
\usepackage{textcomp}
\usepackage{stfloats}
\usepackage{url}
\usepackage{verbatim}
\usepackage{graphicx}
\usepackage{multibib}
\usepackage{capt-of}
\usepackage{cite}
\usepackage[table]{xcolor}
\usepackage[hidelinks]{hyperref}
\newcounter{promptbox}
\renewcommand{\thepromptbox}{S\arabic{promptbox}}
\newcommand{\promptboxlabelnum}[2]{\setcounter{promptbox}{#1}\addtocounter{promptbox}{-1}\refstepcounter{promptbox}\label{#2}}
\definecolor{bestcell}{HTML}{E6F2FF}
\definecolor{baselinepath}{HTML}{E6F2FF}

\usepackage{tcolorbox}
\tcbuselibrary{skins, breakable}

\newtcolorbox{promptbox}[1][]{
  breakable,
  enhanced,
  colback=gray!3!white,
  colframe=black!55,
  coltitle=black,
  fonttitle=\bfseries\footnotesize,
  title=#1,
  boxrule=0.55pt,
  arc=2pt,
  leftrule=0.9pt,
  rightrule=0.4pt,
  toprule=0.4pt,
  bottomrule=0.4pt,
  left=7pt, right=7pt, top=6pt, bottom=6pt,
  before skip=6pt,
  after skip=6pt,
  fontupper=\footnotesize\sffamily,
  colbacktitle=gray!10!white,
  boxed title style={boxrule=0pt, arc=1pt, left=4pt, right=4pt, top=3pt, bottom=3pt},
  before upper={\setlength{\parskip}{2pt}\setlength{\parindent}{1em}},
  overlay unbroken and first={\draw[black!18,line width=0.4pt] ([xshift=4pt,yshift=-4pt]frame.north west) -- ([xshift=-4pt,yshift=-4pt]frame.north east);}
}

\usepackage{listings}
\usepackage{xcolor}

\definecolor{codegreen}{rgb}{0,0.6,0}
\definecolor{codegray}{rgb}{0.5,0.5,0.5}
\definecolor{codepurple}{rgb}{0.58,0,0.82}
\definecolor{backcolour}{rgb}{0.97,0.97,0.97}

\lstdefinestyle{academicstyle}{
  backgroundcolor=\color{backcolour},
  commentstyle=\color{codegreen}\itshape,
  keywordstyle=\color{blue}\bfseries,
  numberstyle=\tiny\color{codegray},
  stringstyle=\color{codepurple},
  basicstyle=\ttfamily\scriptsize,
  breakatwhitespace=false,
  breaklines=true,
  captionpos=b,
  keepspaces=true,
  numbers=left,
  numbersep=7pt,
  xleftmargin=1.4em,
  framexleftmargin=1em,
  framesep=4pt,
  aboveskip=6pt,
  belowskip=6pt,
  showspaces=false,
  showstringspaces=false,
  showtabs=false,
  tabsize=4,
  columns=fullflexible,
  upquote=true,
  frame=single,
  framerule=0.4pt,
  rulecolor=\color{black!22}
}
\newcites{supp}{References}
\makeatletter
\newcommand{\supp@origaddcontentsline}{}
\newcommand{\startsupplementarytoc}{%
  \let\supp@origaddcontentsline\addcontentsline
  \renewcommand{\addcontentsline}[3]{\supp@origaddcontentsline{stc}{##2}{##3}}%
}
\newcommand{\printsupplementarytoc}{%
  \section*{\contentsname}
  \@starttoc{stc}%
}
\makeatother

\newcommand{\startsupplementarydocument}{%
  \clearpage
  \setcounter{page}{1}%
  \setcounter{section}{0}%
  \setcounter{subsection}{0}%
  \setcounter{subsubsection}{0}%
  \setcounter{table}{0}%
  \setcounter{figure}{0}%
  \setcounter{equation}{0}%
  \setcounter{algorithm}{0}%
  \setcounter{lstlisting}{0}%
  \setcounter{promptbox}{0}%
  \setcounter{tocdepth}{2}%
  \setcounter{secnumdepth}{3}%
  \renewcommand{\thetable}{S-\uppercase\expandafter{\romannumeral\arabic{table}}}%
  \renewcommand{\thesection}{S-\uppercase\expandafter{\romannumeral\arabic{section}}}%
  \renewcommand{\thesubsection}{\thesection.\arabic{subsection}}%
  \renewcommand{\thesubsubsection}{\thesubsection.\arabic{subsubsection}}%
  \renewcommand{\thefigure}{S\arabic{figure}}%
  \renewcommand{\theequation}{S\arabic{equation}}%
  \renewcommand{\thealgorithm}{S\arabic{algorithm}}%
  \renewcommand{\thelstlisting}{S\arabic{lstlisting}}%
  \renewcommand{\contentsname}{Supplementary Contents}%
  \startsupplementarytoc
  \twocolumn[%
    \begin{center}
      {\LARGE Supplementary Document of ``Co-evolving Agent Architectures and Interpretable Reasoning for Automated Optimization''\par}
      \vspace{0.75em}
    \end{center}
  ]
  \printsupplementarytoc
  \vspace{0.5em}
}

\begin{document}

\title{Co-evolving Agent Architectures and Interpretable Reasoning for Automated Optimization}


\author{Jiahao~Huang, Peilan~Xu,~\IEEEmembership{Member,~IEEE}, Xiaoya~Nan, and Wenjian~Luo,~\IEEEmembership{Senior Member,~IEEE}
  \thanks{Jiahao Huang, Peilan Xu, and Xiaoya Nan are with the School of Artificial Intelligence, Nanjing University of Information Science and Technology, Nanjing 210044, China (e-mail: 202383300169@nuist.edu.cn; xpl@nuist.edu.cn; xynan@nuist.edu.cn). (\textit{Corresponding author: Peilan Xu.})}
\thanks{Wenjian Luo is with Guangdong Provincial Key Laboratory of Novel Security Intelligence Technologies, Institute of Cyberspace Security, School of Computer Science and Technology, Harbin Institute of Technology, Shenzhen 518055, Guangdong, China (e-mail: luowenjian@hit.edu.cn).}}


\maketitle

\begin{abstract}
  Automating operations research (OR) with large language models (LLMs) remains limited by hand-crafted reasoning--execution workflows. Complex OR tasks require adaptive coordination among problem interpretation, mathematical formulation, solver selection, code generation, and iterative debugging. To address this limitation, we propose EvoOR-Agent, a co-evolutionary framework for automated optimization. EvoOR-Agent represents agent workflows as activity-on-edge (AOE)-style networks, making workflow topology, execution dependencies, and alternative reasoning paths explicit. On this representation, it maintains an architecture graph and evolves a population of reasoning individuals through graph-mediated path-conditioned recombination, multi-granularity semantic mutation, and elitist population update. A knowledge-base-assisted experience-acquisition module further injects reusable OR practices into initialization and semantic variation. Empirical results on seven heterogeneous OR benchmarks, including IndustryOR, MAMO, NL4OPT, BWOR, NLP4LP, and ReSocratic, show that EvoOR-Agent consistently improves over zero-shot LLMs, fixed-pipeline OR agents, specialized OR modeling methods, and representative evolutionary agent frameworks. Ablation studies further verify the contributions of the AOE-style architecture representation and knowledge-base-assisted operators. These results suggest that treating agent architectures and reasoning trajectories as evolvable objects provides an effective route toward adaptive and interpretable automated optimization.
\end{abstract}

\begin{IEEEImpStatement}
  Automated optimization increasingly requires AI systems that can transform natural-language requirements into mathematical models, executable solver code, and reliable debugging procedures with limited human intervention. Existing LLM-based OR agents often rely on fixed reasoning–execution pipelines, which restricts their adaptability across heterogeneous optimization tasks and makes the organization of their workflow difficult to inspect. This work advances automated optimization by representing OR-agent workflows as explicit AOE-style architecture graphs and evolving reasoning trajectories over these graphs. By making workflow topology, execution dependencies, and alternative reasoning paths explicit, EvoOR-Agent supports task-adaptive coordination among problem interpretation, mathematical formulation, solver selection, code generation, and debugging. The evolved trajectories can also expose reusable structural patterns such as formulation decomposition, solver routing, and solver-status-based revision. These capabilities may reduce manual workflow engineering, improve robustness across diverse OR scenarios, and support more transparent decision-support tools for scientific, industrial, and logistics optimization.
\end{IEEEImpStatement}


\section{Introduction}

\begin{figure}[!t]
  \centering
  \includegraphics[width=0.95\columnwidth]{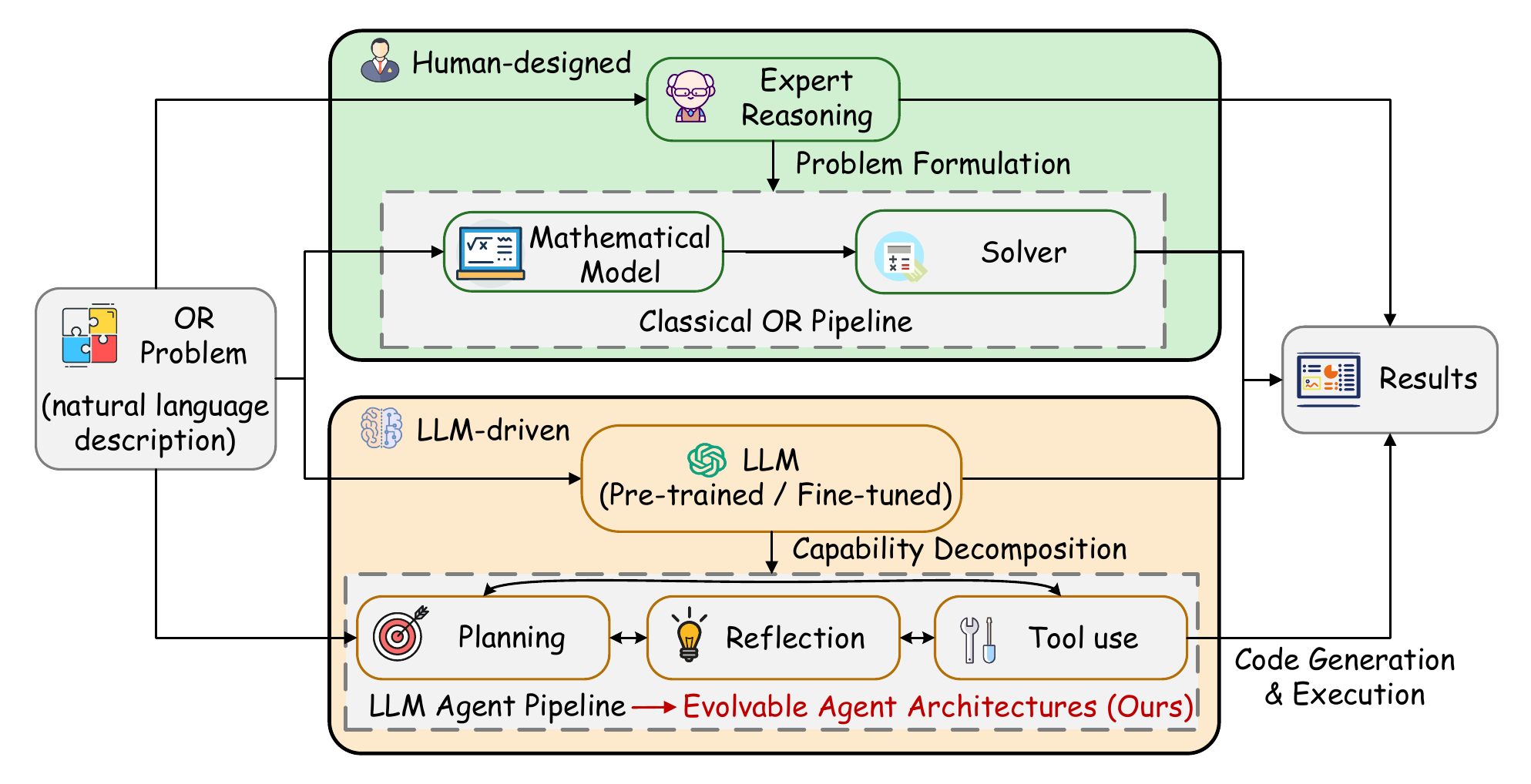}
  \caption{Evolution of OR problem-solving paradigms. The upper path illustrates the classical expert-driven OR workflow, in which practitioners formulate mathematical models from task requirements and then design or select suitable solvers. The lower path shows the LLM-driven paradigm, where planning, reflection, and tool use are integrated into an agentic reasoning--execution chain. Existing LLM-based systems improve semantic automation but usually retain fixed workflow structures, whereas our framework treats agent architecture and the reasoning trajectories instantiated on it as evolvable objects.}
  \label{fig:1}
\end{figure}

\IEEEPARstart{O}{perations} research (OR) is a fundamental methodology for scientific and industrial decision-making, with broad applications in scheduling, resource allocation, logistics, and production planning~\cite{Practicaloptimization,procurement,challengesoptimization}. As illustrated in the upper part of Fig.~\ref{fig:1}, the classical OR paradigm is largely expert-driven: practitioners translate task requirements into mathematical models and then design or select suitable solvers for the resulting optimization problems. Over the past decades, this solver layer has produced a rich family of optimization methods, including gradient-based methods~\cite{ruder2016overview}, classical heuristics~\cite{burke2013hyper}, and metaheuristic methods such as evolutionary algorithms~\cite{xu2022difficulty,zhu2026density,zou2025swarm}. Because solver performance is problem-dependent, considerable effort has also been devoted to automatic solver adaptation, including algorithm selection, configuration, and other forms of instance-specific adjustment~\cite{kerschke2019automated,huang2019survey}. These advances have substantially improved the efficiency and adaptability of solver engineering in OR.

Despite these advances, existing automated optimization methods still operate primarily at the solver layer~\cite{ma2025toward}. Most of these methods act after the optimization problem has already been formulated, and mainly focus on adapting algorithms to structured instances. As a result, they usually assume that the optimization model is already available in a machine-usable form. They do not directly address upstream stages of the OR workflow, such as interpreting natural-language requirements, extracting implicit constraints, constructing mathematical formulations, and organizing executable solution procedures. Therefore, while existing automated optimization methods automate important parts of the solving stage, they do not yet optimize the organization of the full OR pipeline itself.

Recent advances in large language models (LLMs) have created a new opportunity to extend automation beyond the solver layer toward the full OR pipeline~\cite{van2024llamea}. Because LLMs can process natural language, generate code, and perform multi-step reasoning, they have been increasingly used for problem interpretation, mathematical modeling, solver invocation, tool use, and algorithm recommendation within optimization workflows~\cite{COT,reflexion,rawal2021recent,deepseekr1,hagos2024recent}. As illustrated in the lower part of Fig.~\ref{fig:1}, this paradigm shifts OR automation from a purely expert-designed process to an agentic reasoning--execution process in which planning, reflection, and tool use are integrated into a single workflow. Nevertheless, despite this progress, most existing LLM-based OR systems still rely on predefined agent architectures or handcrafted prompting routines~\cite{lu2025optmath,Or-llm-agent,huang2025orlm}. In other words, while semantic capability has improved substantially, the architecture of the reasoning--execution workflow is still fixed in advance and is not itself optimized.

Such architectural rigidity is particularly restrictive for automated optimization. Real-world OR tasks are highly heterogeneous: some require careful extraction of implicit constraints, some depend critically on solver or algorithm selection, and others require iterative revision between modeling and execution. A fixed agent pipeline is therefore unlikely to be uniformly effective across tasks. Moreover, when adaptation is limited to prompt editing or local module refinement, the organization of the reasoning process remains implicit and cannot itself be optimized in a principled way. Consequently, current LLM-based OR systems still lack a mechanism for explicitly adapting how reasoning is structured, coordinated, and executed, rather than merely improving what is generated at individual stages. What is needed instead is a framework in which agent architecture becomes an explicit optimization object and the resulting reasoning trajectories remain structurally interpretable.

To address this challenge, we propose EvoOR-Agent, a co-evolutionary framework for automated optimization. The central idea is to treat the internal organization of an OR agent not as a fixed prompting scaffold, but as an evolvable architecture on which different reasoning trajectories can be instantiated and evaluated. Specifically, we abstract agent workflows into an activity-on-edge (AOE)-style network, where reasoning states, execution dependencies, and alternative solution paths are represented as structured components. This representation exposes the workflow topology of an agent and provides a graph-supported search space for subsequent evolutionary optimization.

Building on this representation, the proposed framework couples architecture evolution with reasoning-trajectory evolution. At the architecture level, the framework maintains a global architecture graph by inserting newly discovered reasoning structures, updating edge weights according to empirical fitness, and pruning persistently weak components. At the trajectory level, it evolves a population of reasoning individuals through graph-mediated path-conditioned recombination, multi-granularity semantic mutation, and elitist population update. In addition, an LLM-driven experience-acquisition module constructs a domain-specific knowledge base from reusable OR practices, which is then used to support initialization and knowledge-guided mutation. Through this coupling, the framework can adapt task decomposition, solver selection, tool use, and downstream execution without relying on a fully predefined reasoning--execution pipeline.

Overall, the main contributions of this work are as follows:
\begin{enumerate}
  \item We formulate agentic OR workflows as explicit \emph{agent architectures} by representing them as AOE-style networks. This converts implicit reasoning--execution organization into a structured and evolvable search space, exposing workflow topology, dependency relations, and alternative execution paths in a form that is analyzable, manipulable, and interpretable.

  \item We develop an evolutionary search mechanism for \emph{reasoning trajectories} instantiated on the maintained agent architecture. The proposed framework evolves reasoning individuals through graph-mediated path-conditioned recombination, semantic mutation, and elitist population update, while the architecture graph is updated in tandem based on the execution traces and fitness of evolved individuals. This design enables task-adaptive reasoning organization for problem formulation, solver selection, tool use, and code execution.
\end{enumerate}

We evaluate the proposed framework on seven heterogeneous OR benchmarks, including IndustryOR, MAMO, NL4OPT, BWOR, NLP4LP, and ReSocratic, covering mathematical formulation, solver-oriented reasoning, industrial optimization, and generalization scenarios. The experiments compare our method with zero-shot reasoning models, fixed-pipeline LLM-based OR agents, specialized OR modeling methods, and representative evolutionary agent frameworks under a unified evaluation protocol. The results show that treating agent architecture and reasoning trajectories as evolvable objects yields consistent improvements over static reasoning pipelines, specialized modeling methods, and representative evolutionary baselines. The ablation study further shows that both the AOE-style architecture representation and the knowledge-base-assisted evolutionary operators contribute to the final performance. In addition, the case study and evolutionary dynamics analyses provide evidence that the learned architecture graph captures interpretable reasoning structures, including formulation decomposition, solver-routing decisions, and semantic debugging behavior.

The remainder of this paper is organized as follows. Section~\ref{sec:Related_Work} reviews related work on automated optimization, LLM-based OR systems, and evolutionary approaches to LLM-related artifacts. Section~\ref{sec:Proposed} introduces the proposed framework for co-evolving agent architectures and reasoning trajectories. Section~\ref{sec:Experiments} presents the experimental setup, benchmark evaluation, and comparative results. Section~\ref{sec:Discussion} provides the mechanistic case study, ablation study, and analyses of population size, convergence behavior, and evolutionary dynamics. Finally, Section~\ref{sec:Conclusion} concludes the paper and discusses future directions.

\section{Related Work}
\label{sec:Related_Work}

In this section, we first review research on LLMs for OR, then discuss recent progress in LLM-based agents, and finally summarize representative studies on LLMs and evolutionary computation.

\subsection{LLMs for Operations Research}

Research on LLMs for OR mainly follows two directions, i.e., fine-tuned methods and prompt-based methods.

Fine-tuned methods improve domain specialization by adapting models to OR-specific corpora and modeling tasks. Representative examples include ORLM~\cite{huang2025orlm}, LLMOPT~\cite{LLMOPT}, OptiBench~\cite{Optibench}, and StepORLM~\cite{zhou2025steporlm}. These methods typically construct large-scale datasets of mathematical formulations, solver-ready scripts, or OR-specific instructions, and then fine-tune language models for downstream optimization tasks. This line of work has shown that domain adaptation can improve modeling fidelity and formal output quality on benchmarks, but it usually operates on fixed modeling interfaces or predefined task formats.

Prompt-based methods instead aim to elicit the reasoning capability of pretrained models through carefully designed prompting or multi-stage interaction. Early work on chain-of-thought prompting~\cite{COT} and zero-shot reasoning~\cite{kojima2022large} established the basis for step-by-step problem solving. Building on this, Ramamonjison et al.~\cite{ramamonjison2022augmenting} explored the possibility of using LLMs as OR scientists. OptiMUS~\cite{OptiMUS-03} further introduced modular agents for solving MILPs from long-form descriptions. Subsequent work incorporated more advanced reasoning mechanisms, such as multi-stage decomposition in OR-LLM-Agent~\cite{Or-llm-agent}, hierarchical search in OptiTree~\cite{liu2025optitree}, and process supervision in OR-PRM~\cite{wangor}, thereby improving semantic modeling and solver-oriented reasoning under predefined workflows or stage decompositions.

\subsection{LLM-based Agents}

Beyond OR-specific systems, a broader line of research has focused on LLM-based agents. Recent progress in this area has expanded the scope of autonomous reasoning and execution~\cite{=llmandagent}.

At the reasoning level, techniques such as chain-of-thought prompting~\cite{COT}, self-reflection~\cite{reflexion}, and meta-reasoning~\cite{gao2024meta} improve the ability of agents to decompose tasks, monitor intermediate decisions, and revise erroneous reasoning traces. Reasoning-oriented models such as OpenAI o1~\cite{jaech2024openai} and DeepSeek-R1~\cite{deepseekr1} further strengthen long-horizon and multi-step reasoning capabilities, making agent systems more effective for complex decision processes.

At the execution level, modern agents increasingly rely on tool use and environment interaction~\cite{masterman2024landscape}. By invoking external APIs, search tools, or domain-specific solvers, agents can go beyond purely parametric generation and access capabilities that are crucial for practical problem solving. Recent developments such as the model context protocol (MCP)~\cite{hou2025model} standardize access to external services, while skill abstractions~\cite{xu2026agent} provide reusable interfaces for modular execution.

At a larger scale, multi-agent system frameworks~\cite{hong2023metagpt,fan2023attention} decompose complex tasks into specialized roles and coordinated interactions, thereby improving modularity and task coverage. These agent frameworks provide strong reasoning, tool-use, and collaboration capabilities~\cite{chen2024multiagent}, but they usually rely on predefined architectural layouts, fixed role assignments, or manually designed interaction protocols.

\subsection{LLMs and Evolutionary Computation}

Another relevant line of work studies the interaction between LLMs and evolutionary computation (EC), including both the use of LLMs as informed evolutionary operators and the use of EC to optimize prompts, programs, and other LLM-related artifacts~\cite{wu2024evolutionary}.

In one direction, LLMs are used to generate, revise, or combine structured search points in a more semantically meaningful way than purely random perturbations. Representative examples include FunSearch~\cite{romera2024mathematical} and evolution through large models~\cite{lehman2023evolution}, which showed that LLM-guided evolution can improve candidate programs for difficult combinatorial problems. AlphaEvolve~\cite{novikov2025alphaevolve} further generalized this idea by maintaining populations of candidate programs and iteratively improving them through mutation, crossover, and selection, illustrating the role of LLMs as high-level semantic variation operators. Specifically, EvoPrompt~\cite{guo2023evoprompt} leverages LLMs to implement GA and DE operators for discrete prompt optimization. Moving toward system-level evolution, EvoAgent~\cite{yuan2025evoagent} automatically extends individual agents into diverse multi-agent systems by evolving their personas and task-specific settings.

In the other direction, EC has also been used to optimize LLM-related artifacts such as prompts, code variants, and system configurations. Examples include GEPA~\cite{gepa}, AdaEvolve~\cite{cemri2026adaevolve}, and ThetaEvolve~\cite{wang2025thetaevolve}, which study prompt optimization, adaptive resource allocation, context-efficient evolutionary updates, and learning-based variation control. This line of work shows that combining LLMs with EC is effective for searching over semantically rich and structured objects, while most existing studies still center on prompts, code, or candidate programs as the primary search artifacts.

\section{The Proposed Framework}
\label{sec:Proposed}

Automated optimization agents need to coordinate problem analysis, mathematical formulation, solver selection, tool invocation, code generation, and debugging. Existing LLM-based OR systems have improved semantic processing in these stages, but their workflows are still largely specified by handcrafted prompts or fixed pipelines. As a result, the internal organization of an agent is usually static and cannot be explicitly adapted to different optimization tasks.

To address this limitation, we model an OR agent as an evolvable architecture and optimize the reasoning trajectories instantiated on this architecture. The proposed framework maintains an AOE-style architecture graph, evolves a population of reasoning individuals on the graph, and updates both the graph and the population throughout search. This section introduces the architecture graph evolution, the reasoning trajectory evolution, and the overall co-evolutionary loop. The prompt templates and implementation details are provided in Section~S-I of the supplementary material.

\subsection{Architecture Graph Evolution}

\begin{figure*}[!t]
  \centering
  \includegraphics[width=2\columnwidth]{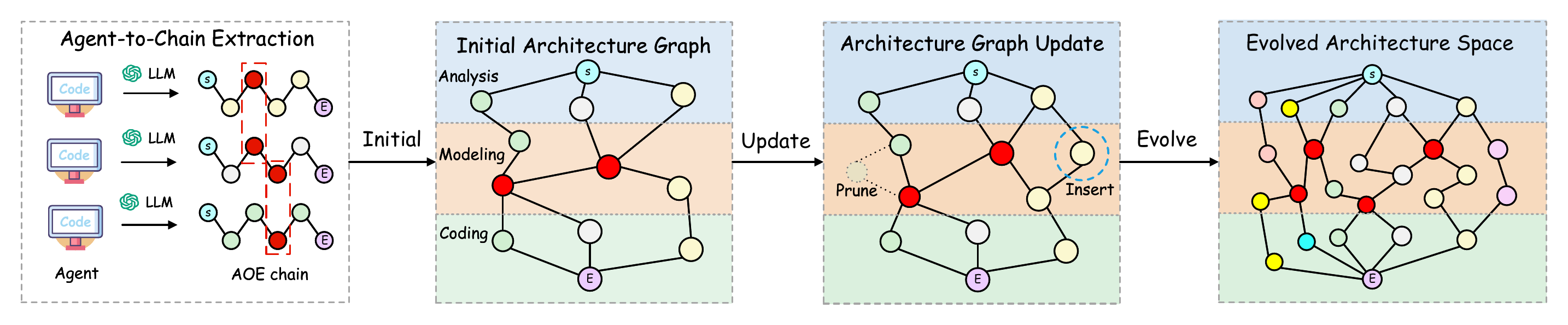}
  \caption{Architecture graph evolution. Individual OR agents are first abstracted into AOE chains, which are merged by phase-local state alignment to form the initial architecture graph. During evolution, newly discovered structures are inserted into the graph, while persistently weak nodes and edges are pruned. The maintained graph defines the current architecture space.}
  \label{fig:architecture_graph_evolution}
\end{figure*}

We first introduce how the architecture graph is represented, initialized, and updated during evolution. At iteration $t$, the architecture graph is denoted by
\[
  \mathcal{G}_t=
  \bigl(
    \mathcal{V}_t,\mathcal{E}_t,
    \pi,\rho,
    w_{\mathrm{fit}},w_{\mathrm{sparse}}
  \bigr),
\]
where $\mathcal{V}_t$ is the set of reasoning states, $\mathcal{E}_t$ is the set of executable transitions, $\pi:\mathcal{V}_t\rightarrow\Pi$ assigns each state to an OR phase, $\rho:\mathcal{E}_t\rightarrow\{\mathrm{work},\mathrm{reason},\mathrm{tool}\}$ assigns each edge to an edge type, and $w_{\mathrm{fit}}$ and $w_{\mathrm{sparse}}$ record the fitness and sparsity weights of edges. The graph exposes the workflow topology of the agent and provides a structured search space for reasoning trajectories.

Each instantiated reasoning individual $a$ is abstracted as a phase-wise AOE chain,
\begin{equation}
  \mathcal{C}(a)=
  \bigl(
    \mathcal{C}_{\mathrm{ana}}(a),
    \mathcal{C}_{\mathrm{mod}}(a),
    \mathcal{C}_{\mathrm{code}}(a)
  \bigr),
  \label{eq:aoe_chain}
\end{equation}
where $\mathcal{C}_{\mathrm{ana}}(a)$, $\mathcal{C}_{\mathrm{mod}}(a)$, and $\mathcal{C}_{\mathrm{code}}(a)$ denote the subchains corresponding to problem analysis, mathematical modeling, and code generation, respectively. Each subchain is a directed node-edge sequence,
\[
  \mathcal{C}_{p}(a)=
  \bigl(
    v^{p}_{0},e^{p}_{1},v^{p}_{1},
    \ldots,
    e^{p}_{\ell_p},v^{p}_{\ell_p}
  \bigr),
  \quad p\in\Pi .
\]
The complete chain records the structured execution trace of the agent while preserving the phase order of the OR workflow. We define $\Pi$ as the ordered set of OR phases, including problem analysis, mathematical modeling, and code generation. These phases provide a structural backbone for state alignment, graph construction, and mutation.

Within the graph, each node represents an intermediate reasoning state or a phase boundary, and each directed edge represents an executable transition between two states. The edge set is partitioned as
\[
  \mathcal{E}_t =
  \mathcal{E}_{\mathrm{work}}
  \cup
  \mathcal{E}_{\mathrm{reason}}
  \cup
  \mathcal{E}_{\mathrm{tool}} .
\]
Edges in $\mathcal{E}_{\mathrm{work}}$ describe inter-phase workflow transitions, edges in $\mathcal{E}_{\mathrm{reason}}$ describe finer-grained reasoning variations, and edges in $\mathcal{E}_{\mathrm{tool}}$ describe calls to external tools or utility interfaces. This partition separates stage-level organization, intra-phase reasoning, and tool invocation. It also allows the framework to extract an AOE chain from an executed individual and to instantiate an executable individual from a feasible path in the architecture graph.

\begin{algorithm}[h]
  \caption{Initial Architecture Graph Construction}
  \label{alg:init_architecture}
  \small
  \begin{algorithmic}[1]
    \STATE \textbf{Input} Initial parent population $\mathcal{P}_0$, ordered phase set $\Pi$
    \STATE \textbf{Output} Initial architecture graph $\mathcal{G}_0$

    \STATE $\mathcal{G}_0 \leftarrow \textsc{InitializeGraph}()$
    \STATE $\mathcal{S} \leftarrow \emptyset$

    \FOR{each individual $a \in \mathcal{P}_0$}
    \STATE Extract a phase-wise AOE chain $\mathcal{C}(a)$ from $a$
    \STATE $\mathcal{S} \leftarrow \mathcal{S} \cup \{\mathcal{C}(a)\}$
    \ENDFOR

    \FOR{each phase $p \in \Pi$}
    \STATE Collect all states assigned to phase $p$
    \STATE Group semantically equivalent states using \textsc{LLMJudge}
    \STATE Replace each group with a representative state
    \ENDFOR

    \FOR{each chain $\mathcal{C} \in \mathcal{S}$}
    \STATE Insert its aligned states and directed edges into $\mathcal{G}_0$
    \ENDFOR

    \RETURN $\mathcal{G}_0$
  \end{algorithmic}
  \normalsize
\end{algorithm}

The initial architecture graph $\mathcal{G}_0$ is constructed from the initial parent population. As shown in Algorithm~\ref{alg:init_architecture}, each individual is first converted into a phase-wise AOE chain. Semantically similar states are then merged within each phase. During this process, the LLM judges state equivalence by considering the semantic role of a state, the function of its incident edges, and its position in the problem-solving workflow. The aligned states and retained directed edges are inserted into $\mathcal{G}_0$, which defines the initial architecture search space.

Two structural constraints are imposed during graph construction. The first is phase-local merging, which restricts state alignment to nodes belonging to the same OR phase. This prevents semantically similar but functionally different states from being merged across phases. The second is the non-skipping constraint, which requires every feasible source-to-sink path to visit the mandatory phase boundaries in the prescribed order. These constraints preserve the global execution order while still allowing multiple admissible workflow variants within and across phases.

The architecture graph is updated after each evolutionary iteration. The update incorporates newly discovered reasoning structures into the graph and reweights existing structures according to the fitness of individuals that traverse them. Algorithm~\ref{alg:update_architecture} summarizes this procedure.

\begin{algorithm}[h!]
  \caption{Architecture Graph Update}
  \label{alg:update_architecture}
  \small
  \begin{algorithmic}[1]
    \STATE \textbf{Input} Current architecture graph $\mathcal{G}_t$, evaluated population $\mathcal{P}_{t+1}$
    \STATE \hspace{1em} Fitness map $\mathcal{F}_{t+1}$, update parameters $\Omega$
    \STATE \textbf{Output} Updated architecture graph $\mathcal{G}_{t+1}$

    \STATE Unpack $\Omega$ as $(\alpha,\tau,\sigma)$
    \STATE $\widetilde{\mathcal{G}}_t \leftarrow \textsc{StateMerging}(\mathcal{P}_{t+1})$
    \STATE $\mathcal{G}_{t+1} \leftarrow \textsc{GraphUnion}(\mathcal{G}_t,\widetilde{\mathcal{G}}_t)$

    \FOR{each edge $e \in \mathcal{E}(\mathcal{G}_{t+1})$}
    \STATE $\mathcal{A}_t(e) \leftarrow \{a \in \mathcal{P}_{t+1} \mid e \in \mathrm{path}(a)\}$
    \IF{$|\mathcal{A}_t(e)| > 0$}
    \STATE Update $w_{\mathrm{fit}}(e)$ using Eq.~\eqref{eq:fit_weight_update}
    \ENDIF
    \STATE Update $\mathrm{count}_t(e)$ and $w_{\mathrm{sparse}}(e)$ using Eq.~\eqref{eq:sparse_weight}
    \ENDFOR

    \FOR{each node $v \in \mathcal{V}(\mathcal{G}_{t+1})$}
    \STATE Compute $w(v)$ using Eq.~\eqref{eq:node_weight}
    \IF{$w(v) < \tau$ for $\sigma$ consecutive iterations}
    \STATE Prune $v$ from $\mathcal{G}_{t+1}$
    \ENDIF
    \ENDFOR

    \FOR{each edge $e \in \mathcal{E}(\mathcal{G}_{t+1})$}
    \IF{$w_{\mathrm{fit}}(e) < \tau$ for $\sigma$ consecutive iterations}
    \STATE Prune $e$ from $\mathcal{G}_{t+1}$
    \ENDIF
    \ENDFOR

    \RETURN $\mathcal{G}_{t+1}$
  \end{algorithmic}
  \normalsize
\end{algorithm}

To incorporate newly discovered structures, we first extract the local graph induced by the reasoning paths in the newly evaluated population and then merge it into the current architecture graph,
\[
  \mathcal{G}_{t+1} \leftarrow
  \textsc{GraphUnion}
  \bigl(
    \mathcal{G}_t,
    \textsc{StateMerging}(\mathcal{P}_{t+1})
  \bigr).
\]
This operation makes newly generated reasoning structures searchable in later generations.

After evaluation, the fitness of individuals is propagated to their traversed edges. For each edge $e \in \mathcal{E}(\mathcal{G}_{t+1})$, let
\[
  \mathcal{A}_t(e)=
  \{a\in \mathcal{P}_{t+1} \mid e\in \mathrm{path}(a)\}.
\]
If $\mathcal{A}_t(e)$ is nonempty, the fitness weight of $e$ is updated by
\begin{equation}
  w_{\mathrm{fit}}(e)\leftarrow
  w_{\mathrm{fit}}(e)+\alpha
  \left(
    \frac{1}{|\mathcal{A}_t(e)|}
    \sum_{a\in \mathcal{A}_t(e)}
    \mathcal{F}_{t+1}(a)
    -
    w_{\mathrm{fit}}(e)
  \right),
  \label{eq:fit_weight_update}
\end{equation}
where $\alpha \in (0,1]$ is the architecture learning rate. Edges that frequently appear in high-performing reasoning trajectories therefore receive larger fitness weights.

To encourage exploration, we also maintain a sparsity weight for each edge,
\begin{equation}
  w_{\mathrm{sparse}}(e)=
  \frac{1}
  {\log(2+\mathrm{count}_t(e))+\epsilon},
  \label{eq:sparse_weight}
\end{equation}
where $\mathrm{count}_t(e)$ is the cumulative number of traversals of $e$ up to iteration $t$, and $\epsilon>0$ is a smoothing constant. Frequently used edges receive smaller sparsity weights, which encourages subsequent path sampling to revisit underexplored regions of the graph.

Node scores are derived from incident edge weights rather than maintained independently,
\begin{equation}
  w(v)=
  \frac{1}{|\mathcal{E}(v)|}
  \sum_{e\in \mathcal{E}(v)}
  w_{\mathrm{fit}}(e),
  \label{eq:node_weight}
\end{equation}
where $\mathcal{E}(v)$ denotes the set of incident edges of $v$. Nodes and edges whose scores remain below the pruning threshold $\tau$ for $\sigma$ consecutive iterations are removed from the graph. This forgetting mechanism prevents obsolete structures from accumulating and keeps the architecture graph compact.

Fig.~\ref{fig:architecture_graph_evolution} illustrates the architecture-level evolution process. AOE chains extracted from individual agents are merged into an initial architecture graph. During evolution, newly discovered structures are inserted, and persistently weak components are pruned. The maintained graph then provides the architecture space for reasoning evolution. Implementation-level prompt templates for code-to-AOE extraction, state merging, AOE-to-code synthesis, semantic mutation, and the complete AOE network representation are provided in Section~S-I of the supplementary material.

\subsection{Reasoning Trajectory Evolution}

Given the current architecture graph, the framework evolves a population of reasoning individuals on this graph. The reasoning-level search includes hybrid initialization, path-conditioned recombination, semantic mutation, and elitist multi-source selection. These operators search for effective trajectories that coordinate problem formulation, solver selection, tool use, code generation, and debugging, with details provided in Section~S-I of the supplementary material.

To support reasoning trajectory evolution with reusable domain experience, we construct a domain-specific knowledge base $K$ through an LLM-agent-based experience acquisition workflow. The workflow retrieves relevant OR papers and code repositories, extracts reusable modeling patterns, solver templates, and implementation snippets, and organizes them into a structured repository. The knowledge base is used in two parts of the evolutionary process. It supports the initialization of reasoning individuals, and it guides semantic mutation when prior OR practices are useful for repairing or improving an individual. To prevent benchmark-specific information from being introduced into the evolutionary process, the construction and use of $K$ follow a leakage-control protocol, with details provided in Section~S-III of the supplementary material.

\begin{figure*}[!t]
  \centering
  \includegraphics[width=\textwidth]{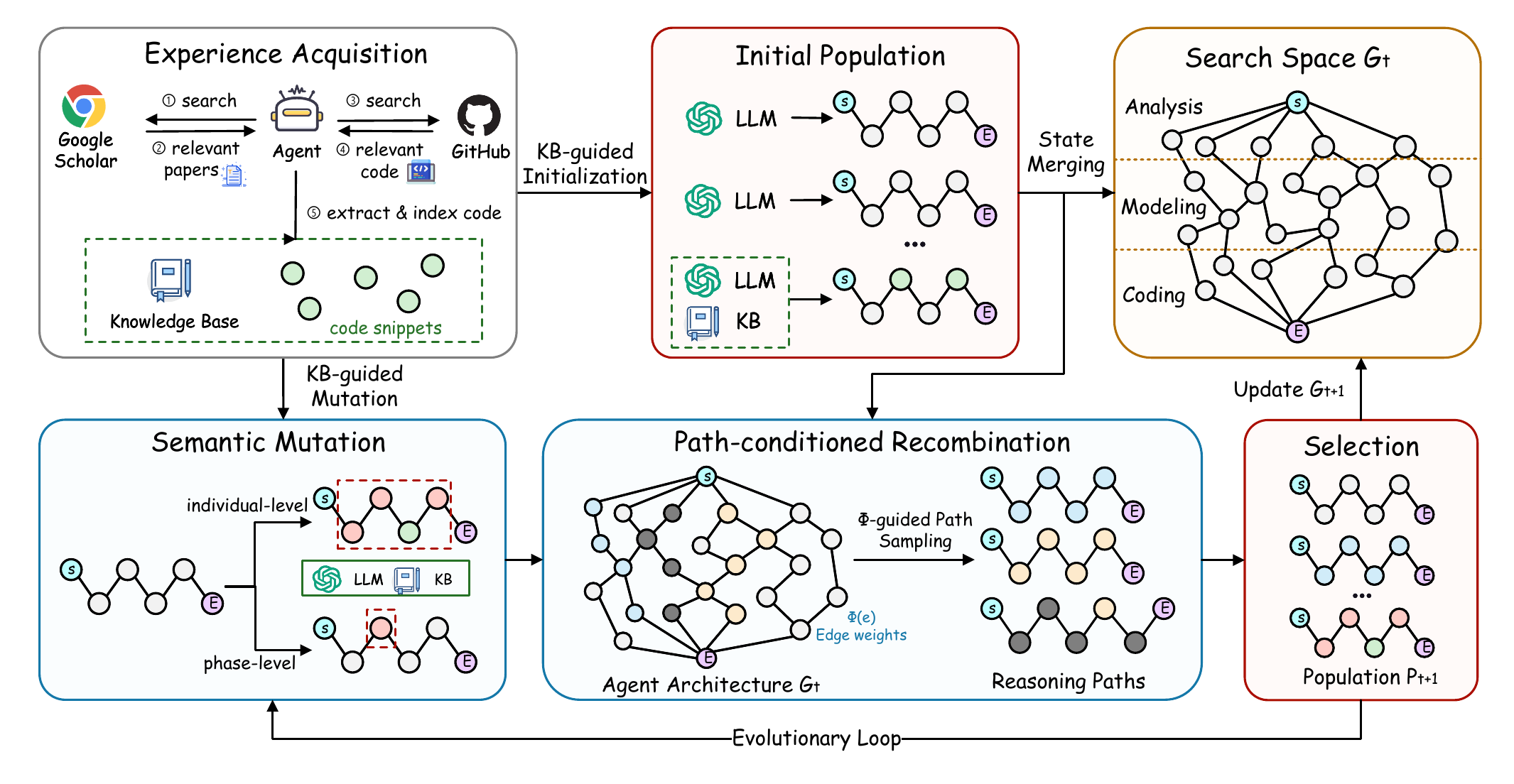}
  \caption{Overview of reasoning trajectory evolution on the current architecture graph. An LLM-agent-based experience acquisition workflow retrieves relevant papers and code repositories, and organizes reusable OR practices into a domain-specific knowledge base. The knowledge base supports initialization and semantic mutation. During each generation, path-conditioned recombination samples new reasoning trajectories from the current graph, semantic mutation revises selected individuals at different granularities, and multi-source selection forms the next population.}
  \label{fig:reasoning_trajectory_evolution}
\end{figure*}

Fig.~\ref{fig:reasoning_trajectory_evolution} provides an overview of reasoning trajectory evolution. The knowledge base provides priors for initialization and mutation. The architecture graph provides the structural search space for path-conditioned recombination. The population is updated by integrating elite individuals, recombined individuals, and mutated individuals.

\subsubsection{Initialization}

\begin{algorithm}[h!]
  \caption{Hybrid Population Initialization}
  \label{alg:init_reasoning}
  \small
  \begin{algorithmic}[1]
    \STATE \textbf{Input} Parent population size $N$, initialization ratio $\alpha_{\mathrm{init}}$, knowledge base $K$
    \STATE \textbf{Output} Initial reasoning population $\mathcal{P}_0$

    \STATE $\mathcal{P}_0 \leftarrow \emptyset$
    \STATE $N_{\mathrm{KB}} \leftarrow \lfloor N \cdot \alpha_{\mathrm{init}} \rfloor$

    \FOR{$i=1$ to $N$}
    \IF{$i \leq N_{\mathrm{KB}}$}
    \STATE $a \leftarrow \textsc{GenerateIndividual}(K,\mathrm{LLM})$
    \ELSE
    \STATE $a \leftarrow \textsc{GenerateIndividual}(\mathrm{LLM})$
    \ENDIF
    \STATE $\mathcal{P}_0 \leftarrow \textsc{Append}(\mathcal{P}_0,a)$
    \ENDFOR

    \RETURN $\mathcal{P}_0$
  \end{algorithmic}
  \normalsize
\end{algorithm}

Reasoning trajectory evolution begins with an initial parent population of size $N$. Algorithm~\ref{alg:init_reasoning} uses a hybrid initialization strategy controlled by $\alpha_{\mathrm{init}}$. A portion of the population is generated with the support of the knowledge base, where the LLM synthesizes reasoning trajectories by using retrieved OR principles and implementation patterns. The remaining individuals are generated directly by the LLM without knowledge augmentation. This design reduces cold-start difficulty while preserving structural diversity in the initial population. The resulting individuals are then abstracted into phase-wise AOE chains and used to construct the initial architecture graph.

\subsubsection{Path-conditioned Recombination}

Once an architecture graph is available, new reasoning individuals can be generated by sampling feasible paths on the graph. Each feasible path corresponds to an executable reasoning trajectory. Since the architecture graph aggregates states and transitions extracted from the population, sampling a new path recombines reasoning fragments discovered by different individuals. The operator is therefore a population-level graph-mediated recombination mechanism rather than a pairwise crossover between two parent strings.

\begin{algorithm}[h!]
  \caption{Path-conditioned Trajectory Recombination}
  \label{alg:recombine_reasoning}
  \small
  \begin{algorithmic}[1]
    \STATE \textbf{Input} Architecture graph $\mathcal{G}_t$, parent population $\mathcal{P}_t$, exploration parameter $\gamma$, number of candidates $\kappa$
    \STATE \textbf{Output} Recombined individual $a_{\mathrm{rec}}$

    \FOR{each edge $e \in \mathcal{E}(\mathcal{G}_t)$}
    \STATE Construct $\Phi(e)=\gamma\,\bar{w}_{\mathrm{fit}}(e)+(1-\gamma)\,\bar{w}_{\mathrm{sparse}}(e)$
    \ENDFOR

    \STATE $\mathcal{L}_{\mathrm{best}} \leftarrow \textsc{WeightedPathSearch}(\mathcal{G}_t, \mathcal{P}_t, \Phi, \kappa)$
    \STATE $a_{\mathrm{rec}} \leftarrow \textsc{InstantiateIndividual}(\mathcal{L}_{\mathrm{best}})$
    \RETURN $a_{\mathrm{rec}}$
  \end{algorithmic}
  \normalsize
\end{algorithm}

Path sampling is guided by the edge-selection score
\begin{equation}
  \Phi(e)=
  \gamma\,\bar{w}_{\mathrm{fit}}(e)
  +(1-\gamma)\,\bar{w}_{\mathrm{sparse}}(e),
  \label{eq:path_score}
\end{equation}
where $\gamma \in [0,1]$ controls the balance between exploitation and exploration. The normalized weight $\bar{w}_{\mathrm{fit}}(e)$ favors transitions associated with high-performing individuals, while $\bar{w}_{\mathrm{sparse}}(e)$ favors underexplored transitions. Min-max normalization is used for both terms, and a constant normalized value is assigned when all edge weights are identical.

The procedure \textsc{WeightedPathSearch} performs a randomized search to collect a set of candidate paths that have not been instantiated by the current parent population. Specifically, it repeatedly performs weighted random walks on the architecture graph starting from the initial node and terminating at a valid end node. Each walk produces a feasible path $\mathcal{L}$. The walk is guided by the per-edge probability derived from $\Phi(e)$, where edges with higher scores are more likely to be selected. This process continues until a predetermined number $\kappa$ of distinct, unvisited paths are collected. For each candidate path $\mathcal{L}$, the path score is computed as the average edge score,
\begin{equation}
  S(\mathcal{L})=\frac{1}{|\mathcal{L}|}\sum_{e\in\mathcal{L}}\Phi(e),
  \label{eq:weighted_path_score}
\end{equation}
where $|\mathcal{L}|$ denotes the number of edges in the path. Among the collected $\kappa$ candidates, the feasible unvisited path with the highest $S(\mathcal{L})$ is selected and converted into an executable reasoning individual.

\subsubsection{Semantic Mutation}

Semantic mutation directly revises existing reasoning individuals. The operator is controlled by the mutation guidance rate $\beta_{\mathrm{learn}}$ and the mutation scope rate $\beta_{\mathrm{strat}}$. The guidance rate determines whether mutation uses the knowledge base, while the scope rate determines whether mutation is applied to a single phase or to the whole individual.

\begin{algorithm}[h!]
  \caption{Multi-granularity Semantic Mutation}
  \label{alg:mutate_reasoning}
  \small
  \begin{algorithmic}[1]
    \STATE \textbf{Input} Target individual $a$, knowledge base $K$, mutation guidance rate $\beta_{\mathrm{learn}}$, mutation scope rate $\beta_{\mathrm{strat}}$
    \STATE \textbf{Output} Mutated individual $a_{\mathrm{mut}}$

    \STATE Sample mutation scope according to $\beta_{\mathrm{strat}}$
    \IF{phase-level mutation is selected}
    \STATE $s \leftarrow \textsc{PhaseLevel}$
    \ELSE
    \STATE $s \leftarrow \textsc{WholeIndividual}$
    \ENDIF

    \STATE Sample guidance mode according to $\beta_{\mathrm{learn}}$
    \IF{knowledge-guided mode is selected}
    \STATE $a_{\mathrm{mut}} \leftarrow \textsc{SemanticMutate}(a,s,K)$
    \ELSE
    \STATE $a_{\mathrm{mut}} \leftarrow \textsc{SemanticMutate}(a,s)$
    \ENDIF

    \RETURN $a_{\mathrm{mut}}$
  \end{algorithmic}
  \normalsize
\end{algorithm}

In the knowledge-guided mode, the LLM uses $K$ to identify and revise weaknesses in the current individual, including missing constraints, inappropriate formulations, or unsuitable solver choices. In the unguided mode, the LLM performs semantic perturbation without external knowledge support, which helps maintain diversity.

Mutation is performed at two granularities. Phase-level mutation revises the reasoning content within one selected OR phase while preserving the remaining phases. This mode is suitable for local refinement, such as improving the modeling stage or correcting solver selection. Individual-level mutation rewrites the reasoning trajectory at the whole-process level and allows larger semantic changes. In both cases, mutation changes the semantic realization of a reasoning trajectory under the current architecture. It does not directly modify the architecture graph. Algorithm~\ref{alg:mutate_reasoning} summarizes this operator.

\subsubsection{Multi-source Selection}

The population update follows an elitist multi-source scheme. In each generation, the next population is constructed from three sources, including elite individuals inherited from the current population, recombined individuals sampled from the architecture graph, and mutated individuals generated from selected parents. This design preserves high-quality reasoning trajectories while continuously introducing new structural and semantic variants.

\begin{algorithm}[h!]
  \caption{Elitist Multi-source Selection}
  \label{alg:multi_source_selection}
  \small
  \begin{algorithmic}[1]
    \STATE \textbf{Input} Parent population $\mathcal{P}_t$, fitness map $\mathcal{F}_t$, architecture graph $\mathcal{G}_t$
    \STATE \hspace{1em} Knowledge base $K$, population-update parameters $\Theta_t$
    \STATE \textbf{Output} Next population $\mathcal{P}_{t+1}$

    \STATE Unpack $\Theta_t$ as $(N_{\mathrm{elite}},N_{\mathrm{rec}},N_{\mathrm{mut}},\gamma,\beta_{\mathrm{learn}},\beta_{\mathrm{strat}})$
    \STATE $\mathcal{P}_{t+1} \leftarrow \emptyset$
    \STATE $\mathcal{P}_{t+1} \leftarrow \textsc{GetBest}(\mathcal{P}_t,\mathcal{F}_t,N_{\mathrm{elite}})$

    \FOR{$i=1$ to $N_{\mathrm{rec}}$}
    \STATE $a_{\mathrm{rec}} \leftarrow \textsc{PathRecombine}(\mathcal{G}_t,\mathcal{P}_t,\gamma)$
    \STATE $\mathcal{P}_{t+1} \leftarrow \textsc{Append}(\mathcal{P}_{t+1},a_{\mathrm{rec}})$
    \ENDFOR

    \FOR{$j=1$ to $N_{\mathrm{mut}}$}
    \STATE $a_{\mathrm{target}} \leftarrow \textsc{RandomSelect}(\mathcal{P}_t)$
    \STATE $a_{\mathrm{mut}} \leftarrow \textsc{SemanticMutate}(a_{\mathrm{target}},K,\beta_{\mathrm{learn}},\beta_{\mathrm{strat}})$
    \STATE $\mathcal{P}_{t+1} \leftarrow \textsc{Append}(\mathcal{P}_{t+1},a_{\mathrm{mut}})$
    \ENDFOR

    \RETURN $\mathcal{P}_{t+1}$
  \end{algorithmic}
  \normalsize
\end{algorithm}

Algorithm~\ref{alg:multi_source_selection} combines inherited and newly generated individuals. Elite inheritance directly transfers the best individuals from $\mathcal{P}_t$ to $\mathcal{P}_{t+1}$. Path-conditioned recombination samples new feasible trajectories from $\mathcal{G}_t$ according to the edge-selection scores. Semantic mutation revises selected parent individuals at different granularities. The three sources jointly form the next population. The new population is then evaluated and used to update the architecture graph.

\subsection{Architecture and Reasoning Co-evolution}

The full framework couples architecture graph evolution and reasoning trajectory evolution into a unified evolutionary loop. The architecture graph provides the structural search space for generating reasoning trajectories, while the evaluated reasoning population provides execution traces and fitness feedback for updating the architecture graph. Algorithm~\ref{alg:framework_main} summarizes the overall procedure.

\begin{algorithm}[h!]
  \caption{Architecture and Reasoning Co-evolution}
  \label{alg:framework_main}
  \small
  \begin{algorithmic}[1]
    \STATE \textbf{Input} Knowledge base $K$, population size $N$, number of iterations $T$, Initialization ratio $\alpha_{\mathrm{init}}$, architecture learning rate $\alpha$, Mutation ratio $\beta_{\mathrm{mut}}$, mutation guidance rate $\beta_{\mathrm{learn}}$, Mutation scope rate $\beta_{\mathrm{strat}}$, exploration parameter $\gamma$, Pruning threshold $\tau$, forgetting horizon $\sigma$, elite rate $\beta_{\mathrm{elite}}$
    \STATE \textbf{Output} Final reasoning population $\mathcal{P}_T$, final architecture graph $\mathcal{G}_T$

    \STATE $\mathcal{P}_0 \leftarrow \textsc{HybridInitialize}(N,\alpha_{\mathrm{init}},K)$
    \STATE $\mathcal{G}_0 \leftarrow$ $\textsc{InitializeArchitectureGraph}(\mathcal{P}_0,\Pi)$
    \STATE $\mathcal{F}_0 \leftarrow \textsc{EvaluatePopulation}(\mathcal{P}_0)$

    \FOR{$t=0$ to $T-1$}
    \STATE $N_{\mathrm{mut}} \leftarrow \lfloor N \cdot \beta_{\mathrm{mut}} \rfloor$
    \STATE $N_{\mathrm{elite}} \leftarrow \lfloor N \cdot \beta_{\mathrm{elite}} \rfloor$
    \STATE $N_{\mathrm{rec}} \leftarrow N-N_{\mathrm{mut}}-N_{\mathrm{elite}}$

    \STATE $\Theta_t \leftarrow$ $(N_{\mathrm{elite}},N_{\mathrm{rec}},N_{\mathrm{mut}},\gamma,\beta_{\mathrm{learn}},\beta_{\mathrm{strat}})$

    \STATE $\mathcal{P}_{t+1} \leftarrow$ $\textsc{MultiSourceSelection}(\mathcal{P}_t,\mathcal{F}_t,\mathcal{G}_t,K,\Theta_t)$

    \STATE $\mathcal{F}_{t+1} \leftarrow \textsc{EvaluatePopulation}(\mathcal{P}_{t+1})$

    \STATE $\Omega \leftarrow (\alpha,\tau,\sigma)$

    \STATE $\mathcal{G}_{t+1} \leftarrow$  $\textsc{UpdateArchitectureGraph}(\mathcal{G}_t,\mathcal{P}_{t+1},\mathcal{F}_{t+1},\Omega)$
    \ENDFOR

    \RETURN $\mathcal{P}_T,\mathcal{G}_T$
  \end{algorithmic}
  \normalsize
\end{algorithm}

For compact notation, $\Theta_t$ collects the parameters used by \textsc{MultiSourceSelection}, and $\Omega$ collects the parameters used by \textsc{UpdateArchitectureGraph}. The procedure starts by generating the initial reasoning population with \textsc{HybridInitialize}. The initialized individuals are then abstracted into phase-wise AOE chains and merged by \textsc{InitializeArchitectureGraph} to construct the initial architecture graph. After evaluation, the pair $(\mathcal{P}_0,\mathcal{G}_0)$, together with the fitness map $\mathcal{F}_0$, defines the initial state of the co-evolutionary process.

At iteration $t$, the current architecture graph $\mathcal{G}_t$ guides the generation of new reasoning trajectories. Path-conditioned recombination samples feasible trajectories from the graph, while semantic mutation revises selected parent individuals with or without knowledge-base guidance. Elite inheritance preserves high-performing individuals from the current population. These three sources are integrated by \textsc{MultiSourceSelection} to form the next population $\mathcal{P}_{t+1}$.

The newly formed population is then evaluated to obtain $\mathcal{F}_{t+1}$. The evaluated individuals provide two types of feedback to architecture graph evolution. Their execution traces indicate which states and transitions should be inserted into the graph, and their fitness values determine how the traversed edges are reweighted. Persistently weak nodes and edges are pruned according to the forgetting mechanism. The updated graph $\mathcal{G}_{t+1}$ then serves as the architecture space for the next generation.

This procedure creates a bidirectional coupling between the two evolutionary levels. Reasoning trajectory evolution explores executable paths on the current architecture graph, while architecture graph evolution accumulates, reweights, and prunes structural patterns discovered by the population. Through this repeated interaction, the framework gradually refines both the population of reasoning individuals and the architecture graph that constrains their search. After $T$ iterations, the framework returns the final reasoning population $\mathcal{P}_T$ and the final architecture graph $\mathcal{G}_T$. The final population records high-quality reasoning trajectories, while the final graph records the structural patterns accumulated during search.

\section{Experiments}
\label{sec:Experiments}

We evaluate \textit{EvoOR-Agent} through comparative experiments across heterogeneous OR tasks. The evaluation addresses four core questions. First, we examine whether \textit{EvoOR-Agent} improves over zero-shot LLMs, prompt-optimization baselines, and representative agent frameworks. Second, we compare our training-free framework against specialized fine-tuned operations research models. Third, we analyze whether the proposed co-evolutionary architecture provides stable performance gains across different foundation models. Fourth, we investigate the computational overhead by analyzing the token consumption patterns during the search and inference phases. Detailed benchmark statistics, API configurations, and software/hardware environments are provided in Section~S-II of the supplementary material. The experimental code and full implementation of
this paper are available in the anonymous repository at \url{https://anonymous.4open.science/r/pL8xY2mQ8rA5fC3vB7nK}.

\subsection{Experimental Setup}

\subsubsection*{Benchmarks and Evaluation Metrics}

This evaluation employs a comprehensive suite of seven benchmark datasets covering both academic and industrial OR scenarios. Originally, NL4OPT~\cite{ramamonjison2022augmenting} contained 289 linear programming problems. MAMO~\cite{huang2024mamo} evaluates mathematical modeling ability, comprising 652 problems in EasyLP and 211 problems in ComplexLP. NLP4LP~\cite{OptiMUS-03} includes 269 abstract optimization formulations, while IndustryOR~\cite{huang2025orlm} contains 100 real-world OR problems collected from various industrial sectors. ReSocratic~\cite{Optibench} consists of 605 formatted reasoning instances, and BWOR~\cite{Or-llm-agent} comprises textbook-level OR problems translated into LaTeX-formatted English, which require exact solver-based optimization to reach ground-truth objective values.

However, substantial text formatting anomalies and incorrect ground-truth annotations are identified in these original public subsets, severely compromising the reliability of experimental evaluations. Consequently, six of these benchmarks are evaluated using their post-cleaned versions~\cite{xiao2025survey}. The systematic changes in instance counts before and after the filtering protocol are contrasted in detail in Table~\ref{tab:cleaning_comparison}. In total, 1,564 cleaned OR instances are used in the benchmark evaluation. A detailed description of the data is provided in Section~S-II-A of the supplementary material.

\begin{table}[htbp]
  \centering
  \caption{Benchmark dataset scales before and after data cleaning.}
  \label{tab:cleaning_comparison}
  \scriptsize
  \setlength{\tabcolsep}{2.2pt}
  \renewcommand{\arraystretch}{1.08}
  \begin{tabular*}{\columnwidth}{@{\extracolsep{\fill}} l c c c c c c c c @{}}
    \toprule
    \rowcolor{gray!8}
    \textbf{Scale} & \textbf{N4O} & \textbf{ELP} & \textbf{CLP} & \textbf{N4LP} & \textbf{IOR} & \textbf{ReSoc} & \textbf{BWOR} & \textbf{Total} \\
    \midrule
    \textbf{Original} & 289 & 652 & 211 & 269 & 100 & 605 & 82 & \textbf{2,208} \\
    \textbf{Cleaned} & 213 & 545 & 111 & 178 & 42 & 403 & 82 & \textbf{1,564} \\
    \bottomrule
  \end{tabular*}
  \vspace{6pt}

  \centering
  \parbox{0.96\columnwidth}{\centering\scriptsize \textit{Note:} N4O = NL4OPT; ELP = EasyLP; CLP = ComplexLP; N4LP = NLP4LP; IOR = IndustryOR; ReSoc = ReSocratic.}
\end{table}

For final evaluation, we use standard accuracy. A prediction is regarded as correct if the relative error between the generated objective value and the ground-truth objective value is no larger than $\delta=10^{-3}$. For an evaluation subset $\mathcal{D}$, accuracy is computed as
\begin{equation}
  A(\mathcal{D}) =
  \frac{1}{|\mathcal{D}|}
  \sum_{i\in\mathcal{D}}
  \mathbb{I}
  \left[
    \frac{|\hat{y}_i-y_i|}{\max(|y_i|,\epsilon_y)}
    \leq \delta
  \right],
  \label{eq:accuracy}
\end{equation}
where $\hat{y}_i$ is the generated objective value, $y_i$ is the ground-truth objective value, and $\epsilon_y$ is a small constant used only to avoid division by zero.

\subsubsection*{Language Models}

We instantiate \textit{EvoOR-Agent} with four reasoning-oriented foundation models, including DeepSeek-v3.2~\cite{liu2025deepseek}, GPT-5~\cite{singh2025openai}, Gemini 3 Flash~\cite{deepmind2025gemini3}, and Qwen 3Max~\cite{yang2025qwen3}. These models are selected to cover different reasoning, coding, and long-context capabilities. All LLM calls are made through official APIs, and a detailed description of the configuration is provided in Section~S-II-B of the supplementary material.

\subsubsection*{Evolutionary Training and Experimental Settings}

The framework is implemented in Python, and all LLM calls are made via official APIs. To ensure fair comparison across three evolutionary paradigms, we impose a strict computational budget of 400,000 tokens as the termination criterion for all evolutionary processes, instead of using fixed generations or population sizes. This budget is based on the fact that \textit{EvoOR-Agent} with population size $N=10$ and generation limit $T=8$ consumes roughly 400,000 tokens.

Five of the seven benchmark datasets are used in the evolutionary training phase, while the remaining two—NLP4LP and ReSocratic—are reserved exclusively for evaluating out-of-distribution generalization. Before evolutionary training, we construct a fixed training–test split from the 983 OR instances belonging to these five benchmarks. Let $\mathcal{S}$ denote the set of these five evaluation subsets. To prevent any single benchmark with a large sample size from dominating the selection process, each subset $s \in \mathcal{S}$ is assigned an equal subset-level fitness weight $\omega_s = 1$, resulting in a balanced $1:1:1:1:1$ weight ratio across the evolutionary benchmarks. These weights are used solely for constructing the training split and computing evolutionary fitness.

For an instance $i$ belonging to subset $s$, its instance-level training weight is defined as
\begin{equation}
  u_i = \frac{\omega_s}{|\mathcal{D}_s|},
  \label{eq:instance_weight}
\end{equation}
where $\mathcal{D}_s$ denotes the set of instances in subset $s$. The training set $\mathcal{D}_{\mathrm{train}}$ is sampled without replacement using $u_i$ as the sampling weight, subject to the constraint that it contains 120 instances and accounts for approximately 15\% of the total weighted mass. The remaining 863 instances form the evolutionary test set $\mathcal{D}_{\mathrm{test}}$. The split is fixed before evolution begins and is shared across all compared methods and all independent runs. Test instances are not used for evolutionary fitness evaluation, architecture-graph updates, or model selection.

During evolutionary training, weighted accuracy serves as the fitness value. The purpose of this weighting is to prevent large but relatively simple subsets from dominating evolutionary selection. The weighted accuracy on the training set is computed as
\begin{equation}
  WA(\mathcal{D}_{\mathrm{train}}) =
  \frac{
    \sum_{i \in \mathcal{D}_{\mathrm{train}}}
    u_i \,
    \mathbb{I}
    \left[
      \frac{|\hat{y}_i - y_i|}{\max(|y_i|, \epsilon_y)}
      \leq \delta
    \right]
  }{
    \sum_{i \in \mathcal{D}_{\mathrm{train}}} u_i
  }.
  \label{eq:weighted_accuracy}
\end{equation}
Thus, $WA$ is used as the evolutionary fitness, whereas the final comparison in Table~\ref{table:comparison} uses the standard accuracy defined in Eq.~\eqref{eq:accuracy}.

The knowledge base used by \textit{EvoOR-Agent} is constructed before evolutionary training. To prevent benchmark-specific information from entering the evolutionary process, the construction and use of the knowledge base follow the leakage-control protocol described in Section~S-III of the supplementary material.

The core evolutionary hyperparameters are summarized in Table~\ref{tab:evolution_parameters}. DeepSeek-v3.2 is used as the default evolutionary operator in parameter and dynamics analyses due to its favorable reasoning-to-cost ratio. In the comparative experiments, \textit{EvoOR-Agent} is instantiated with each foundation model under the same evaluation protocol.

\begin{table}[htbp]
  \centering
  \caption{Core hyperparameter settings}
  \label{tab:evolution_parameters}
  \footnotesize
  \setlength{\tabcolsep}{3pt}
  \renewcommand{\arraystretch}{1.08}
  \begin{tabular*}{\columnwidth}{@{\extracolsep{\fill}} p{0.30\columnwidth} c p{0.34\columnwidth} c @{}}
    \toprule
    \textbf{Parameter} & \textbf{Value} & \textbf{Parameter} & \textbf{Value} \\
    \midrule
    Population size ($N$) & 10 & Iteration depth ($T$) & 8 \\
    Init. ratio ($\alpha_{\mathrm{init}}$) & 50\% & Mutation ratio ($\beta_{\mathrm{mut}}$) & 50\% \\
    Guidance rate ($\beta_{\mathrm{learn}}$) & 50\% & Scope rate ($\beta_{\mathrm{strat}}$) & 50\% \\
    Architecture rate ($\alpha$) & 0.5 & Exploration parameter ($\gamma$) & 0.5 \\
    Pruning threshold ($\tau$) & 0.1 & Elite rate ($\beta_{\mathrm{elite}}$) & 20\% \\
    \bottomrule
  \end{tabular*}
\end{table}

After evolution, the individual with the highest training $WA$ is selected as the final evolved agent for test evaluation. All reproduced methods and \textit{EvoOR-Agent} variants are evaluated over ten independent runs, and we report the mean accuracy and standard deviation.

\subsection{Comparative Results}

We compare \textit{EvoOR-Agent} with five categories of baseline methods. The first group consists of zero-shot reasoning LLMs, where each model directly generates executable solver code from the natural-language problem description. The second group includes \textit{EvoPrompt}~\cite{guo2023evoprompt}, representing evolutionary prompt optimization. The third group is \textit{OR-LLM-Agent}~\cite{Or-llm-agent}, which features a fixed manually designed OR-agent pipeline. The fourth group is \textit{EvoAgent}~\cite{yuan2025evoagent}, which evolves agent personas and task-specific configurations. The last group consists of specialized fine-tuned operations research language models, including \textit{ORLM}~\cite{huang2025orlm} and \textit{StepORLM}~\cite{zhou2025steporlm}.

Table~\ref{table:comparison} reports the comparative performance results across the seven benchmarks. For \textit{EvoOR-Agent} and all reproduced baselines, the evaluation metrics are reported as the mean accuracy and standard deviation over ten independent runs. In the table, bold values indicate the best performance within each method group, a background indicates the best overall result among all compared methods, and underlining indicates the second-highest accuracy in the corresponding column.

\begin{table*}[t]
  \centering
  \caption{Comparison of accuracy performance on NL4OPT, MAMO (ComplexLP/EasyLP), IndustryOR, BWOR, NLP4LP, and ReSocratic benchmarks. Results report the mean and standard deviation ($\pm$ SD) over ten independent runs where available. Tokens denote the average token consumption per problem.}
  \label{table:comparison}
  \scriptsize
  \setlength{\tabcolsep}{3pt}
  \renewcommand{\arraystretch}{1.1}
  \resizebox{\textwidth}{!}{%
    \begin{tabular}{l c c c c c c c c}
      \toprule
      \textbf{Model} & \textbf{IndustryOR} & \textbf{ComplexLP} & \textbf{EasyLP} & \textbf{NL4OPT} & \textbf{BWOR} & \textbf{NLP4LP} & \textbf{ReSocratic} & \textbf{Tokens} \\
      \midrule
      \rowcolor{gray!12}\multicolumn{9}{c}{\textit{Reasoning LLMs (Zero-shot)}} \\
      DeepSeek-v3.2 & 48.72 $\pm$ 1.98\% & \textbf{63.75 $\pm$ 1.24\%} & \textbf{78.91 $\pm$ 0.71\%} & \textbf{75.86 $\pm$ 1.16\%} & 45.12 $\pm$ 1.03\% & 76.78 $\pm$ 1.63\% & 75.92 $\pm$ 0.61\% & 1586 \\
      GPT-5 & \textbf{50.48 $\pm$ 2.09\%} & 55.05 $\pm$ 1.62\% & 76.99 $\pm$ 0.83\% & 71.02 $\pm$ 1.78\% & 57.32 $\pm$ 1.33\% & \textbf{79.46 $\pm$ 1.22\%} & \textbf{78.06 $\pm$ 0.79\%} & 1714 \\
      Gemini 3 Flash & 46.71 $\pm$ 1.78\% & 54.80 $\pm$ 1.36\% & 76.23 $\pm$ 0.91\% & 71.84 $\pm$ 1.27\% & \textbf{58.54 $\pm$ 1.00\%} & 74.87 $\pm$ 1.27\% & 73.57 $\pm$ 0.95\% & 1693 \\
      Qwen 3Max & 36.38 $\pm$ 1.85\% & 51.17 $\pm$ 1.26\% & 73.62 $\pm$ 0.97\% & 64.08 $\pm$ 1.64\% & 40.24 $\pm$ 1.18\% & 69.85 $\pm$ 1.14\% & 68.29 $\pm$ 0.99\% & 1651 \\
      \midrule
      \rowcolor{gray!12}\multicolumn{9}{c}{\textit{EvoPrompt}~\cite{guo2023evoprompt} \textit{(Reproduction)}} \\
      DeepSeek-v3.2 & 64.91 $\pm$ 1.97\% & 68.28 $\pm$ 1.55\% & \textbf{83.76 $\pm$ 0.90\%} & 80.62 $\pm$ 1.19\% & 66.83 $\pm$ 1.36\% & \textbf{84.92 $\pm$ 1.62\%} & 82.95 $\pm$ 0.66\% & 3398 \\
      GPT-5 & \textbf{67.82 $\pm$ 1.89\%} & \textbf{69.72 $\pm$ 1.18\%} & 81.55 $\pm$ 0.98\% & \textbf{82.82 $\pm$ 1.65\%} & \textbf{68.73 $\pm$ 1.26\%} & 82.45 $\pm$ 1.61\% & \textbf{83.86 $\pm$ 0.74\%} & 3512 \\
      Gemini 3 Flash & 57.89 $\pm$ 1.98\% & 64.39 $\pm$ 1.57\% & 78.33 $\pm$ 0.83\% & 79.62 $\pm$ 1.66\% & 68.19 $\pm$ 0.97\% & 80.08 $\pm$ 1.37\% & 78.98 $\pm$ 0.74\% & 3472 \\
      Qwen 3Max & 45.21 $\pm$ 2.15\% & 53.79 $\pm$ 1.16\% & 75.43 $\pm$ 0.62\% & 76.24 $\pm$ 1.40\% & 59.02 $\pm$ 1.37\% & 75.67 $\pm$ 1.40\% & 75.67 $\pm$ 0.80\% & 3394 \\
      \midrule
      \rowcolor{gray!12}\multicolumn{9}{c}{\textit{OR-LLM-Agent}~\cite{Or-llm-agent} \textit{(Reproduction)}} \\
      DeepSeek-v3.2 & 69.04 $\pm$ 1.69\% & 75.67 $\pm$ 1.48\% & \textbf{86.45 $\pm$ 0.89\%} & 86.38 $\pm$ 1.66\% & \textbf{76.83 $\pm$ 0.91\%} & 88.21 $\pm$ 1.74\% & 87.44 $\pm$ 0.66\% & 3903 \\
      GPT-5 & \textbf{71.51 $\pm$ 1.95\%} & 73.29 $\pm$ 1.67\% & 84.55 $\pm$ 0.71\% & \textbf{87.92 $\pm$ 1.80\%} & 70.73 $\pm$ 1.13\% & 89.61 $\pm$ 1.64\% & \textbf{88.41 $\pm$ 0.79\%} & 3987 \\
      Gemini 3 Flash & 69.87 $\pm$ 2.08\% & \textbf{75.93 $\pm$ 1.44\%} & 80.96 $\pm$ 0.64\% & 85.21 $\pm$ 1.17\% & 73.17 $\pm$ 1.14\% & \textbf{89.78 $\pm$ 1.67\%} & 85.98 $\pm$ 0.72\% & 3821 \\
      Qwen 3Max & 50.21 $\pm$ 1.83\% & 68.03 $\pm$ 1.11\% & 77.15 $\pm$ 0.73\% & 78.16 $\pm$ 1.54\% & 67.07 $\pm$ 1.08\% & 80.63 $\pm$ 1.73\% & 78.93 $\pm$ 0.72\% & 4096 \\
      \midrule
      \rowcolor{gray!12}\multicolumn{9}{c}{\textit{EvoAgent}~\cite{yuan2025evoagent} \textit{(Reproduction)}} \\
      DeepSeek-v3.2 & \textbf{72.82 $\pm$ 1.91\%} & 69.81 $\pm$ 1.37\% & \textbf{83.34 $\pm$ 0.69\%} & \textbf{85.89 $\pm$ 1.16\%} & 74.95 $\pm$ 1.18\% & 85.62 $\pm$ 1.68\% & \textbf{86.79 $\pm$ 0.62\%} & 4874 \\
      GPT-5 & 69.76 $\pm$ 1.79\% & \textbf{70.57 $\pm$ 1.76\%} & 80.87 $\pm$ 0.90\% & 83.21 $\pm$ 1.21\% & \textbf{75.61 $\pm$ 0.96\%} & \textbf{87.01 $\pm$ 1.14\%} & 85.76 $\pm$ 0.99\% & 4673 \\
      Gemini 3 Flash & 70.43 $\pm$ 1.90\% & 70.31 $\pm$ 1.79\% & 83.26 $\pm$ 0.79\% & 82.36 $\pm$ 1.58\% & 73.78 $\pm$ 1.04\% & 83.27 $\pm$ 1.20\% & 84.37 $\pm$ 0.68\% & 4788 \\
      Qwen 3Max & 62.58 $\pm$ 2.19\% & 62.08 $\pm$ 1.65\% & 78.12 $\pm$ 0.80\% & 77.26 $\pm$ 1.19\% & 65.79 $\pm$ 1.26\% & 79.53 $\pm$ 1.48\% & 78.53 $\pm$ 0.66\% & 5023 \\
      \midrule
      \rowcolor{gray!12}\multicolumn{9}{c}{\textit{Fine-tuned LLMs (Reproduction)}} \\
      ORLM~\cite{huang2025orlm} & 48.56 $\pm$ 1.34\% & 60.77 $\pm$ 1.21 \% & 85.43 $\pm$ 0.84\% & 72.35 $\pm$ 0.75 \% & 29.27 $\pm$ 1.87\% & 75.32 $\pm$ 1.36\% & 65.92 $\pm$ 1.01\% & 1365 \\
      StepORLM & 63.88 $\pm$ 1.17\% & \underline{80.18$\pm$ 1.04\%} & \cellcolor{bestcell}\textbf{97.06 $\pm$ 0.76\%} & \underline{93.89 $\pm$ 0.81\%} & 42.38 $\pm$ 0.89\% & \underline{96.06 $\pm$ 1.02\%} & 84.81 $\pm$ 0.86\% & 2187 \\
      \midrule
      \rowcolor{gray!12}\multicolumn{9}{c}{\textit{EvoOR-Agent (Ours)}} \\
      DeepSeek-v3.2 & \cellcolor{bestcell}\textbf{81.55 $\pm$ 1.73\%} & \cellcolor{bestcell}\textbf{81.98 $\pm$ 1.14\%} & 93.51 $\pm$ 0.65\% & \cellcolor{bestcell}\textbf{94.03 $\pm$ 1.71\%} & \cellcolor{bestcell}\textbf{84.15 $\pm$ 1.08\%} & 93.28 $\pm$ 1.59 \% & \cellcolor{bestcell}\textbf{92.55 $\pm$ 0.74\%} & 5023 \\
      GPT-5 & 75.88 $\pm$ 1.84\% & 75.53 $\pm$ 1.48\% & \underline{\textbf{94.19 $\pm$ 0.74\%}} & 90.76 $\pm$ 1.20\% & \underline{80.49 $\pm$ 1.20\%} & \cellcolor{bestcell}\textbf{97.75 $\pm$ 1.35\%} & \underline{90.17 $\pm$ 0.93\%} & 5129 \\
      Gemini 3 Flash & \underline{77.12 $\pm$ 1.96\%} & 74.89 $\pm$ 1.36\% & 90.82 $\pm$ 0.63\% & 92.59 $\pm$ 1.14\% & 79.88 $\pm$ 1.16\% & 92.23 $\pm$ 1.23\% & 88.72 $\pm$ 0.63\% & 5217 \\
      Qwen 3Max & 65.76 $\pm$ 2.14\% & 68.82 $\pm$ 1.75\% & 85.76 $\pm$ 0.77\% & 86.72 $\pm$ 1.20\% & 73.17 $\pm$ 1.12\% & 85.10 $\pm$ 1.79\% & 83.07 $\pm$ 0.80\% & 5334 \\
      \bottomrule
    \end{tabular}%
  }
  \vspace{2pt}
  \footnotesize
\end{table*}

Compared with \textit{OR-LLM-Agent}, which relies on a fixed manually designed pipeline, \textit{EvoOR-Agent} demonstrates clear advantages by making the workflow topology and reasoning trajectory evolvable. Whereas \textit{OR-LLM-Agent} applies a static sequential solving process to all problems, our framework dynamically adapts its architecture graph to the structural constraints of each OR instance. As shown in Table~\ref{table:comparison}, \textit{EvoOR-Agent} (with DeepSeek-v3.2) outperforms this fixed baseline on all seven evaluation subsets, achieving particularly large margins on more challenging benchmarks: ComplexLP (81.98\% vs. 75.67\%), IndustryOR (81.55\% vs. 71.51\%), and BWOR (84.15\% vs. 76.83\%). These results confirm that an evolvable workflow trajectory yields substantial gains over a static, expert-designed agent architecture.

When compared to general evolutionary frameworks such as \textit{EvoPrompt} and \textit{EvoAgent}, the performance improvements of \textit{EvoOR-Agent} highlight the importance of selecting an appropriate search object. \textit{EvoPrompt} optimizes surface-level prompt instructions, and \textit{EvoAgent} evolves agent personas or task configurations, but neither alters the underlying operational workflow. In contrast, \textit{EvoOR-Agent} performs a structural search over the architecture graph to directly optimize the mathematical formulation and execution trajectories. By shifting the evolutionary focus from generic linguistic patterns to task-aligned structural patterns, our method consistently outperforms the best evolutionary baseline results across all benchmarks, with particularly clear gains on IndustryOR (e.g., 81.55\% vs. 72.82\%) and BWOR (84.15\% vs. 76.83\%).

The deep comparison between our framework and specialized fine-tuned models (\textit{ORLM} and \textit{StepORLM}) reveals a fundamental distinction in optimization paradigms. Parametric fine-tuning is inherently data-driven, modifying model weights to store domain-specific patterns, which tightly couples the learned capability with the training distribution. Conversely, \textit{EvoOR-Agent} follows a reasoning-driven paradigm that keeps the underlying LLM weights completely frozen. Instead of adapting the parametric model to data patterns through weight updates, our framework keeps the LLM fixed and optimizes the AOE-style agent architecture graph together with the reasoning trajectories instantiated on it. This is fundamentally different from fine-tuning: we change the agent's way of thinking, not the model's weights.

This paradigm difference becomes particularly evident when analyzing out-of-distribution generalization. As reported in Table~\ref{table:comparison}, the data-driven fine-tuning approach of \textit{StepORLM} achieves strong results on benchmarks aligned with its training distribution, such as EasyLP (97.06\%) and NL4OPT (93.89\%). However, its performance degrades sharply on the BWOR benchmark, falling to 42.38\%, indicating severe overfitting to simpler textual structures. In contrast, \textit{EvoOR-Agent}  maintains high robustness across heterogeneous tasks, securing a substantial lead on BWOR (84.15\%) and also delivering strong results on NLP4LP (97.75\% ) and ReSocratic (92.55\%), which are entirely excluded from the evolutionary training phase. This demonstrates that structural evolution provides more reliable transferability than weight modification when facing unseen problem layouts.

Regarding computational overhead, the final column of Table~\ref{table:comparison} shows that while \textit{EvoOR-Agent} consumes more tokens than single-pass zero-shot or fine-tuned models, this expenditure yields consistent and substantial accuracy dividends. Across all iterative and evolutionary baselines, our framework maintains a stable token consumption interval (between 5,023 and 5,334 tokens). This cost is practical because \textit{EvoOR-Agent} effectively translates the allocated token budget into reliable iterative reasoning and self-reflection, rather than mere code execution. The stable accuracy improvements across diverse tasks justify the token budget as a scalable mechanism for complex OR problem solving.

In summary, the comparative results validate the effectiveness of structural agent evolution for operations research tasks. \textit{EvoOR-Agent} achieves state-of-the-art performance on six out of the seven benchmarks: with DeepSeek-v3.2, it tops IndustryOR (81.55\%), ComplexLP (81.98\%), NL4OPT (94.03\%), BWOR (84.15\%), and ReSocratic (92.55\%); with GPT-5, it achieves the highest accuracy on NLP4LP (97.75\%). This strikes a favorable balance between optimization accuracy and computational efficiency, confirming that structural evolution can outperform both static agent designs and parametric fine-tuning.

\section{Discussion}
\label{sec:Discussion}

In this section, we analyze EvoOR-Agent from three perspectives. First, the case study examines the structure of an evolved reasoning trajectory and compares it with the static OR-LLM-Agent path. Second, the ablation study evaluates the contribution of the main algorithmic components, including chain-of-thought prompting, evolutionary search, the AOE-style architecture graph, and knowledge-base priors. Third, the population size, convergence, and population dynamics analyses investigate how the evolutionary process affects search efficiency, generalization, and structural diversity.

\subsection{Case Study}
\label{subsec:case_study}

We first examine the reasoning trajectory learned by EvoOR-Agent. To make the evolved workflow interpretable, we extract the state-action trace of the best individual selected from Generation 8 and map it onto the architecture graph. Table~\ref{tab:case_study_trajectory} reports the resulting trajectory. The shaded rows indicate the edges traversed by the static OR-LLM-Agent when it is mapped onto the same architecture space.

\begin{table}[h]
  \centering
  \caption{Evolved reasoning trajectory and mapped OR-LLM-Agent path}
  \label{tab:case_study_trajectory}
  \scriptsize
  \renewcommand{\arraystretch}{1.12}
  \setlength{\tabcolsep}{2pt}
  \begin{tabular*}{\columnwidth}{@{\extracolsep{\fill}} >{\centering\arraybackslash}p{0.06\columnwidth} >{\raggedright\arraybackslash}p{0.17\columnwidth} >{\raggedright\arraybackslash}p{0.17\columnwidth} >{\raggedright\arraybackslash}p{0.52\columnwidth} @{}}
    \toprule
    \textbf{ID} & \textbf{Start State} & \textbf{End State} & \multicolumn{1}{c}{\textbf{Key Action / Prompt / Tool}} \\
    \midrule
    \rowcolor{baselinepath}[0pt][0pt] 1 & Agent Start & Log Init & \texttt{log\_llm\_chat} \\
    2 & Log Init & Config Ready & Initialize logging module and dialogue \\
    \rowcolor{baselinepath}[0pt][0pt] 3 & Config Ready & Path Extracted & Load target dataset path from configuration \\
    \rowcolor{baselinepath}[0pt][0pt] 4 & Path Extracted & Ques Loaded & \texttt{load\_dataset} \\
    \rowcolor{baselinepath}[0pt][0pt] 5 & Ques Loaded & Ques Format & Format raw problem into reasoning template \\
    6 & Ques Format & Sets Defined & \textit{``...analyze the problem and explicitly define Sets $S$...''} \\
    7 & Sets Defined & Para Defined & \textit{``...extract and list all Parameters $P$ from the text...''} \\
    8 & Para Defined & Vars Defined & \textit{``...define Decision Variables $V$ with their bounds...''} \\
    9 & Vars Defined & Obj Defined & \textit{``...formulate the Objective Function...''} \\
    10 & Obj Defined & All Defined & \textit{``...construct logical Constraints mathematically...''} \\
    \rowcolor{baselinepath}[0pt][0pt] 11 & All Defined & Raw S1 Out & \texttt{query\_llm} \\
    \rowcolor{baselinepath}[0pt][0pt] 12 & Raw S1 Out & Txt Ready & Parse generated mathematical model \\
    \midrule
    13 & Txt Ready & Props Parsed & \textit{``...analyze the model's linearity and MILP scale properties...''} \\
    14 & Props Parsed & Limits Eva & \textit{``...evaluate if a Gurobi exact solve is computationally viable...''} \\
    15 & Limits Eva & Route Dec & \textit{``...decide algorithmic path: exact solver vs. designing new algorithms...''} \\
    16 & Route Dec & Raw Route & \texttt{query\_llm} \\
    17 & Raw Route & Route & Extract algorithmic routing strategy \\
    18 & Route & Algo Struct & \textit{``...design Gurobi implementation strategy based on routing...''} \\
    19 & Algo Struct & Strat Verified & \textit{``...verify logical consistency of the algorithm design...''} \\
    20 & Strat Verified & Raw S2 Out & \texttt{query\_llm} \\
    21 & Raw S2 Out & Txt Ready & Parse and isolate Gurobi algorithm design \\
    \midrule
    \rowcolor{baselinepath}[0pt][0pt] 22 & Txt Ready & Final Txt & Merge model and algorithm context \\
    \rowcolor{baselinepath}[0pt][0pt] 23 & Final Txt & Generate & \textit{``...generate complete executable Python code using Gurobi...''} \\
    \rowcolor{baselinepath}[0pt][0pt] 24 & Generate & Raw Code & \texttt{query\_llm} \\
    \rowcolor{baselinepath}[0pt][0pt] 25 & Raw Code & Code Parsed & Extract executable Python code block \\
    \rowcolor{baselinepath}[0pt][0pt] 26 & Code Parsed & Code Saved & \texttt{save\_generated\_code} \\
    \rowcolor{baselinepath}[0pt][0pt] 27 & Code Saved & Exec Output & \texttt{extract\_and\_execute\_python\_code} \\
    28 & Exec Output & Obj Extracted & \texttt{extract\_best\_objective} \\
    29 & Obj Extracted & Type Check & Validate if extracted objective is numeric \\
    30 & Type Check & Status Check & Check solver execution \\
    31 & Status Check & Refl. Synth'd & \textit{``...model yielded NO solution. Backtrack and check Stage 1...''} \\
    32 & Refl. Synth'd & Corrected Res & \texttt{query\_llm} \\
    33 & Corrected Res & Code Updated & Extract and update the corrected code \\
    \rowcolor{baselinepath}[0pt][0pt] 34 & Code Updated & Process Term & Set final solve success flag based on debug iterations \\
    \rowcolor{baselinepath}[0pt][0pt] 35 & Process Term & Bench. Done & [External] \texttt{eval\_model\_result} \\
    \bottomrule
  \end{tabular*}
  \vspace{2pt}
  \footnotesize

  \noindent\textit{Note:} Shaded rows indicate the edges traversed by OR-LLM-Agent after being mapped onto the evolved AOE space. The external evaluator in the last row is used only for offline scoring after solution generation and is not available to the LLM during reasoning or debugging.
\end{table}

The 35-edge trajectory in Table~\ref{tab:case_study_trajectory} reveals three structures learned by EvoOR-Agent. The first structure is formulation decomposition. Edges 6 to 10 split the modeling stage into a sequence of smaller steps, including the definitions of sets, parameters, decision variables, objective function, and constraints. This design avoids generating the full mathematical model in one monolithic step and makes intermediate modeling states explicit. Such decomposition is useful for OR tasks because errors in early modeling elements can propagate directly to solver code.

The second structure is algorithmic routing before code generation. Edges 13 to 17 analyze model properties, including linearity, scale, and solver feasibility, before selecting an implementation strategy. This routing stage separates formulation from solver choice. Compared with directly translating the mathematical model into code, it introduces an intermediate decision point that can adapt the downstream implementation to the problem structure.

The third structure is solver-status-based debugging. Edges 30 to 33 check the execution status and trigger revision when the solver returns an infeasible, unbounded, or invalid result. The debugging process uses only internal solver feedback and generated execution outputs. It does not use benchmark answers or external scoring information. Therefore, the correction step remains isolated from the final evaluator.

The shaded rows in Table~\ref{tab:case_study_trajectory} show the path traversed by OR-LLM-Agent after it is mapped onto the evolved AOE space. Under this shared representation, the static baseline mainly follows a coarse sequence of problem loading, modeling, code generation, execution, and final evaluation. In contrast, EvoOR-Agent inserts additional states for formulation decomposition, property analysis, routing, verification, and solver-status-based debugging. This comparison illustrates how architecture evolution expands a fixed OR-agent pipeline into a more fine-grained and interpretable reasoning trajectory.

Overall, the case study provides a qualitative explanation for the empirical results in Section~\ref{sec:Experiments}. The evolved trajectory exposes intermediate reasoning states that can be recombined, mutated, and pruned during evolution. This provides a structural basis for adapting OR-solving workflows, rather than relying only on a fixed pipeline or a single end-to-end generation step.

\subsection{Ablation Study}
\label{subsec:ablation}

We conduct an ablation study using DeepSeek-v3.2 to isolate the contribution of each major component in \textit{EvoOR-Agent}. For variants involving evolutionary training, the inference engine, population size, iteration depth, and evaluation protocol are kept consistent with Section~\ref{sec:Experiments}. Table~\ref{table:ablation} reports mean accuracy over ten independent runs.

We evaluate five configurations across seven datasets. The base LLM directly generates executable solver code from the problem statement. The +CoT variant adds explicit intermediate reasoning but still uses a fixed generation process. The +EC variant introduces evolutionary search through direct text-level mutation. The +AOE variant replaces unconstrained text mutation with trajectory-level evolution over the AOE-style architecture graph, but does not use knowledge-base priors. The full \textit{EvoOR-Agent} further incorporates the knowledge base into initialization and semantic variation.

\begin{table}[h]
  \centering
  \caption{Ablation study on the core components of EvoOR-Agent using DeepSeek-v3.2. Results are reported as mean accuracy over ten independent runs.}
  \label{table:ablation}
  \scriptsize
  \setlength{\tabcolsep}{3pt}
  \renewcommand{\arraystretch}{1.08}
  \resizebox{\columnwidth}{!}{%
    \begin{tabular}{l c c c c c c c c}
      \toprule
      \textbf{Variant} & \textbf{IndustryOR} & \textbf{ComplexLP} & \textbf{EasyLP} & \textbf{NL4OPT} & \textbf{BWOR} & \textbf{WA} & \textbf{NLP4LP} & \textbf{ReSocratic} \\
      \midrule
      LLM & 48.72\% & 63.75\% & 78.91\% & 75.86\% & 45.12\% & 62.9\% & 76.78\% & 75.92\% \\
      +CoT & 54.20\% & 66.18\% & 81.36\% & 79.24\% & 53.68\% & 67.4\% & 80.92\% & 79.84\% \\
      +EC & 64.85\% & 70.46\% & 84.28\% & 84.12\% & 67.34\% & 75.8\% & 86.38\% & 84.96\% \\
      +AOE & 76.10\% & 78.62\% & 91.08\% & 91.36\% & 80.72\% & 84.1\% & 94.62\% & 89.70\% \\
      \midrule
      \textbf{Full} & \textbf{81.55\%} & \textbf{81.98\%} & \textbf{93.51\%} & \textbf{94.03\%} & \textbf{84.15\%} & \textbf{87.0\%} & \textbf{93.28\%} & \textbf{92.55\%} \\
      \bottomrule
    \end{tabular}%
  }
\end{table}

Table~\ref{table:ablation} shows a clear stepwise improvement as components are added. The base LLM obtains a weighted accuracy (WA) of 62.9\% over the five evolutionary benchmarks. Adding CoT raises WA to 67.4\%, with visible gains on IndustryOR, NL4OPT, and BWOR. This indicates that explicit intermediate reasoning is useful, but a fixed reasoning template alone is insufficient for adapting the solving workflow to heterogeneous OR instances.

Introducing evolutionary search through +EC further increases WA to 75.8\%. The largest gains appear on IndustryOR and BWOR, where accuracy improves from 54.20\% to 64.85\% and from 53.68\% to 67.34\%, respectively, relative to +CoT. These improvements show that evolutionary variation can discover more effective reasoning patterns than a single static prompt. However, because +EC mutates the text representation directly, it does not explicitly preserve workflow topology or phase-level execution dependencies.

The +AOE variant provides the largest component-level gain, increasing WA from 75.8\% to 84.1\%. Compared with +EC, it improves IndustryOR from 64.85\% to 76.10\%, ComplexLP from 70.46\% to 78.62\%, EasyLP from 84.28\% to 91.08\%, NL4OPT from 84.12\% to 91.36\%, and BWOR from 67.34\% to 80.72\%. This confirms the central role of the AOE-style architecture graph: by evolving reasoning trajectories in a structured graph space, the method can preserve useful intermediate states, localize structural changes, and reuse effective formulation--execution paths.

The full \textit{EvoOR-Agent} achieves the best overall WA of 87.0\% after adding knowledge-base priors. It further improves IndustryOR, ComplexLP, EasyLP, NL4OPT, BWOR, and ReSocratic over +AOE, while NLP4LP decreases slightly from 94.62\% to 93.28\%. This pattern suggests that the knowledge base mainly acts as a refinement mechanism on top of architecture evolution, improving initialization and semantic mutation for most heterogeneous OR tasks without replacing the dominant contribution of the AOE representation. Overall, the ablation results indicate that workflow-level structural evolution is the main driver of performance, and knowledge-guided variation provides an additional but smaller gain.

\subsection{Population Size Analysis}
\label{subsec:population_size}

\begin{figure}[h]
  \centering
  \includegraphics[width=0.95\columnwidth]{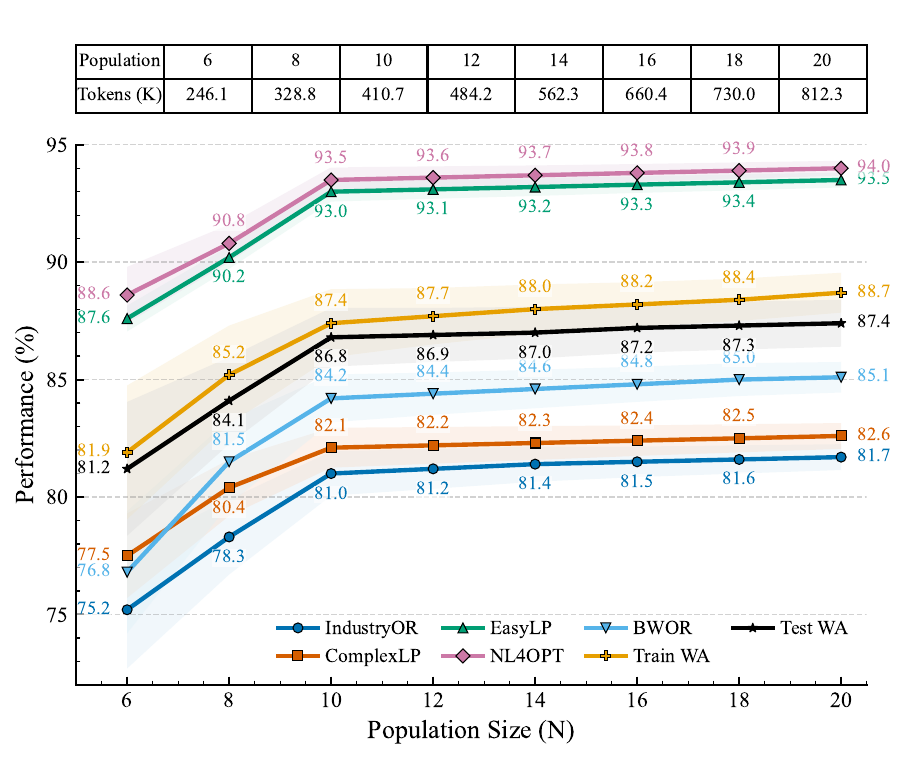}
  \caption{Population-size sensitivity at a fixed iteration depth of $T=8$. The curves show Train WA, Test WA, subset accuracies, and the cumulative token budget as the population size increases. Tokens include population initialization and all eight evolutionary updates; shaded regions denote standard deviations over ten runs. DeepSeek-v3.2 is used as the evolutionary operator.}
  \label{fig:population_size_analysis}
\end{figure}

We analyze the sensitivity of \textit{EvoOR-Agent} to the population size $N$ while fixing the iteration depth at $T=8$. The population size is varied from 6 to 20 in increments of two. Fig.~\ref{fig:population_size_analysis} reports the five evolutionary benchmark accuracies, Train WA, Test WA, and the corresponding cumulative token budget. This experiment examines the trade-off between population diversity, search stability, and computational cost.

Fig.~\ref{fig:population_size_analysis} shows that increasing $N$ from 6 to 10 produces the main performance gain. Test WA rises from 81.2\% to 86.8\%, and Train WA rises from 81.9\% to 87.4\%. The subset curves follow the same pattern: IndustryOR improves from 75.2\% to 81.0\%, ComplexLP from 77.5\% to 82.1\%, EasyLP from 87.6\% to 93.0\%, NL4OPT from 88.6\% to 93.5\%, and BWOR from 76.8\% to 84.2\%. These gains indicate that a small population does not provide enough architectural diversity for reliable trajectory evolution, especially on heterogeneous or less standardized benchmarks such as IndustryOR and BWOR.

After $N=10$, the curves enter a plateau. From $N=10$ to $N=20$, Test WA increases only from 86.8\% to 87.4\%, while most subset accuracies change by less than one percentage point. In contrast, the token budget continues to grow almost linearly, increasing from 410.7K tokens at $N=10$ to 812.3K tokens at $N=20$. This indicates that larger populations mainly add redundant exploration once the architecture graph already contains enough useful reasoning states and transitions.

The shaded regions also narrow as the population increases from 6 to 10, suggesting that moderate population diversity stabilizes the evolutionary process. Beyond $N=10$, however, the reduction in variance is limited compared with the additional token cost. Therefore, $N=10$ provides the best balance in the present setting: it supplies enough diverse individuals for path-conditioned recombination and semantic mutation, while avoiding the rapidly increasing cost of maintaining larger populations.

\subsection{Convergence Analysis}
\label{subsec:convergence}

\begin{figure}[h]
  \centering
  \includegraphics[width=0.95\columnwidth]{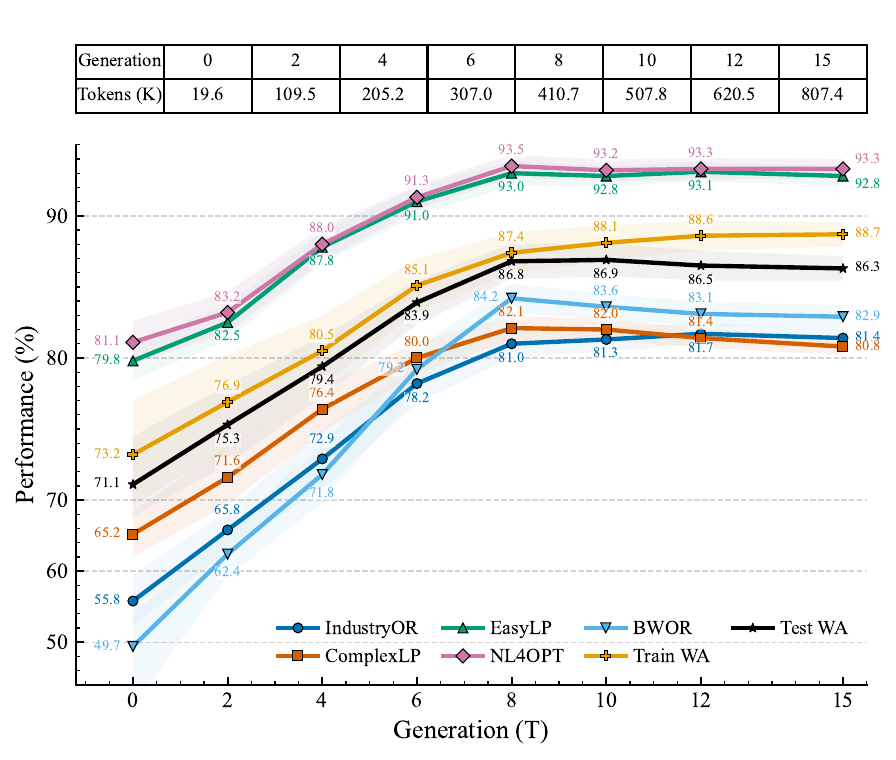}
  \caption{Convergence behavior with a fixed population size of $N=10$. The curves show Train WA, Test WA, subset accuracies, and the actual cumulative token budget as the generation number increases. Tokens include population initialization and all evolutionary updates completed up to generation $T$; shaded regions denote standard deviations over ten runs. DeepSeek-v3.2 is used as the evolutionary operator.}
  \label{fig:convergence_analysis}
\end{figure}

We analyze the convergence behavior of \textit{EvoOR-Agent} over 15 generations with a fixed population size of $N=10$. Fig.~\ref{fig:convergence_analysis} reports the five evolutionary benchmark accuracies, Train WA, Test WA, and the actual cumulative token budget recorded during the evolutionary process. Tokens include population initialization and all evolutionary updates completed up to generation $T$; for example, the value at $T=3$ includes initialization and the first three update rounds. This experiment examines the trade-off between iterative refinement, generalization stability, and computational cost.

Fig.~\ref{fig:convergence_analysis} shows that increasing the generation number from 0 to 8 produces the main performance gain. Test WA rises from 71.1\% to 86.8\%, and Train WA rises from 73.2\% to 87.4\%. The subset curves follow the same pattern: IndustryOR improves from 55.8\% to 81.0\%, ComplexLP from 65.2\% to 82.1\%, EasyLP from 79.8\% to 93.0\%, NL4OPT from 81.1\% to 93.5\%, and BWOR from 49.7\% to 84.2\%. These gains indicate that early and middle generations effectively accumulate useful reasoning states and transitions in the architecture graph.

The largest improvements occur on the more difficult and less standardized subsets. BWOR rises by 34.5 percentage points from Generation 0 to Generation 8, and IndustryOR rises by 25.2 percentage points. This suggests that trajectory evolution is especially valuable when the problem requires iterative formulation, solver routing, and execution repair rather than direct translation from text to code. By contrast, EasyLP and NL4OPT start from higher initial accuracy and therefore show smaller but still consistent gains.

After Generation 8, the testing curves enter a plateau and then slightly decline. From Generation 8 to Generation 15, Test WA decreases from 86.8\% to 86.3\%, while Train WA continues to increase from 87.4\% to 88.7\%. Similar mild declines appear on IndustryOR, ComplexLP, BWOR, and NL4OPT. This divergence indicates the onset of overfitting: later generations increasingly adapt to the training split, while their benefit to unseen test instances becomes weaker.

In contrast, the actual cumulative token budget continues to grow substantially after the best test point, increasing from 410.7K tokens at Generation 8 to 807.4K tokens at Generation 15. This indicates that additional generations mainly add redundant or overly specialized exploration once the architecture graph already contains effective reasoning trajectories. Therefore, Generation 8 provides the best balance in the present setting: it reaches the highest Test WA while avoiding the rapidly increasing token cost of later evolutionary updates.

\subsection{Population Dynamics Analysis}
\label{subsec:population_dynamics}

\begin{figure}[h]
  \centering
  \includegraphics[width=0.95\columnwidth]{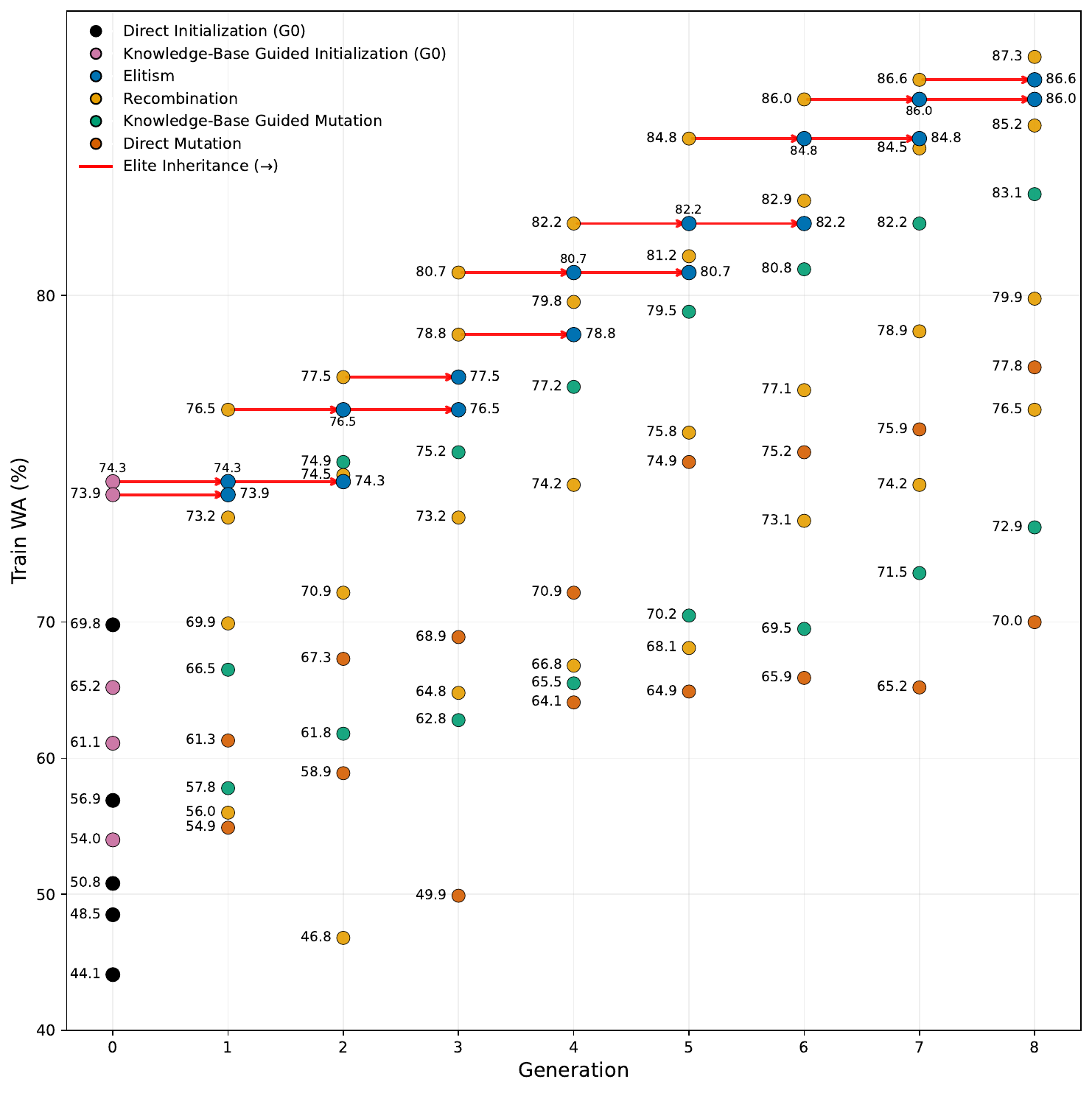}
  \caption{Population dynamics across generations. The vertical axis denotes training weighted accuracy. Colors indicate the applied evolutionary operators. Red arrows denote elite inheritance. A piecewise scaling is applied to the vertical axis to improve visibility.}
  \label{fig:population_dynamics}
\end{figure}

We further inspect the population dynamics during the default $T=8$ evolutionary process. Fig.~\ref{fig:population_dynamics} tracks the Train WA of all ten population slots across generations and marks the operator that produces each individual. This analysis illustrates how direct initialization, knowledge-base-guided initialization, elitism, recombination, and mutation jointly shape the population.

The initial population contains both weak and competitive individuals, reflecting the diversity introduced by direct and knowledge-base-guided initialization. At Generation 0, Train WA ranges from 44.1\% to 74.3\%, and the best initialized individual is produced by knowledge-base-guided initialization. This wide spread indicates that initialization provides diverse starting trajectories rather than a single homogeneous reasoning pattern.

Across generations, the elite inheritance path provides a stable upper envelope for the population. The best Train WA increases from 74.3\% at Generation 1 to 77.5\% at Generation 2, 80.7\% at Generation 3, 82.2\% at Generation 4, 84.8\% at Generation 5, 86.0\% at Generation 6, 86.6\% at Generation 7, and 87.3\% at Generation 8. This monotonic elite trajectory shows that elitism prevents high-quality reasoning trajectories from being lost while allowing the rest of the population to continue exploring new variants.

Recombination contributes several major improvements in the upper population. For example, recombination generates individuals with Train WA of 77.5\% at Generation 2, 80.7\% at Generation 3, 82.2\% at Generation 4, 84.8\% at Generation 5, 86.0\% at Generation 6, and 87.3\% at Generation 8. These jumps suggest that once useful reasoning fragments have accumulated in the architecture graph, path-conditioned recombination can assemble them into stronger trajectories than direct inheritance alone.

The same dynamics also show the cost and value of exploration. Some recombination and mutation attempts produce low-fitness individuals, such as 46.8\% at Generation 2, 49.9\% at Generation 3, and 64.9\% at Generation 5, indicating that structural or semantic changes can disrupt useful reasoning sequences. Meanwhile, knowledge-base-guided mutation often generates competitive mid- and high-level variants, such as 77.2\% at Generation 4, 80.8\% at Generation 6, 82.2\% at Generation 7, and 83.1\% at Generation 8. Overall, Fig.~\ref{fig:population_dynamics} shows that \textit{EvoOR-Agent} maintains exploration diversity through recombination and mutation, while elite inheritance preserves the best discovered trajectories and steadily raises the population frontier.

\section{Conclusion}
\label{sec:Conclusion}

In this paper, we proposed EvoOR-Agent, a co-evolutionary framework for automated OR problem solving. EvoOR-Agent represents OR-agent workflows as AOE-style architecture graphs and evolves reasoning trajectories over the maintained graph. This design makes the organization of problem interpretation, mathematical formulation, solver selection, code generation, and debugging explicit and evolvable. In addition, knowledge-base priors are incorporated into initialization and semantic mutation to provide reusable OR modeling and implementation patterns. Experiments on heterogeneous OR benchmarks show that EvoOR-Agent improves over zero-shot reasoning LLMs, fixed OR-agent pipelines, and representative evolutionary LLM-agent baselines. The ablation study further shows that the AOE-style architecture representation contributes the largest component-level gain, while the knowledge base provides consistent additional improvement. Case studies and evolutionary analyses also indicate that the evolved trajectories can expose interpretable structures such as formulation decomposition, algorithmic routing, and solver-status-based debugging.

In future work, we will extend EvoOR-Agent in three directions. First, we will study dynamic and stochastic OR settings, where the architecture graph needs to adapt to changing environments and uncertain inputs. Second, we will investigate continual evolution mechanisms so that the framework can update its knowledge base and reasoning trajectories from long-term deployment experience. Third, we will explore cooperative multi-agent extensions, where multiple evolved agents can specialize in different modeling, solving, and verification roles for large-scale or multi-objective optimization tasks.

\bibliographystyle{IEEEtran}
\bibliography{ref}

\begin{thebibliography}{10}
\providecommand{\url}[1]{#1}
\csname url@samestyle\endcsname
\providecommand{\newblock}{\relax}
\providecommand{\bibinfo}[2]{#2}
\providecommand{\BIBentrySTDinterwordspacing}{\spaceskip=0pt\relax}
\providecommand{\BIBentryALTinterwordstretchfactor}{4}
\providecommand{\BIBentryALTinterwordspacing}{\spaceskip=\fontdimen2\font plus
\BIBentryALTinterwordstretchfactor\fontdimen3\font minus \fontdimen4\font\relax}
\providecommand{\BIBforeignlanguage}[2]{{%
\expandafter\ifx\csname l@#1\endcsname\relax
\typeout{** WARNING: IEEEtran.bst: No hyphenation pattern has been}%
\typeout{** loaded for the language `#1'. Using the pattern for}%
\typeout{** the default language instead.}%
\else
\language=\csname l@#1\endcsname
\fi
#2}}
\providecommand{\BIBdecl}{\relax}
\BIBdecl

\bibitem{Practicaloptimization}
P.~E. Gill, W.~Murray, and M.~H. Wright, \emph{Practical optimization}.\hskip 1em plus 0.5em minus 0.4em\relax SIAM, 2019.

\bibitem{procurement}
D.~Saban and G.~Y. Weintraub, ``Procurement mechanisms for assortments of differentiated products,'' \emph{Operations Research}, vol.~69, no.~3, pp. 795--820, 2021.

\bibitem{challengesoptimization}
A.~Abbas, A.~Ambainis, B.~Augustino, A.~B{\"a}rtschi, H.~Buhrman, C.~Coffrin, G.~Cortiana, V.~Dunjko, D.~J. Egger, B.~G. Elmegreen \emph{et~al.}, ``Challenges and opportunities in quantum optimization,'' \emph{Nature Reviews Physics}, vol.~6, no.~12, pp. 718--735, 2024.

\bibitem{ruder2016overview}
S.~Ruder, ``An overview of gradient descent optimization algorithms,'' \emph{arXiv preprint arXiv:1609.04747}, 2016.

\bibitem{burke2013hyper}
E.~K. Burke, M.~Gendreau, M.~Hyde, G.~Kendall, G.~Ochoa, E.~{\"O}zcan, and R.~Qu, ``Hyper-heuristics: A survey of the state of the art,'' \emph{Journal of the Operational Research Society}, vol.~64, no.~12, pp. 1695--1724, 2013.

\bibitem{xu2022difficulty}
P.~Xu, W.~Luo, X.~Lin, Y.~Chang, and K.~Tang, ``Difficulty and contribution-based cooperative coevolution for large-scale optimization,'' \emph{IEEE Transactions on Evolutionary Computation}, vol.~27, no.~5, pp. 1355--1369, 2022.

\bibitem{zhu2026density}
Y.~Zhu, P.~Xu, J.~Huang, X.~Lin, and W.~Luo, ``Density-assisted evolutionary dynamic multimodal optimization,'' \emph{ACM Transactions on Evolutionary Learning and Optimization}, vol.~6, no.~1, pp. 1--30, 2026.

\bibitem{zou2025swarm}
Y.~Zou, P.~Xu, H.~Dai, H.~Song, and W.~Luo, ``Swarm optimization with intra- and inter-hierarchical competition for large-scale berth allocation and crane assignment,'' \emph{IEEE Transactions on Emerging Topics in Computational Intelligence}, vol.~9, no.~2, pp. 1307--1321, Apr. 2025.

\bibitem{kerschke2019automated}
P.~Kerschke, H.~H. Hoos, F.~Neumann, and H.~Trautmann, ``Automated algorithm selection: Survey and perspectives,'' \emph{Evolutionary computation}, vol.~27, no.~1, pp. 3--45, 2019.

\bibitem{huang2019survey}
C.~Huang, Y.~Li, and X.~Yao, ``A survey of automatic parameter tuning methods for metaheuristics,'' \emph{IEEE transactions on evolutionary computation}, vol.~24, no.~2, pp. 201--216, 2019.

\bibitem{ma2025toward}
Z.~Ma, H.~Guo, Y.-J. Gong, J.~Zhang, and K.~C. Tan, ``Toward automated algorithm design: A survey and practical guide to meta-black-box-optimization,'' \emph{IEEE Transactions on Evolutionary Computation}, 2025.

\bibitem{van2024llamea}
N.~Van~Stein and T.~B{\"a}ck, ``Llamea: A large language model evolutionary algorithm for automatically generating metaheuristics,'' \emph{IEEE Transactions on Evolutionary Computation}, vol.~29, no.~2, pp. 331--345, 2024.

\bibitem{COT}
J.~Wei, X.~Wang, D.~Schuurmans, M.~Bosma, F.~Xia, E.~Chi, Q.~V. Le, D.~Zhou \emph{et~al.}, ``Chain-of-thought prompting elicits reasoning in large language models,'' \emph{Advances in neural information processing systems}, vol.~35, pp. 24\,824--24\,837, 2022.

\bibitem{reflexion}
N.~Shinn, F.~Cassano, A.~Gopinath, K.~Narasimhan, and S.~Yao, ``Reflexion: Language agents with verbal reinforcement learning,'' \emph{Advances in neural information processing systems}, vol.~36, pp. 8634--8652, 2023.

\bibitem{rawal2021recent}
A.~Rawal, J.~McCoy, D.~B. Rawat, B.~M. Sadler, and R.~S. Amant, ``Recent advances in trustworthy explainable artificial intelligence: Status, challenges, and perspectives,'' \emph{IEEE Transactions on Artificial Intelligence}, vol.~3, no.~6, pp. 852--866, 2021.

\bibitem{deepseekr1}
D.~Guo, D.~Yang, H.~Zhang, J.~Song, P.~Wang, Q.~Zhu, R.~Xu, R.~Zhang, S.~Ma, X.~Bi \emph{et~al.}, ``Deepseek-r1 incentivizes reasoning in llms through reinforcement learning,'' \emph{Nature}, vol. 645, no. 8081, pp. 633--638, 2025.

\bibitem{hagos2024recent}
D.~H. Hagos, R.~Battle, and D.~B. Rawat, ``Recent advances in generative ai and large language models: Current status, challenges, and perspectives,'' \emph{IEEE transactions on artificial intelligence}, vol.~5, no.~12, pp. 5873--5893, 2024.

\bibitem{lu2025optmath}
H.~Lu, Z.~Xie, Y.~Wu, C.~Ren, Y.~Chen, and Z.~Wen, ``Optmath: A scalable bidirectional data synthesis framework for optimization modeling,'' \emph{arXiv preprint arXiv:2502.11102}, 2025.

\bibitem{Or-llm-agent}
B.~Zhang and P.~Luo, ``Or-llm-agent: Automating modeling and solving of operations research optimization problem with reasoning large language model,'' \emph{arXiv e-prints}, pp. arXiv--2503, 2025.

\bibitem{huang2025orlm}
C.~Huang, Z.~Tang, S.~Hu, R.~Jiang, X.~Zheng, D.~Ge, B.~Wang, and Z.~Wang, ``Orlm: A customizable framework in training large models for automated optimization modeling,'' \emph{Operations Research}, vol.~73, no.~6, pp. 2986--3009, 2025.

\bibitem{LLMOPT}
C.~Jiang, X.~Shu, H.~Qian, X.~Lu, J.~Zhou, A.~Zhou, and Y.~Yu, ``Llmopt: Learning to define and solve general optimization problems from scratch,'' \emph{arXiv preprint arXiv:2410.13213}, 2024.

\bibitem{Optibench}
Z.~Yang, Y.~Wang, Y.~Huang, Z.~Guo, W.~Shi, X.~Han, L.~Feng, L.~Song, X.~Liang, and J.~Tang, ``Optibench meets resocratic: Measure and improve llms for optimization modeling,'' \emph{arXiv preprint arXiv:2407.09887}, 2024.

\bibitem{zhou2025steporlm}
C.~Zhou, T.~Xu, J.~Lin, and D.~Ge, ``Steporlm: A self-evolving framework with generative process supervision for operations research language models,'' \emph{arXiv preprint arXiv:2509.22558}, 2025.

\bibitem{kojima2022large}
T.~Kojima, S.~S. Gu, M.~Reid, Y.~Matsuo, and Y.~Iwasawa, ``Large language models are zero-shot reasoners,'' \emph{Advances in neural information processing systems}, vol.~35, pp. 22\,199--22\,213, 2022.

\bibitem{ramamonjison2022augmenting}
R.~Ramamonjison, H.~Li, T.~Yu, S.~He, V.~Rengan, A.~Banitalebi-Dehkordi, Z.~Zhou, and Y.~Zhang, ``Augmenting operations research with auto-formulation of optimization models from problem descriptions,'' in \emph{Proceedings of the 2022 Conference on Empirical Methods in Natural Language Processing: Industry Track}, 2022, pp. 29--62.

\bibitem{OptiMUS-03}
A.~AhmadiTeshnizi, W.~Gao, H.~Brunborg, S.~Talaei, C.~Lawless, and M.~Udell, ``Optimus-0.3: Using large language models to model and solve optimization problems at scale,'' \emph{arXiv preprint arXiv:2407.19633}, 2024.

\bibitem{liu2025optitree}
H.~Liu, J.~Wang, Y.~Cai, X.~Han, Y.~Kuang, and J.~Hao, ``Optitree: Hierarchical thoughts generation with tree search for llm optimization modeling,'' \emph{arXiv preprint arXiv:2510.22192}, 2025.

\bibitem{wangor}
Y.~Wang, H.~Zhou, D.~Mao, L.~Li, J.~Tan, H.~Han, Z.~Yang, A.~J. Wang, and M.~Li, ``Or-prm: A process reward model for algorithmic problem in operations research,'' in \emph{The Fourteenth International Conference on Learning Representations}.

\bibitem{=llmandagent}
M.~A. Ferrag, N.~Tihanyi, and M.~Debbah, ``From llm reasoning to autonomous ai agents: A comprehensive review,'' \emph{arXiv preprint arXiv:2504.19678}, 2025.

\bibitem{gao2024meta}
P.~Gao, A.~Xie, S.~Mao, W.~Wu, Y.~Xia, H.~Mi, and F.~Wei, ``Meta reasoning for large language models,'' \emph{arXiv preprint arXiv:2406.11698}, 2024.

\bibitem{jaech2024openai}
A.~Jaech, A.~Kalai, A.~Lerer, A.~Richardson, A.~El-Kishky, A.~Low, A.~Helyar, A.~Madry, A.~Beutel, A.~Carney \emph{et~al.}, ``Openai o1 system card,'' \emph{arXiv preprint arXiv:2412.16720}, 2024.

\bibitem{masterman2024landscape}
T.~Masterman, S.~Besen, M.~Sawtell, and A.~Chao, ``The landscape of emerging ai agent architectures for reasoning, planning, and tool calling: A survey,'' \emph{arXiv preprint arXiv:2404.11584}, 2024.

\bibitem{hou2025model}
X.~Hou, Y.~Zhao, S.~Wang, and H.~Wang, ``Model context protocol (mcp): Landscape, security threats, and future research directions,'' \emph{ACM Transactions on Software Engineering and Methodology}, 2025.

\bibitem{xu2026agent}
R.~Xu and Y.~Yan, ``Agent skills for large language models: Architecture, acquisition, security, and the path forward,'' \emph{arXiv preprint arXiv:2602.12430}, 2026.

\bibitem{hong2023metagpt}
S.~Hong, M.~Zhuge, J.~Chen, X.~Zheng, Y.~Cheng, J.~Wang, C.~Zhang, Z.~Wang, S.~K.~S. Yau, Z.~Lin \emph{et~al.}, ``Metagpt: Meta programming for a multi-agent collaborative framework,'' in \emph{The twelfth international conference on learning representations}, 2023.

\bibitem{fan2023attention}
B.~Fan, X.~Liu, G.~Xiao, Y.~Kang, D.~Wang, and P.~Wang, ``Attention-based multiagent graph reinforcement learning for service restoration,'' \emph{IEEE Transactions on Artificial Intelligence}, vol.~5, no.~5, pp. 2163--2178, 2023.

\bibitem{chen2024multiagent}
Z.~Chen, L.~Yu, S.~Zhang, S.~Hu, and C.~Shen, ``Multiagent hierarchical deep reinforcement learning for operation optimization of grid-interactive efficient commercial buildings,'' \emph{IEEE Transactions on Artificial Intelligence}, vol.~5, no.~8, pp. 4280--4292, 2024.

\bibitem{wu2024evolutionary}
X.~Wu, S.-h. Wu, J.~Wu, L.~Feng, and K.~C. Tan, ``Evolutionary computation in the era of large language model: Survey and roadmap,'' \emph{IEEE Transactions on Evolutionary Computation}, vol.~29, no.~2, pp. 534--554, 2024.

\bibitem{romera2024mathematical}
B.~Romera-Paredes, M.~Barekatain, A.~Novikov, M.~Balog, M.~P. Kumar, E.~Dupont, F.~J. Ruiz, J.~S. Ellenberg, P.~Wang, O.~Fawzi \emph{et~al.}, ``Mathematical discoveries from program search with large language models,'' \emph{Nature}, vol. 625, no. 7995, pp. 468--475, 2024.

\bibitem{lehman2023evolution}
J.~Lehman, J.~Gordon, S.~Jain, K.~Ndousse, C.~Yeh, and K.~O. Stanley, ``Evolution through large models,'' in \emph{Handbook of evolutionary machine learning}.\hskip 1em plus 0.5em minus 0.4em\relax Springer, 2023, pp. 331--366.

\bibitem{novikov2025alphaevolve}
A.~Novikov, N.~V{\~u}, M.~Eisenberger, E.~Dupont, P.-S. Huang, A.~Z. Wagner, S.~Shirobokov, B.~Kozlovskii, F.~J. Ruiz, A.~Mehrabian \emph{et~al.}, ``Alphaevolve: A coding agent for scientific and algorithmic discovery,'' \emph{arXiv preprint arXiv:2506.13131}, 2025.

\bibitem{guo2023evoprompt}
Q.~Guo, R.~Wang, J.~Guo, B.~Li, K.~Song, X.~Tan, G.~Liu, J.~Bian, and Y.~Yang, ``Evoprompt: Connecting llms with evolutionary algorithms yields powerful prompt optimizers,'' \emph{arXiv e-prints}, pp. arXiv--2309, 2023.

\bibitem{yuan2025evoagent}
S.~Yuan, K.~Song, J.~Chen, X.~Tan, D.~Li, and D.~Yang, ``Evoagent: Towards automatic multi-agent generation via evolutionary algorithms,'' in \emph{Proceedings of the 2025 Conference of the Nations of the Americas Chapter of the Association for Computational Linguistics: Human Language Technologies (Volume 1: Long Papers)}, 2025, pp. 6192--6217.

\bibitem{gepa}
L.~A. Agrawal, S.~Tan, D.~Soylu, N.~Ziems, R.~Khare, K.~Opsahl-Ong, A.~Singhvi, H.~Shandilya, M.~J. Ryan, M.~Jiang \emph{et~al.}, ``Gepa: Reflective prompt evolution can outperform reinforcement learning,'' \emph{arXiv preprint arXiv:2507.19457}, 2025.

\bibitem{cemri2026adaevolve}
M.~Cemri, S.~Agrawal, A.~Gupta, S.~Liu, A.~Cheng, Q.~Mang, A.~Naren, L.~E. Erdogan, K.~Sen, M.~Zaharia \emph{et~al.}, ``Adaevolve: Adaptive llm driven zeroth-order optimization,'' \emph{arXiv preprint arXiv:2602.20133}, 2026.

\bibitem{wang2025thetaevolve}
Y.~Wang, S.-R. Su, Z.~Zeng, E.~Xu, L.~Ren, X.~Yang, Z.~Huang, X.~He, L.~Ma, B.~Peng \emph{et~al.}, ``Thetaevolve: Test-time learning on open problems,'' \emph{arXiv preprint arXiv:2511.23473}, 2025.

\bibitem{huang2024mamo}
X.~Huang, Q.~Shen, Y.~Hu, A.~Gao, and B.~Wang, ``Mamo: a mathematical modeling benchmark with solvers,'' \emph{arXiv preprint arXiv:2405.13144}, 2024.

\bibitem{xiao2025survey}
Z.~Xiao, J.~Xie, L.~Xu, S.~Guan, J.~Zhu, X.~Han, X.~Fu, W.~Yu, H.~Wu, W.~Shi \emph{et~al.}, ``A survey of optimization modeling meets llms: Progress and future directions,'' \emph{arXiv preprint arXiv:2508.10047}, 2025.

\bibitem{liu2025deepseek}
A.~Liu, A.~Mei, B.~Lin, B.~Xue, B.~Wang, B.~Xu, B.~Wu, B.~Zhang, C.~Lin, C.~Dong \emph{et~al.}, ``Deepseek-v3. 2: Pushing the frontier of open large language models,'' \emph{arXiv preprint arXiv:2512.02556}, 2025.

\bibitem{singh2025openai}
A.~Singh, A.~Fry, A.~Perelman, A.~Tart, A.~Ganesh, A.~El-Kishky, A.~McLaughlin, A.~Low, A.~Ostrow, A.~Ananthram \emph{et~al.}, ``Openai gpt-5 system card,'' \emph{arXiv preprint arXiv:2601.03267}, 2025.

\bibitem{deepmind2025gemini3}
\BIBentryALTinterwordspacing
DeepMind, ``Gemini 3 pro model card,'' Google DeepMind, Model Card, 2025b. [Online]. Available: \url{https://storage.googleapis.com/deepmind-media/Model-Cards/Gemini-3-Pro-Model-Card.pdf}
\BIBentrySTDinterwordspacing

\bibitem{yang2025qwen3}
A.~Yang, A.~Li, B.~Yang, B.~Zhang, B.~Hui, B.~Zheng, B.~Yu, C.~Gao, C.~Huang, C.~Lv \emph{et~al.}, ``Qwen3 technical report,'' \emph{arXiv preprint arXiv:2505.09388}, 2025.

\end{thebibliography}

\vfill
\clearpage
\startsupplementarydocument
\section{Prompt Templates and Process}
\label{app:workflow_prompts}

This supplementary document provides a reproducible account of the prompt design, workflow organization, experimental configuration, and knowledge-base construction underlying the Co-evolving Agent framework. To remain fully aligned with the methodology developed in the main paper, Section~S.I presents the prompt and process components according to the same end-to-end lifecycle followed by the agent itself, including initialization, AOE-based structural abstraction, topology-preserving recompilation, semantic mutation, and tool-grounded execution. In this way, the supplementary material does not treat prompts as isolated text artifacts; instead, it documents them as interdependent elements of a unified reasoning--evolution pipeline, clarifying both their individual functions and their roles in the overall system.

\subsection{Prompt Templates}
\label{app:prompt_templates}

This subsection presents the prompt families that implement the main components of the proposed framework in a form consistent with the \textit{The Proposed Framework} section of the main paper. The prompts are organized according to the same methodological order used there: initialization of executable reasoning individuals, conversion between agent code and phase-wise AOE chains, phase-local state alignment for architecture graph construction, and semantic mutation for reasoning trajectory evolution. Table~\ref{tab:prompt_family_summary} summarizes this organization before the full prompt texts are presented.

\begin{table}[h]
  \centering
  \caption{Prompt families and their roles in architecture graph evolution and reasoning trajectory evolution.}
  \label{tab:prompt_family_summary}
  \footnotesize
  \setlength{\tabcolsep}{5pt}
  \renewcommand{\arraystretch}{1.22}
  \begin{tabularx}{\columnwidth}{@{}>{\centering\arraybackslash}p{0.16\columnwidth} >{\raggedright\arraybackslash}p{0.25\columnwidth} >{\raggedright\arraybackslash}X@{}}
    \toprule
    \textbf{Prompt Box} & \textbf{Module} & \textbf{Primary Role} \\
    \midrule
    S1 & Zero-shot initialization & Generates executable reasoning individuals that satisfy the ordered OR phases, mandatory tool calls, and execution-repair behavior required for the initial parent population. \\
    S2 & KB-guided initialization & Incorporates knowledge-base priors into initialization so that the initial parent population starts from more credible OR modeling and debugging patterns. \\
    S4 & Code $\rightarrow$ AOE-Chain & Converts executable agent code into a phase-wise AOE chain that can be inserted into the architecture graph and reused by downstream evolutionary operators. \\
    S5 & Phase-local state alignment & Merges semantically equivalent states within the same OR phase to support architecture graph construction while preserving interpretability. \\
    S6 & AOE-Chain $\rightarrow$ code & Reconstructs executable reasoning individuals from feasible phase-wise AOE chains while preserving explicit topological differences. \\
    S7--S8 & Semantic mutation & Revises existing reasoning individuals with unguided or knowledge-guided mutation to balance diversity preservation and mathematically informed improvement. \\
    \bottomrule
  \end{tabularx}
  \vspace{2pt}
\end{table}

\subsubsection{Initialization Prompts}
\label{app:prompt_init}

Initialization is the entry point of the full framework because all later operators act on executable reasoning individuals rather than on free-form text alone. In the terminology of the main paper, this stage produces the initial parent population by instantiating feasible OR-agent workflows that already follow the ordered phases of problem analysis, mathematical modeling, and code generation. Prompt Box~\ref{box:zero_shot_init} therefore defines the minimal executable template for an individual in the co-evolutionary process. It specifies the mandatory phase order, required tool invocation, and repair behavior needed before an individual can be abstracted into a phase-wise AOE chain and inserted into the architecture graph.

\promptboxlabelnum{1}{box:zero_shot_init}
\begin{promptbox}[Prompt Box~\thepromptbox: Zero-shot Initialization Workflow Prompt]
  You are a Heuristic operations research Agent Generator.

  [Task Description]
  You are tasked with constructing an OR optimization agent strictly adhering to a three-stage main workflow:
  \begin{enumerate}
    \item Stage 1: Problem Analysis. Requirement: Thoroughly analyze the problem description to identify and extract the sets, parameters, decision variables, objective functions, and constraints.
    \item Stage 2: Mathematical Modeling. Requirement: Based on the analytical components derived in Stage 1, formulate a rigorous and computable mathematical model.
    \item Stage 3: Code Generation and Execution Repair. Requirement: Translate the mathematical model from Stage 2 into a complete Python Gurobi script, execute it immediately, and autonomously repair the code upon failure until successful execution or a maximum retry limit is reached.
  \end{enumerate}

  [Heuristic Guidelines]
  \begin{enumerate}
    \item Single-file output: The generated script must be a single, standalone Python file.
    \item Strict dependency chain: Stage 2 must utilize the exact outputs from Stage 1; Stage 3 must integrate the mathematical model formulated in Stage 2.
    \item Constraints preceding generation: Explicitly define the mathematical boundaries and logic before generating the code.
    \item Minimum viable retry mechanism:
      \begin{itemize}
        \item Execute the code immediately after initial generation.
        \item If execution fails, return the complete error traceback and explicitly prompt for a ``complete code rewrite.''
        \item The default maximum number of retries is set to 3.
      \end{itemize}
  \end{enumerate}

  [Architectural Framework]
  \begin{enumerate}
    \item Three-stage core workflow: Problem Analysis $\rightarrow$ Mathematical Modeling $\rightarrow$ Code Generation and Repair.
    \item Strict intermediate state transfer: Stage 1 output flows into Stage 2; Stage 2 output flows into Stage 3.
    \item Error handling: Trigger a repair loop upon execution failure; trigger a structural backtrack if execution succeeds but yields a non-numerical result.
    \item Essential tool integration: query\_llm, extract\_and\_execute\_python\_code, save\_generated\_code, eval\_model\_result.
    \item Evaluation pipeline: The run\_eval function must iterate over the dataset, evaluating each problem and computing the run pass and solve correct metrics.
  \end{enumerate}

  [Operational Constraints]
  Strictly utilize the load\_dataset tool to read experimental data (do not assume or synthesize dummy test data). Evaluate results exclusively via the eval\_model\_result tool.

  [Tool Invocation Specifications]
  The following contexts must be strictly referenced, learned, and implemented:

  [tool.txt]
  \verb|{tool_doc}|

  [new\_utils.py Source Code]
  \verb|{new_utils_source}|

  Mandatory Invocation Requirements:
  \begin{enumerate}
    \item The statement from new\_utils import must include at least: query\_llm, save\_generated\_code, extract\_and\_execute\_python\_code, eval\_model\_result, load\_dataset, extract\_best\_objective.
    \item The code generation and repair paths must invoke: query\_llm, save\_generated\_code, extract\_and\_execute\_python\_code.
    \item The optimization result extraction path must invoke: extract\_best\_objective.
    \item The evaluation path must invoke: eval\_model\_result.
  \end{enumerate}

  [Final Output Formatting]
  \begin{enumerate}
    \item Output exclusively the complete Python executable code.
    \item Do not include any natural language explanations or Markdown code fences.
  \end{enumerate}
\end{promptbox}

Following Prompt Box~\ref{box:zero_shot_init}, the foundational heuristic agent script is generated. Listing~\ref{lst:zero_shot_code} shows the core control logic of this baseline implementation. Even in this simplified form, the listing already exhibits the architectural commitments emphasized in the methodology section: a staged transition from mathematical reasoning to code generation, explicit invocation of tool interfaces, and a bounded self-repair loop that keeps the workflow executable under runtime failure.

\begin{lstlisting}[language=Python, caption={Listing~S1. Core execution and repair logic of the zero-shot initialized EvoOR-Agent.}, label={lst:zero_shot_code}]
  def agent(user_question, model_name=DEFAULT_MODEL_NAME, max_attempts=3):
  messages = [
  {
    "role": "system",
    "content": (
    "You are an expert in operations research. Based on the optimization problem, please construct a mathematical model..."
    ),
  },
  {"role": "user", "content": user_question},
  ]

  math_model = query_llm(messages, model_name)
  messages.append({"role": "assistant", "content": math_model})
  messages.append({
    "role": "user",
    "content": "Based on the above mathematical model, use Gurobi to write complete and reliable Python code...",
  })

  is_solve_success, result, messages = generate_or_code_solver(messages, model_name, max_attempts)

  # Structural Backtrack for infeasible outputs
  if is_solve_success and not is_number_string(str(result)):
  messages.append({
    "role": "user",
    "content": "The current model yields *no feasible solution*. Please check the mathematical model and Gurobi code...",
  })
  is_solve_success, result, messages = generate_or_code_solver(messages, model_name, max_attempts=1)

  return is_solve_success, result
\end{lstlisting}

Building upon the zero-shot baseline, we next introduce the knowledge base (KB)-guided initialization prompt to reduce logical drift in complex OR formulation. This variant corresponds to the knowledge-augmented branch of the methodology: instead of letting the model rely only on generic reasoning ability, the prompt explicitly injects retrieved domain priors so that stage-wise analysis, modeling, and debugging are conditioned on validated OR heuristics. In practice, this shifts initialization from a purely heuristic scaffold to a guided architectural prior, improving the likelihood that the first generated agent already occupies a structurally meaningful region of the search space. The full prompt template is given in Prompt Box~\ref{box:kb_init}.

\promptboxlabelnum{2}{box:kb_init}
\begin{promptbox}[Prompt Box~\thepromptbox: Knowledge-Base Guided Initialization Prompt]
  You are a Knowledge-Guided Heuristic OR Agent Generator.

  [Retrieved Expert Knowledge]
  You have been provided with the following expert heuristics and modeling patterns retrieved from the domain knowledge base:
  \begin{quote}
    \verb|{retrieved_knowledge}|
  \end{quote}
  Instruction: You must rigorously analyze this retrieved knowledge. Apply the relevant modeling formulations, constraint structures, and semantic debugging strategies to the current problem to prevent common logical errors.

  [Task Description]
  You are tasked with constructing an OR optimization agent strictly adhering to a three-stage main workflow, while integrating the expert knowledge provided above:
  \begin{enumerate}
    \item Stage 1: Problem Analysis. Requirement: Thoroughly analyze the problem description to identify sets, parameters, variables, objectives, and constraints. Cross-reference these elements with the [Retrieved Expert Knowledge] to ensure accurate structural extraction.
    \item Stage 2: Mathematical Modeling. Requirement: Formulate a rigorous and computable mathematical model. Explicitly utilize the constraint formulation patterns suggested in the knowledge base.
    \item Stage 3: Code Generation and Execution Repair. Requirement: Translate the mathematical model into a complete Python Gurobi script, execute it immediately, and autonomously repair the code upon failure. If debugging is required, prioritize the debugging heuristics from the knowledge base.
  \end{enumerate}

  [Heuristic Guidelines]
  \begin{enumerate}
    \item Single-file output: The generated script must be a single, standalone Python file.
    \item Strict dependency chain: Stage 2 must utilize the exact outputs from Stage 1; Stage 3 must integrate the mathematical model formulated in Stage 2.
    \item Constraints preceding generation: Explicitly define the mathematical boundaries and logic before generating the code.
  \end{enumerate}

  [Architectural Framework]
  \begin{enumerate}
    \item Three-stage core workflow: Problem Analysis $\rightarrow$ Mathematical Modeling $\rightarrow$ Code Generation and Repair.
    \item Strict intermediate state transfer: Stage 1 output flows into Stage 2; Stage 2 output flows into Stage 3.
    \item Error handling: Trigger a repair loop upon execution failure; trigger a structural backtrack if execution succeeds but yields a non-numerical result.
    \item Essential tool integration: query\_llm, extract\_and\_execute\_python\_code, save\_generated\_code, eval\_model\_result.
    \item Evaluation pipeline: The run\_eval function must iterate over the dataset, evaluating each problem and computing the run pass and solve correct metrics.
  \end{enumerate}

  [Operational Constraints]
  Strictly utilize the load\_dataset tool to read experimental data (do not assume or synthesize dummy test data). Evaluate results exclusively via the eval\_model\_result tool.

  [Tool Invocation Specifications]
  The following contexts must be strictly referenced, learned, and implemented:

  [tool.txt]
  \verb|{tool_doc}|

  [new\_utils.py Source Code]
  \verb|{new_utils_source}|

  Mandatory Invocation Requirements:
  \begin{enumerate}
    \item The statement from new\_utils import must include at least: query\_llm, save\_generated\_code, extract\_and\_execute\_python\_code, eval\_model\_result, load\_dataset, extract\_best\_objective.
    \item The code generation and repair paths must invoke: query\_llm, save\_generated\_code, extract\_and\_execute\_python\_code.
    \item The optimization result extraction path must invoke: extract\_best\_objective.
    \item The evaluation path must invoke: eval\_model\_result.
  \end{enumerate}

  [Final Output Formatting]
  \begin{enumerate}
    \item Output exclusively the complete Python executable code.
    \item Do not include any natural language explanations or Markdown code fences.
  \end{enumerate}
\end{promptbox}

Listing~\ref{lst:kb_init_code} presents the corresponding KB-guided variant. Relative to the zero-shot baseline, this implementation makes the stage decomposition more explicit and repeatedly re-injects retrieved priors during both formulation and repair. This design is important from an evolutionary perspective: it does not merely change wording, but alters how information flows through the architecture by binding expert priors to the three-stage reasoning chain and the subsequent debugging loop.

\begin{lstlisting}[language=Python, caption={Listing~S2. Core execution logic of the KB-guided EvoOR-Agent, demonstrating the explicit three-stage pipeline and prior knowledge integration.}, label={lst:kb_init_code}]
  import copy
  from new_utils import (
  query_llm, save_generated_code, extract_and_execute_python_code, is_number_string
  )

  def generate_or_code_solver(messages_bak, model_name, max_attempts):
  messages = copy.deepcopy(messages_bak)

  gurobi_code = query_llm(messages, model_name)
  print("[Python Gurobi Code]:\n", gurobi_code)
  save_generated_code(gurobi_code, prefix="agent")

  text = f"{gurobi_code}"
  attempt = 0
  while attempt < max_attempts:
  success, error_msg = extract_and_execute_python_code(text)
  if success:
  messages_bak.append({"role": "assistant", "content": gurobi_code})
  return True, error_msg, messages_bak

  print(f"\nAttempt {attempt + 1} failed, requesting LLM to fix code...\n")
  messages.append({"role": "assistant", "content": gurobi_code})
  messages.append({
    "role": "user",
    "content": f"An error occurred during code execution. The error message is as follows:\n{error_msg}\nPlease fix the code and provide the complete executable code again.",
  })

  gurobi_code = query_llm(messages, model_name)
  save_generated_code(gurobi_code, prefix="agent_fix")
  text = f"{gurobi_code}"
  attempt += 1

  messages_bak.append({"role": "assistant", "content": gurobi_code})
  print(f"Reached maximum number of attempts ({max_attempts}).")
  return False, None, messages_bak

  def or_agent(user_question, model_name=DEFAULT_MODEL_NAME):
  messages = [
  {
    "role": "system",
    "content": (
    "You are an expert in operations research. Integrate the following "
    ),
  }
  ]

  # Stage 1: Problem Analysis (Applying KB structural patterns)
  messages.append({
    "role": "user",
    "content": f"Stage 1: Thoroughly analyze the following problem to identify sets, parameters, and variables:\n{user_question}"
  })
  analysis_result = query_llm(messages, model_name)
  messages.append({"role": "assistant", "content": analysis_result})

  # Stage 2: Mathematical Modeling
  messages.append({
    "role": "user",
    "content": "Stage 2: Based on the analysis, formulate a rigorous mathematical model. Explicitly utilize the constraint formulation patterns"
  })
  math_model = query_llm(messages, model_name)
  messages.append({"role": "assistant", "content": math_model})

  # Stage 3: Code Generation
  messages.append({
    "role": "user",
    "content": "Stage 3: Translate the mathematical model into a complete Python Gurobi script. Format as ```python\n{code}\n```."
  })

  is_solve_success, result, messages = generate_or_code_solver(messages, model_name, max_attempts, retrieved_knowledge)

  return is_solve_success, result
\end{lstlisting}

\subsubsection{Transformation: Code to AOE-Chain}
\label{app:prompt_code_to_aoe}

To support architecture graph evolution, each executable agent must next be converted into the phase-wise AOE chain defined in the main paper. This conversion is the step that maps a runnable reasoning individual into a structured execution trace over ordered OR phases. Prompt Box~\ref{box:code_to_aoe} therefore does more than summarize code. It defines a reversible extraction protocol in which the workflow is decomposed into ordered state-action units that are fine-grained enough for path-conditioned recombination and semantic mutation, while still preserving the information needed to reconstruct an executable individual.

\promptboxlabelnum{4}{box:code_to_aoe}
\begin{promptbox}[Prompt Box~\thepromptbox: Code to AOE-Chain Extraction Prompt]
  Please reverse-engineer a ``reversible state-action trajectory JSON array'' based on the optimization agent Python code provided below. Output ONLY JSON, without any explanations.

  [Objectives]
  \begin{enumerate}
    \item This JSON must be sufficiently precise to support a lossless bidirectional transformation: JSON $\leftrightarrow$ Code.
    \item You must strictly adhere to the provided code; do not hallucinate or invent steps that do not exist in the code.
    \item The trajectory must be linear, executable, and non-skippable.
    \item Each action must explicitly declare a start state, an end state, and an action description.
    \item Each action must be granularized into the ``minimal but meaningful'' work unit, which should be neither too coarse nor fragmented to the point of losing semantic integrity.
    \item The array must cover the complete workflow, including code control flow, prompt-driven LLM reasoning nodes, and tool invocations.
  \end{enumerate}

  This analysis targets variant number \verb|{variant_index}| / \verb|{total_variants}|.

  [Output Requirements]
  \begin{enumerate}
    \item Output strictly a JSON array.
    \item Do NOT use markdown code fences (e.g., ```json or ```).
    \item The array cannot be empty.
    \item Continuity constraint: The end\_state of the previous action must perfectly match the start\_state of the subsequent action.
    \item type must be strictly one of: code, prompt, or tool.
    \item key must encapsulate the most critical, minimized, and executable mapping core information of that action.
    \item Do not omit any critical state transitions.
    \item phase must be restricted to the three standard stages. The start and end states for each phase must strictly adhere to the following logic:
      \begin{itemize}
        \item 1. Problem Analysis
          \begin{itemize}
            \item start\_state: Agent Initialization
            \item end\_state: Problem Analysis Complete
          \end{itemize}
        \item 2. Mathematical Modeling
          \begin{itemize}
            \item start\_state: Problem Analysis Complete
            \item end\_state: Mathematical Modeling Complete
          \end{itemize}
        \item 3. Code Generation
          \begin{itemize}
            \item start\_state: Mathematical Modeling Complete
            \item end\_state: Code Generation Complete
          \end{itemize}
      \end{itemize}
  \end{enumerate}

  [Element Fields Definition]
  Each JSON element must contain the following fields: phase, type, action, start\_state, end\_state, key.

  [Field Constraints]
  \begin{itemize}
    \item phase: The name of the phase where the current action resides.
    \item type = code: Denotes a workflow action within the Python code.
    \item type = prompt: Denotes a prompt-driven cognitive node of the Large Language Model.
    \item type = tool: Denotes the invocation of new\_utils.py or other tool functions.
    \item action: A precise description of the action.
    \item start\_state: The system state at the beginning of the action.
    \item end\_state: The system state achieved upon execution of the action.
    \item key: The minimal critical unit.
      \begin{itemize}
        \item If type = code: Extract the complete code for that step exactly as it is. If it is a function, extract the complete function.
        \item If type = prompt: Extract the complete minimal cognitive node verbatim from the prompt, not the entire prompt. Multiple cognitive nodes collectively form the complete prompt.
        \item If type = tool: Write the specific tool invocation rule for that step.
      \end{itemize}
  \end{itemize}

  [Strict Constraints]
  \begin{enumerate}
    \item By sequentially combining the key values in the JSON structure, one must be able to fundamentally reconstruct the source code. Pay strict attention to the completeness of the key and its consistency with the source code.
    \item Avoid extracting trivial input/output operations as primary workflow-advancing actions.
  \end{enumerate}

  You may refer to the following tool constraints:
  \begin{quote}
    \verb|{tool_doc}|
  \end{quote}

  Agent code to be reverse-engineered:
  \begin{quote}
    \verb|{agent_code}|
  \end{quote}
\end{promptbox}

The resulting decomposition converts a continuous script into a phase-wise AOE chain whose edges are typed as executable code, tool invocation, or prompt-driven reasoning operations. A particularly important design choice is that long prompts are split into minimal semantic units rather than kept as indivisible text blocks. This gives the evolutionary algorithm a manipulable search space at the level where reasoning behavior actually changes, making it possible to perform path-conditioned recombination and localized mutation without breaking the integrity of the full reasoning trajectory.

\subsubsection{State Merging}
\label{app:prompt_state_merge}

Because different phase-wise AOE chains may describe equivalent reasoning states with different surface forms, a dedicated state-alignment step is required before the architecture graph can be initialized or updated. Prompt Box~\ref{box:state_merging} operationalizes this phase-local merging step. Its role is to merge only functionally identical states while preserving type isolation across prompt, tool, and programmatic semantics. This is consistent with the graph-construction constraints in the main paper. If semantically different states were merged too aggressively, the architecture graph would lose interpretability and no longer support faithful structural reuse.

\promptboxlabelnum{5}{box:state_merging}
\begin{promptbox}[Prompt Box~\thepromptbox: Semantic State Merging Prompt]
  You are a State Merging Assistant. Based on the input JSON array of records, your task is to merge nodes that are semantically similar and can be considered the identical state.

  [Input Specification]
  Each record represents a state node candidate containing rich context (consistent with the previously extracted AOE-Chain format).

  [Merging Objectives \& Constraints]
  \begin{itemize}
    \item Group state\_text nodes that share identical semantics, synonymy, or referential consistency into the same cluster.
    \item The state\_text within a group may be phrased differently, but their underlying semantics must be absolutely identical.
    \item Do NOT merge states with distinctly different semantics. If there are contextual conflicts (e.g., inconsistent dependencies, file paths, or functional meanings), do not merge them.
    \item Cross-role merging is permitted conditionally, but you must strictly adhere to semantic consistency.
    \item Type Isolation: Programmatic function state nodes MUST NOT be merged with prompt-driven state nodes.
    \item Every node\_id must appear in exactly ONE group; omissions are strictly prohibited.
  \end{itemize}

  [Output Format (Strict JSON)]
  \begin{verbatim}
    {
      "groups": [
      {
        "canonical_state": "...",
        "members": [
        {"node_id": "..."},
        {"node_id": "..."}
        ]
      }
      ]
    }
  \end{verbatim}

  [Formatting Requirements]
  \begin{enumerate}
    \item Output strictly in JSON format. Do not append any natural language explanations.
    \item Each node\_id must be uniquely assigned to one and only one group.
    \item If a record cannot be logically merged with any others, it must be assigned to its own independent group.
  \end{enumerate}
\end{promptbox}

\subsubsection{Transformation: AOE-Chain to Code}
\label{app:prompt_aoe_to_code}

To complete the reversible mapping, any extracted or mutated phase-wise AOE chain must be compiled back into executable Python code. Prompt Box~\ref{box:aoe_to_code} defines this inverse transformation. Methodologically, this step closes the loop between architecture graph evolution and runnable reasoning individuals. Structural edits along a feasible path in the graph are required to reappear as concrete changes in prompts, tool calls, control flow, and repair logic. The synthesis constraints intentionally forbid the insertion of unlisted operations so that downstream performance changes can be attributed to explicit topological differences rather than uncontrolled regeneration noise.

\promptboxlabelnum{6}{box:aoe_to_code}
\begin{promptbox}[Prompt Box~\thepromptbox: AOE-Chain to Code Synthesis Prompt]
  You are an AOE-Chain Synthesis Compiler. Based on the provided activity-on-edge (AOE) structural JSON array, your task is to reconstruct the complete and executable Python optimization agent code.

  [Input Specification]
  You are provided with a sequential JSON array representing the structural state-action trajectory of an optimization workflow. Each node contains a phase, type, start\_state, end\_state, and an executable key.

  [Synthesis Objectives \& Protocol]
  \begin{enumerate}
    \item Deterministic Reconstruction: You must sequentially traverse the provided JSON array from the first node to the last. The key fields serve as the core building blocks for the code reconstruction.
    \item Type-Specific Translation:
      \begin{itemize}
        \item If type = code: Directly implement the exact programmatic logic defined in the key.
        \item If type = prompt: Formulate the key into the appropriate message payload for the LLM invocation, ensuring context retention.
        \item If type = tool: Implement the exact invocation of the specified tool function (e.g., query\_llm, save\_generated\_code).
      \end{itemize}
    \item Topological Fidelity: You must strictly adhere to the provided state-action trajectory. Do NOT invent, hallucinate, or inject any programmatic steps, prompts, or tool invocations that do not explicitly exist in the JSON array.
    \item Executable Integrity: The synthesized code must be a complete, structurally sound, and immediately executable Python script. Ensure all necessary variable assignments, loop structures, and conditional branches implied by the sequence are properly synthesized.
  \end{enumerate}

  [Output Requirements]
  \begin{enumerate}
    \item Output ONLY the reconstructed Python code.
    \item Use the standard Markdown formatting: \texttt{```python\textbackslash n\{code\}\textbackslash n```}.
    \item Do not append any natural language explanations, debugging suggestions, or comments outside the code block.
  \end{enumerate}

  AOE-Chain JSON Array to be synthesized:
  \begin{quote}
    \verb|{aoe_chain_json}|
  \end{quote}
\end{promptbox}
\subsubsection{Mutation Prompts}
\label{app:prompt_mutation}

In the reasoning trajectory evolution stage, semantic mutation is the main operator for revising existing individuals and escaping locally stable but suboptimal workflows. We implement the same two mutation modes described in the main paper. Prompt Box~\ref{box:direct_mutation} defines unguided semantic mutation, which perturbs the current individual without external knowledge support and helps preserve diversity. Prompt Box~\ref{box:kb_mutation} defines knowledge-guided semantic mutation, which revises the current individual using retrieved OR modeling heuristics and implementation patterns. Together, these prompts realize the exploration--guidance balance of the mutation operator: one broadens the search space, while the other steers revision toward mathematically credible regions.

\promptboxlabelnum{7}{box:direct_mutation}
\begin{promptbox}[Prompt Box~\thepromptbox: Direct Semantic Mutation Prompt]
  You are an Evolutionary Code Mutator. Your task is to apply semantic mutations to the provided Python optimization agent code to generate a novel, mathematically sound variant.

  [Task Description]
  Analyze the input code and introduce meaningful structural or algorithmic variations. The goal is to explore alternative logic paths that might yield better performance or robustness in operations research tasks.

  [Mutation Guidelines]
  \begin{enumerate}
    \item Algorithmic Variation: Modify Gurobi parameters (e.g., MIPGap, TimeLimit), introduce alternative constraint formulations (e.g., big-M vs. indicator constraints), or alter the loop structures in the repair mechanism.
    \item Prompt Engineering: If the code contains internal prompts sent to the LLM (e.g., messages.append(...)), mutate the phrasing to elicit different reasoning styles (e.g., step-by-step logic, mathematical formalization).
    \item Preserve Executability: The mutated code MUST remain syntactically correct and fully executable. Do not break the core data loading or evaluation pipeline (new\_utils).
    \item Maintain Objective: The fundamental goal of the code (solving the OR problem) must remain unchanged.
  \end{enumerate}

  [Output Requirements]
  \begin{enumerate}
    \item Output strictly the complete, mutated Python code.
    \item Use the standard Markdown formatting: \texttt{```python\textbackslash n\{code\}\textbackslash n```}.
    \item Do not append any natural language explanations, change logs, or comments outside the code block.
  \end{enumerate}

  Source Code to Mutate:
  \begin{quote}
    \verb|{source_code}|
  \end{quote}
\end{promptbox}

\promptboxlabelnum{8}{box:kb_mutation}
\begin{promptbox}[Prompt Box~\thepromptbox: Knowledge-Guided Mutation Prompt]
  You are a Knowledge-Guided Evolutionary Code Mutator. Your task is to apply semantic mutations to the provided Python optimization code strictly guided by the retrieved expert operations research heuristics.

  [Retrieved Expert Knowledge]
  \begin{quote}
    \verb|{retrieved_knowledge}|
  \end{quote}

  [Task Description]
  Analyze the input code and introduce structural variations by explicitly integrating the strategies, constraint patterns, or debugging logic suggested in the expert knowledge above.

  [Mutation Guidelines]
  \begin{enumerate}
    \item Knowledge Integration: Identify areas in the source code (especially within modeling constraints or the LLM prompt payloads) that contradict or fail to utilize the retrieved knowledge. Mutate these sections to rigorously align with the expert heuristics.
    \item Structural Optimization: If the knowledge suggests a more efficient mathematical formulation (e.g., symmetry breaking, cutting planes), mutate the Gurobi generation logic to adopt this structure.
    \item Preserve Executability: The mutated code MUST remain syntactically correct and fully executable. Do not break the core data loading or evaluation pipeline (new\_utils).
    \item Maintain Objective: The fundamental goal of the code (solving the OR problem) must remain unchanged.
  \end{enumerate}

  [Output Requirements]
  \begin{enumerate}
    \item Output strictly the complete, mutated Python code.
    \item Use the standard Markdown formatting: \texttt{```python\textbackslash n\{code\}\textbackslash n```}.
    \item Do not append any natural language explanations, change logs, or comments outside the code block.
  \end{enumerate}

  Source Code to Mutate:
  \begin{quote}
    \verb|{source_code}|
  \end{quote}
\end{promptbox}

\subsection{Tool Configuration Appendix}
\label{app:tools}

Evaluated EvoOR-Agent relies on a stable utility layer that connects prompt-based reasoning with executable evaluation. These functions, implemented in \texttt{new\_utils.py}, form the operational basis of the full co-evolutionary loop. \texttt{query\_llm} provides a unified interface for formulation, code generation, reflection, and repair. \texttt{load\_dataset} supplies benchmark instances and the fields required by the evaluation pipeline. \texttt{save\_generated\_code} preserves intermediate artifacts so that each generated candidate remains traceable across iterations.

\texttt{extract\_and\_execute\_python\_code} links generated text to actual solver behavior. It extracts Python code blocks from model output, executes them in an isolated environment, and returns either a valid result or a full traceback. The companion function \texttt{extract\_best\_objective} reads the target objective value from raw solver output and returns \texttt{None} when infeasibility or parsing failure prevents valid extraction. After that, \texttt{eval\_model\_result} evaluates each candidate using the same criteria emphasized in the main paper, namely whether the program runs successfully and whether the numerical answer is correct within tolerance. \texttt{log\_llm\_chat} records prompt-response trajectories for reproducibility, diagnosis, and evolutionary analysis. Together, these utilities keep the reasoning process inspectable, reproducible, and tied to benchmark data instead of ad hoc execution behavior.

Table~\ref{tab:tool_config_summary} summarizes the role of each utility in the executable agent workflow. The framework does not treat these functions as generic helpers. Instead, each one is assigned a fixed control responsibility, including model interaction, dataset access, artifact persistence, execution, objective parsing, metric evaluation, and trajectory logging. This separation is important for the methodology in the main paper because evolutionary search is meant to change reasoning structure and prompt content without changing the measurement interface. As a result, architectural variation takes place above a stable utility layer, which keeps comparisons across variants fair and makes failure modes easier to interpret.

\begin{table}[h]
  \centering
  \caption{Configuration-level summary of the utility layer used by the Co-evolving Agent.}
  \label{tab:tool_config_summary}
  \footnotesize
  \setlength{\tabcolsep}{5pt}
  \renewcommand{\arraystretch}{1.22}
  \begin{tabularx}{\columnwidth}{@{}>{\centering\arraybackslash}m{0.27\columnwidth} >{\raggedright\arraybackslash}p{0.22\columnwidth} >{\raggedright\arraybackslash}X@{}}
    \toprule
    \textbf{Tools} & \textbf{Primary Role} & \textbf{Operational Constraint in the Framework} \\
    \midrule
    \shortstack[c]{\scriptsize\texttt{query}\\\scriptsize\texttt{\_llm}} & Unified LLM access layer & Used for all major reasoning stages so that initialization, repair, and mutation share the same invocation interface and logging semantics. \\
    \shortstack[c]{\scriptsize\texttt{load}\\\scriptsize\texttt{\_dataset}} & Benchmark instance loader & Serves as the exclusive data ingress path; generated agents are not allowed to fabricate placeholder instances or bypass benchmark loading. \\
    \shortstack[c]{\scriptsize\texttt{save\_generated}\\\scriptsize\texttt{\_code}} & Artifact persistence & Saves every generated or repaired solver candidate with a stage-aware prefix, enabling replay and lineage tracing across evolutionary iterations. \\
    \shortstack[c]{\scriptsize\texttt{extract\_and}\\\scriptsize\texttt{\_execute}\\\scriptsize\texttt{\_python\_code}} & Sandboxed execution bridge & Converts textual code output into an executable trial and returns either runtime success or the full traceback needed for repair. \\
    \shortstack[c]{\scriptsize\texttt{extract\_best}\\\scriptsize\texttt{\_objective}} & Objective-value parser & Converts raw solver terminal output into the scalar objective used for downstream correctness judgment; parse failure is treated as a non-valid result. \\
    \shortstack[c]{\scriptsize\texttt{eval\_model}\\\scriptsize\texttt{\_result}} & Benchmark scoring interface & Computes the run-pass and solve-correct signals under the same dataset-driven evaluation policy for every evolved agent. \\
    \shortstack[c]{\scriptsize\texttt{log\_llm}\\\scriptsize\texttt{\_chat}} & Reproducibility logger & Stores prompt--response trajectories so that architectural decisions, debugging loops, and mutation effects remain auditable after execution. \\
    \bottomrule
  \end{tabularx}
  \vspace{2pt}
\end{table}

From a process perspective, the utility layer defines a fixed execution contract. Problem instances enter the system only through \texttt{load\_dataset}. Every reasoning step that requires language generation passes through \texttt{query\_llm}. All generated solver programs are written out by \texttt{save\_generated\_code} before execution, so no candidate exists only in transient memory. Runtime validation is handled by \texttt{extract\_and\_execute\_python\_code}, and the resulting solver output is converted into benchmark-level signals by \texttt{extract\_best\_objective} and \texttt{eval\_model\_result}. The full interaction history is then preserved by \texttt{log\_llm\_chat}. This ordering forms the fixed tool-grounded loop assumed by the prompts in Section~\ref{app:prompt_templates} and by the trajectory reconstruction mechanism introduced later in this supplementary document.

To keep the configuration auditable, generated agents are also required to reference the same tool names in their import statements and execution paths. This avoids a common reproducibility problem in LLM-based systems, where semantically similar but implementation-divergent helper functions are silently substituted across variants. By binding all candidates to a stable utility vocabulary, the framework makes it possible to interpret performance gains as genuine improvements in architecture or reasoning rather than artifacts of inconsistent infrastructure.

\subsection{AOE Network Representation}
\label{app:aoe_network}

Consistent with the architecture graph formulation in the main paper, we represent each executable agent as a structured AOE network whose nodes denote intermediate reasoning states and whose directed edges denote executable transitions between states. This representation makes the internal organization of the agent explicit. It also provides the same structured view used by architecture graph evolution, where executable reasoning individuals are abstracted into phase-wise AOE chains and then merged into a maintained architecture space.

\begin{figure*}[!htbp]
  \centering
  \includegraphics[width=\textwidth]{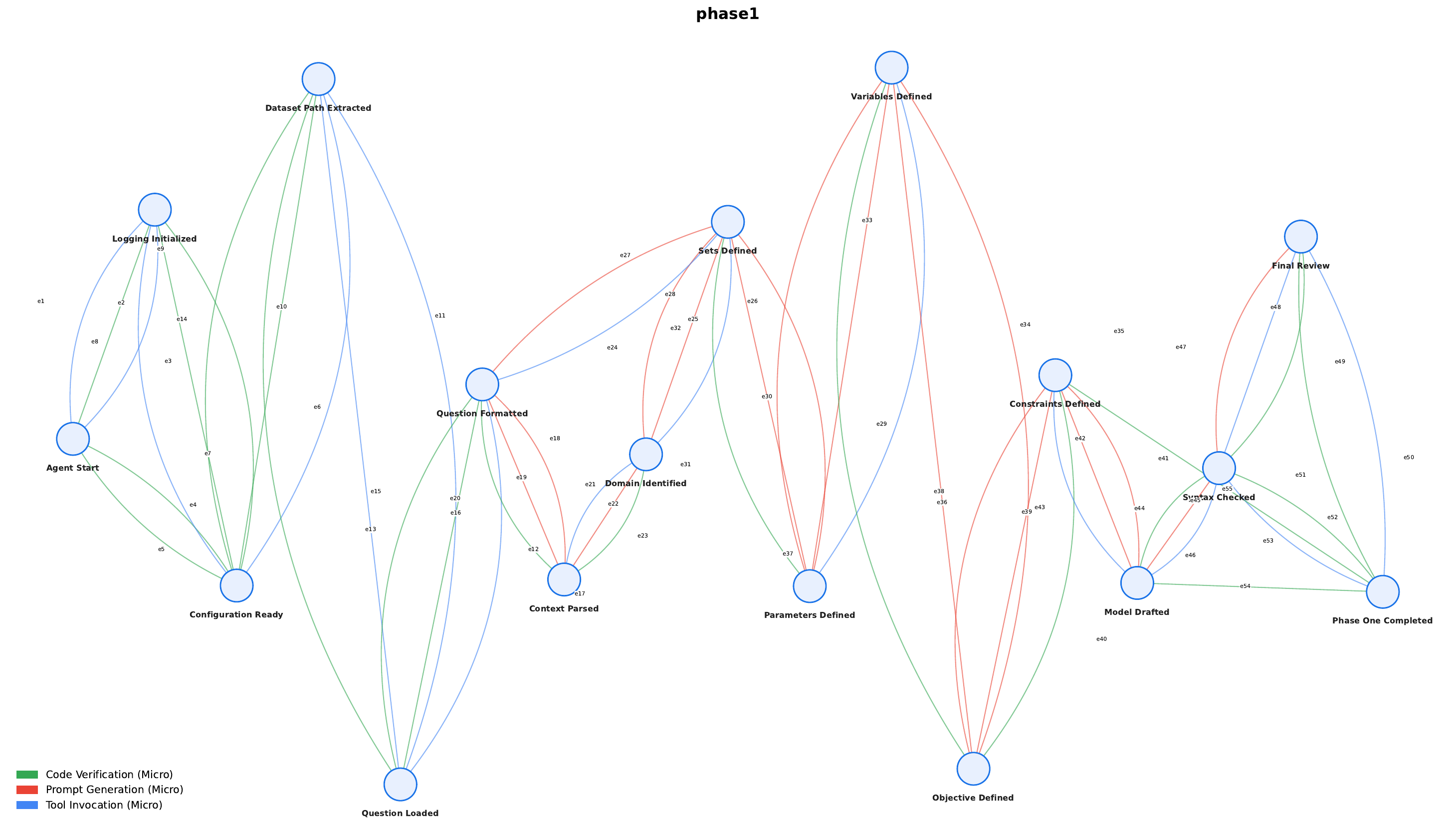}
  \caption{{AOE topology for \textbf{Phase 1: Problem Analysis and Mathematical Modeling}. The graph exposes the alternative transitions used to parse task context, identify OR structure, define the mathematical model, and assemble the formulation draft.}}
  \label{fig:aoe_phase1}
\end{figure*}

\begin{figure*}[!htbp]
  \centering
  \includegraphics[width=0.94\textwidth]{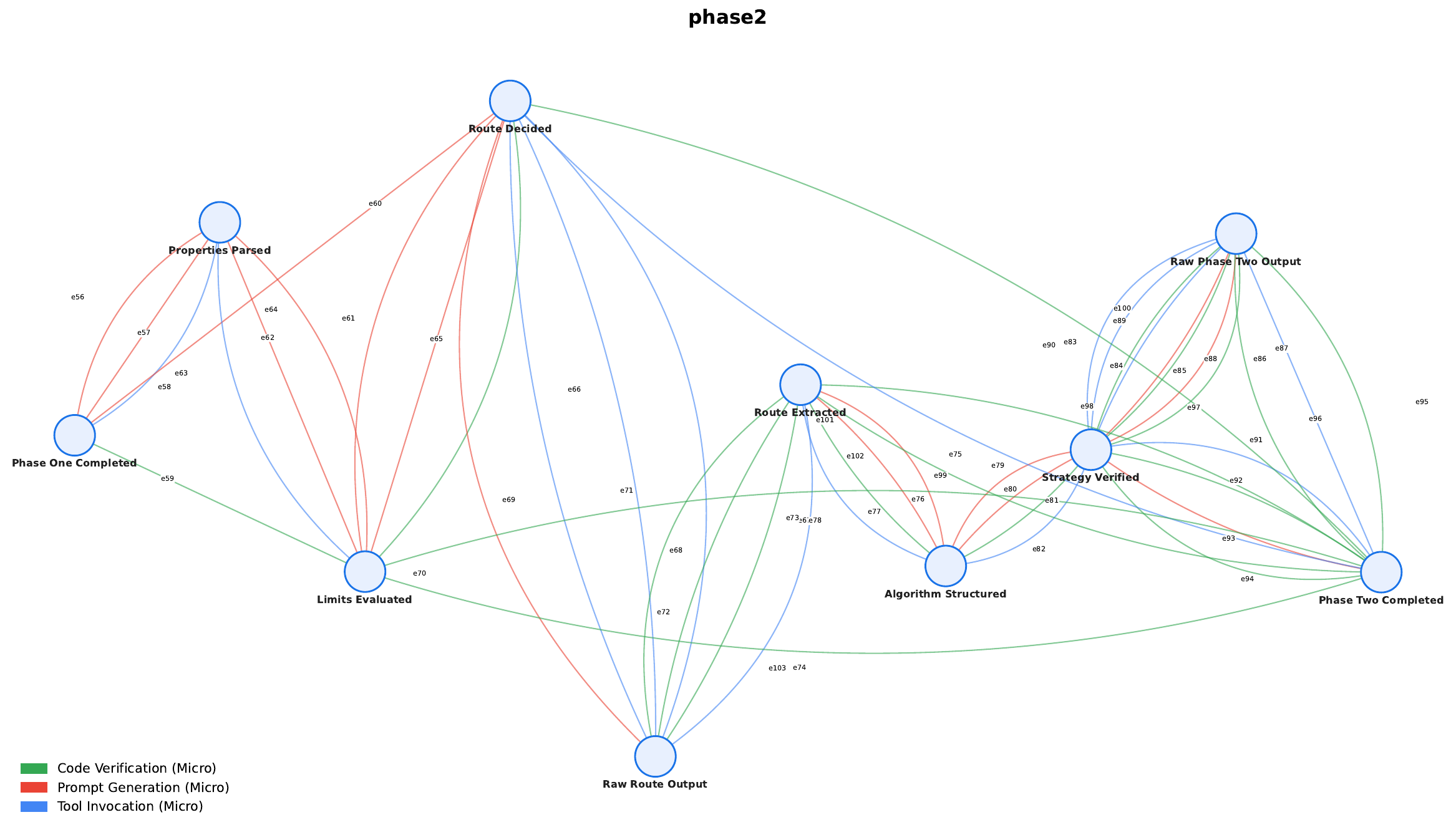}
  \caption{AOE topology for \textbf{Phase 2: Algorithm Design and Strategy Verification}. This phase captures branching decisions for solver selection, algorithmic route planning, verification, and early commitment to standard or heuristic solution pathways.}
  \label{fig:aoe_phase2}
\end{figure*}
\begin{figure*}[!t]
  \centering
  \includegraphics[width=0.94\textwidth]{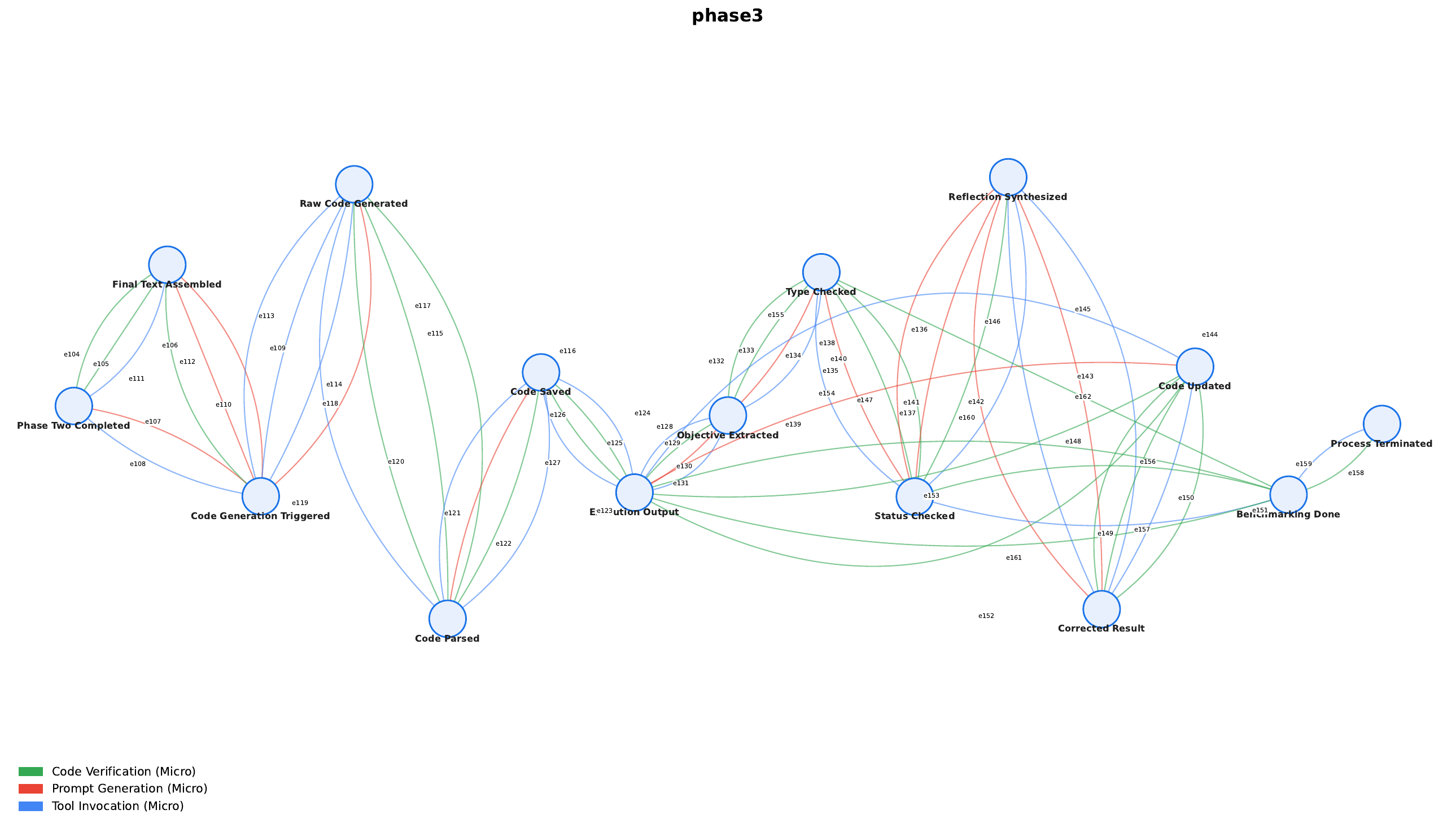}
  \caption{AOE topology for \textbf{Phase 3: Code Generation, Execution, and Repair}. The graph emphasizes executable code synthesis, runtime evaluation, objective extraction, reflective debugging, and loop-based correction dynamics.}
  \label{fig:aoe_phase3}
\end{figure*}

At the level of the supplementary material, we visualize the resulting network in three ordered functional segments. Together, these segments cover problem analysis, mathematical modeling, intermediate strategy organization, code generation, execution, and repair. Figures~\ref{fig:aoe_phase1}--\ref{fig:aoe_phase3} show these parts separately so that readers can inspect how prompt-based reasoning steps, tool invocations, and executable control transitions are distributed across the full reasoning trajectory.

Across the three segments, the network contains \textbf{162} micro-level trajectory edges, with 55, 48, and 59 edges in the three parts, respectively. This decomposition reflects a central point of the main paper. The framework does not optimize only local prompt wording. Instead, it evolves reasoning trajectories over a structured architecture space. The network therefore makes visible where the agent can branch, where alternative reasoning paths appear, and where execution and repair loops are re-entered.

Table~\ref{tab:full_AOE} provides the complete edge-level mapping of the network. Each row lists the edge ID, transition type (\texttt{prompt}, \texttt{tool}, or \texttt{code}), the start state, the end state, and the corresponding action detail. Used together, the phase diagrams and the full edge table provide a direct link between graph topology and executable semantics. This makes it possible to trace evolved reasoning trajectories and to examine how the maintained architecture space is instantiated in executable agent behavior.

\clearpage
\onecolumn

\begin{longtable}{@{\extracolsep{\fill}} >{\centering\arraybackslash}p{0.05\textwidth} >{\centering\arraybackslash}p{0.08\textwidth} >{\raggedright\arraybackslash}p{0.16\textwidth} >{\raggedright\arraybackslash}p{0.16\textwidth} >{\raggedright\arraybackslash}p{0.45\textwidth} @{}}
  \caption{\centering Comprehensive Micro-level State-Action Trajectory Network (162 Edges)}
  \label{tab:full_AOE} \\
  \toprule
  \textbf{ID} & \textbf{Type} & \textbf{Start State} & \textbf{End State} & \multicolumn{1}{c}{\textbf{Action Detail}} \\
  \midrule
  \endfirsthead

  \multicolumn{5}{c}%
  {{\bfseries \tablename\ \thetable{} -- continued from previous page}} \\
  \toprule
  \textbf{ID} & \textbf{Type} & \textbf{Start State} & \textbf{End State} & \multicolumn{1}{c}{\textbf{Action Detail}} \\
  \midrule
  \endhead

  \midrule \multicolumn{5}{r}{{Continued on next page}} \\
  \endfoot

  \bottomrule
  \endlastfoot

  \multicolumn{5}{c}{\textbf{Phase 1: Problem Analysis and Mathematical Modeling (55 Edges)}} \\
  \midrule
  1 & tool & Agent Start & Logging Initialized & \texttt{log\_llm\_chat} \\
  2 & code & Agent Start & Logging Initialized & Initialize an asynchronous file handler for storing evolutionary interaction logs \\
  3 & tool & Agent Start & Logging Initialized & \texttt{initialize\_wandb\_experiment\_tracker} \\
  4 & code & Agent Start & Configuration Ready & Load deeply cached hyper-parameters to bypass redundant logging initialization \\
  5 & code & Agent Start & Configuration Ready & Parse the local environment variables to establish execution boundaries immediately \\
  6 & code & Logging Initialized & Configuration Ready & Verify and load system environment variables and API keys \\
  7 & code & Logging Initialized & Configuration Ready & Validate the integrity of the YAML configuration files programmatically \\
  8 & tool & Logging Initialized & Configuration Ready & \texttt{load\_system\_configuration} \\
  9 & code & Configuration Ready & Dataset Path Extracted & Construct the absolute file pathway for the operations research benchmark dataset \\
  10 & code & Configuration Ready & Dataset Path Extracted & Validate the JSON schema configuration for the target input stream \\
  11 & tool & Configuration Ready & Dataset Path Extracted & \texttt{verify\_dataset\_checksum\_integrity} \\
  12 & tool & Dataset Path Extracted & Question Loaded & \texttt{load\_dataset} \\
  13 & tool & Dataset Path Extracted & Question Loaded & \texttt{stream\_jsonl\_benchmark\_data} \\
  14 & code & Dataset Path Extracted & Question Loaded & Retrieve multi-modal context from the local vectorized database \\
  15 & code & Question Loaded & Question Formatted & Format the raw natural language problem description into the standardized reasoning template \\
  16 & code & Question Loaded & Question Formatted & Inject retrieved few-shot mathematical modeling examples from the knowledge base \\
  17 & tool & Question Loaded & Question Formatted & \texttt{clean\_markdown\_and\_html\_artifacts} \\
  18 & prompt & Question Formatted & Context Parsed & \textit{``...comprehensively analyze the background context of the supply chain operations research problem...''} \\
  19 & prompt & Question Formatted & Context Parsed & \textit{``...extract all implicit and explicit logical dependencies mentioned in the text...''} \\
  20 & code & Question Formatted & Context Parsed & Execute a deterministic regular expression sweep for known domain-specific terminology \\
  21 & tool & Context Parsed & Domain Identified & \texttt{query\_llm} \\
  22 & prompt & Context Parsed & Domain Identified & \textit{``...classify the underlying mathematical structure as a Vehicle Routing or Facility Location problem...''} \\
  23 & code & Context Parsed & Domain Identified & Map the extracted context to the internal operations research taxonomy tree \\
  24 & prompt & Domain Identified & Sets Defined & \textit{``...explicitly define all mathematical index sets required for spatial and temporal dimensions...''} \\
  25 & prompt & Domain Identified & Sets Defined & \textit{``...identify all network nodes, vehicle fleets, and operational periods as independent sets...''} \\
  26 & tool & Domain Identified & Sets Defined & \texttt{extract\_mathematical\_sets\_via\_regex} \\
  27 & prompt & Question Formatted & Sets Defined & \textit{``...bypass intermediate context parsing and directly define the fundamental mathematical sets...''} \\
  28 & tool & Question Formatted & Sets Defined & \texttt{query\_llm\_reasoner\_model} \\
  29 & prompt & Sets Defined & Parameters Defined & \textit{``...extract all explicitly stated numerical parameters such as cost matrices and distances...''} \\
  30 & prompt & Sets Defined & Parameters Defined & \textit{``...determine structural bounds and identify implicit zero-cost parameter assumptions...''} \\
  31 & code & Sets Defined & Parameters Defined & Verify the dimensional consistency of the extracted parameter matrices programmatically \\
  32 & prompt & Parameters Defined & Variables Defined & \textit{``...formulate continuous flow variables and binary assignment variables with tight bounds...''} \\
  33 & prompt & Parameters Defined & Variables Defined & \textit{``...define auxiliary tracking variables required for sub-tour elimination constraints...''} \\
  34 & tool & Parameters Defined & Variables Defined & \texttt{validate\_variable\_dimension\_alignment} \\
  35 & prompt & Variables Defined & Objective Defined & \textit{``...formulate the overarching Objective Function to minimize aggregate operational costs...''} \\
  36 & prompt & Variables Defined & Objective Defined & \textit{``...identify multi-objective components and establish appropriate scalarization weights...''} \\
  37 & code & Variables Defined & Objective Defined & Execute symbolic verification using the SymPy library to guarantee mathematical convexity \\
  38 & prompt & Objective Defined & Constraints Defined & \textit{``...construct rigorous mathematical constraints ensuring flow conservation across all network nodes...''} \\
  39 & prompt & Objective Defined & Constraints Defined & \textit{``...apply Big-M relaxation techniques to linearize conditional 'if-then' logical statements...''} \\
  40 & code & Objective Defined & Constraints Defined & Isolate raw constraint text blocks into independent executable Python dictionary elements \\
  41 & prompt & Constraints Defined & Model Drafted & \textit{``...synthesize the defined sets, parameters, variables, objective, and constraints into a cohesive model...''} \\
  42 & prompt & Constraints Defined & Model Drafted & \textit{``...output the fully assembled Mixed-Integer Linear Programming formulation in LaTeX format...''} \\
  43 & tool & Constraints Defined & Model Drafted & \texttt{query\_llm\_fast\_model} \\
  44 & code & Model Drafted & Syntax Checked & Parse the generated mathematical draft algorithmically to detect missing brackets and symbols \\
  45 & prompt & Model Drafted & Syntax Checked & \textit{``...ensure the synthesized mathematical syntax is strictly linear and computationally resolvable...''} \\
  46 & tool & Model Drafted & Syntax Checked & \texttt{execute\_regular\_expression\_syntax\_cleaner} \\
  47 & prompt & Syntax Checked & Final Review & \textit{``...cross-reference the generated mathematical model against known operations research fallacies...''} \\
  48 & tool & Syntax Checked & Final Review & \texttt{query\_knowledge\_base\_heuristics} \\
  49 & code & Syntax Checked & Final Review & Programmatic inspection ensuring all five fundamental mathematical components are present \\
  50 & tool & Final Review & Phase One Completed & \texttt{save\_phase\_one\_intermediate\_context} \\
  51 & code & Final Review & Phase One Completed & Compile the validated mathematical model components into a serialized JSON payload \\
  52 & code & Syntax Checked & Phase One Completed & Bypass heuristic final review to preserve raw syntactical generation \\
  53 & tool & Syntax Checked & Phase One Completed & \texttt{clear\_gpu\_memory\_cache\_and\_flush} \\
  54 & code & Model Drafted & Phase One Completed & Force an early exit for simplistic unconstrained optimization problems \\
  55 & code & Constraints Defined & Phase One Completed & Directly commit constraints to system memory and terminate the modeling phase \\

  \midrule
  \multicolumn{5}{c}{\textbf{Phase 2: Algorithm Design and Strategy Verification (48 Edges)}} \\
  \midrule
  56 & prompt & Phase One Completed & Properties Parsed & \textit{``...analyze the extracted model to determine linearity, convexity, and integer variable scale...''} \\
  57 & prompt & Phase One Completed & Properties Parsed & \textit{``...examine the constraint matrix to identify potential symmetry-breaking opportunities...''} \\
  58 & tool & Phase One Completed & Properties Parsed & \texttt{execute\_structural\_property\_analyzer} \\
  59 & code & Phase One Completed & Limits Evaluated & Extract hardware specification limits and solver timeout boundaries from the configuration \\
  60 & prompt & Phase One Completed & Route Decided & \textit{``...evaluate the overarching mathematical structure to propose an immediate heuristic framework...''} \\
  61 & prompt & Properties Parsed & Limits Evaluated & \textit{``...evaluate if invoking a Gurobi exact solver is computationally viable given the scale...''} \\
  62 & prompt & Properties Parsed & Limits Evaluated & \textit{``...assess dynamic memory allocation requirements for the anticipated branch-and-bound tree...''} \\
  63 & tool & Properties Parsed & Limits Evaluated & \texttt{query\_historical\_performance\_database} \\
  64 & prompt & Limits Evaluated & Route Decided & \textit{``...decide algorithmic path: exact solver compilation versus designing a custom meta-heuristic...''} \\
  65 & prompt & Limits Evaluated & Route Decided & \textit{``...propose alternative column generation or Benders decomposition strategies if NP-hard...''} \\
  66 & code & Limits Evaluated & Route Decided & Apply deterministic surrogate models to estimate optimal algorithmic routing \\
  67 & tool & Route Decided & Raw Route Output & \texttt{query\_llm} \\
  68 & tool & Route Decided & Raw Route Output & \texttt{query\_llm\_expert\_mixture\_of\_agents} \\
  69 & tool & Route Decided & Raw Route Output & \texttt{query\_llm\_low\_temperature\_greedy} \\
  70 & prompt & Route Decided & Raw Route Output & \textit{``...generate a rigorous architectural design for the chosen combinatorial optimization algorithm...''} \\
  71 & code & Raw Route Output & Route Extracted & Extract the core algorithmic routing strategy from the verbose generation output \\
  72 & code & Raw Route Output & Route Extracted & Execute regular expression matching to isolate specific algorithm nomenclature tags \\
  73 & code & Raw Route Output & Route Extracted & Handle missing algorithmic classification tags gracefully by assigning fallback default solvers \\
  74 & tool & Raw Route Output & Route Extracted & \texttt{validate\_json\_routing\_schema\_integrity} \\
  75 & prompt & Route Extracted & Algorithm Structured & \textit{``...design a precise Gurobi Python implementation strategy based on the extracted routing...''} \\
  76 & prompt & Route Extracted & Algorithm Structured & \textit{``...design customized cutting-plane logic tailored specifically for the formulated constraints...''} \\
  77 & code & Route Extracted & Algorithm Structured & Inject standardized knowledge base algorithmic implementation templates into the active context \\
  78 & tool & Route Extracted & Algorithm Structured & \texttt{retrieve\_algorithmic\_code\_templates} \\
  79 & prompt & Algorithm Structured & Strategy Verified & \textit{``...verify the overarching logical consistency of the proposed algorithm design...''} \\
  80 & prompt & Algorithm Structured & Strategy Verified & \textit{``...verify that Big-M relaxation bounds are sufficiently tight to prevent numerical instability...''} \\
  81 & code & Algorithm Structured & Strategy Verified & Compare the proposed implementation strategy against an internal registry of anti-patterns \\
  82 & tool & Algorithm Structured & Strategy Verified & \texttt{execute\_static\_algorithmic\_analysis} \\
  83 & tool & Strategy Verified & Raw Phase Two Output & \texttt{query\_llm} \\
  84 & tool & Strategy Verified & Raw Phase Two Output & \texttt{query\_llm\_reasoner\_with\_reflection} \\
  85 & prompt & Strategy Verified & Raw Phase Two Output & \textit{``...ensure the detailed algorithm design is structurally sound and logically complete...''} \\
  86 & prompt & Strategy Verified & Raw Phase Two Output & \textit{``...review multi-threading configurations and solver environment parameters...''} \\
  87 & code & Raw Phase Two Output & Strategy Verified & [Backtrack] Incomplete pseudocode detected, forcing secondary algorithmic verification \\
  88 & code & Raw Phase Two Output & Strategy Verified & [Backtrack] Unrecognized optimization library imported, request framework correction \\
  89 & code & Raw Phase Two Output & Strategy Verified & [Backtrack] \textit{``...correct the flawed decomposition logic identified in the output...''} \\
  90 & tool & Raw Phase Two Output & Strategy Verified & \texttt{trigger\_local\_algorithmic\_backtrack} \\
  91 & tool & Strategy Verified & Phase Two Completed & \texttt{serialize\_algorithm\_design\_document} \\
  92 & code & Strategy Verified & Phase Two Completed & Compile the verified strategy into an intermediate representation suitable for code generation \\
  93 & prompt & Strategy Verified & Phase Two Completed & \textit{``...finalize the algorithmic structure and prepare the operational context for synthesis...''} \\
  94 & code & Strategy Verified & Phase Two Completed & Bypass raw textual output processing due to absolute confidence in heuristic logic \\
  95 & code & Raw Phase Two Output & Phase Two Completed & Clean logical formatting and Python indentation artifacts from the raw text \\
  96 & tool & Raw Phase Two Output & Phase Two Completed & \texttt{save\_phase\_two\_intermediate\_state} \\
  97 & code & Raw Phase Two Output & Phase Two Completed & Detect syntax anomalies in the isolated algorithm block and commit to memory \\
  98 & code & Route Extracted & Phase Two Completed & Fallback to default Gurobi exact solver due to complex routing engine evaluation failure \\
  99 & code & Route Extracted & Phase Two Completed & Direct path execution leveraging cached historically optimal routing strategies \\
  100 & code & Route Decided & Phase Two Completed & Heuristically bypass deep algorithm design for strictly linear and highly sparse matrices \\
  101 & tool & Route Decided & Phase Two Completed & \texttt{force\_trivial\_solver\_pathway} \\
  102 & code & Limits Evaluated & Phase Two Completed & Execution limits prohibit custom algorithms, forcing immediate standard solver synthesis \\
  103 & code & Limits Evaluated & Phase Two Completed & Memory constraints detected, triggering immediate low-footprint solver deployment \\

  \midrule
  \multicolumn{5}{c}{\textbf{Phase 3: Code Generation, Execution, and Repair (59 Edges)}} \\
  \midrule
  104 & code & Phase Two Completed & Final Text Assembled & Merge the validated mathematical model dictionary with the algorithmic strategy context \\
  105 & code & Phase Two Completed & Final Text Assembled & Perform context window truncation algorithms if maximum token limits are exceeded \\
  106 & tool & Phase Two Completed & Final Text Assembled & \texttt{inject\_secure\_api\_keys\_to\_context} \\
  107 & prompt & Phase Two Completed & Code Generation Triggered & \textit{``...bypass intermediate assembly and directly generate executable Python code using Gurobi...''} \\
  108 & tool & Phase Two Completed & Code Generation Triggered & \texttt{query\_llm\_monolithic\_generation} \\
  109 & prompt & Final Text Assembled & Code Generation Triggered & \textit{``...synthesize the complete executable Python script prioritizing constraint execution efficiency...''} \\
  110 & prompt & Final Text Assembled & Code Generation Triggered & \textit{``...ensure software imports explicitly include both networkx and gurobipy libraries...''} \\
  111 & code & Final Text Assembled & Code Generation Triggered & Enforce strict Python PEP8 line-length constraints and block formatting programmatically \\
  112 & tool & Code Generation Triggered & Raw Code Generated & \texttt{query\_llm} \\
  113 & tool & Code Generation Triggered & Raw Code Generated & \texttt{query\_llm\_coding\_specialist\_agent} \\
  114 & tool & Code Generation Triggered & Raw Code Generated & \texttt{query\_llm\_maximum\_token\_allocation} \\
  115 & prompt & Code Generation Triggered & Raw Code Generated & \textit{``...review generated syntax internally before finalizing the executable output block...''} \\
  116 & code & Raw Code Generated & Code Parsed & Extract the executable Python code block from within the Markdown fences \\
  117 & code & Raw Code Generated & Code Parsed & Handle missing Markdown code fences gracefully via fuzzy regular expression matching \\
  118 & code & Raw Code Generated & Code Parsed & Detect Python indentation and structural formatting errors programmatically \\
  119 & tool & Raw Code Generated & Code Parsed & \texttt{parse\_python\_abstract\_syntax\_tree} \\
  120 & tool & Code Parsed & Code Saved & \texttt{save\_generated\_code\_to\_disk} \\
  121 & prompt & Code Parsed & Code Saved & \textit{``...confirm the extracted code block is fundamentally runnable before saving...''} \\
  122 & code & Code Parsed & Code Saved & Write the Python artifact to a localized high-speed memory cache \\
  123 & tool & Code Parsed & Code Saved & \texttt{generate\_timestamped\_version\_backup} \\
  124 & tool & Code Saved & Execution Output & \texttt{extract\_and\_execute\_python\_code} \\
  125 & code & Code Saved & Execution Output & Check subprocess dynamic memory allocation limits prior to triggering execution \\
  126 & code & Code Saved & Execution Output & Monitor execution timeout thresholds and terminate hanging solver processes \\
  127 & tool & Code Saved & Execution Output & \texttt{execute\_script\_in\_docker\_sandbox} \\
  128 & tool & Execution Output & Objective Extracted & \texttt{extract\_best\_objective\_value} \\
  129 & code & Execution Output & Objective Extracted & Execute regular expression fallback mechanisms for missing standard 'Optimal cost' strings \\
  130 & prompt & Execution Output & Objective Extracted & \textit{``...locate and extract the lower bound value if the relative optimality gap is greater than zero...''} \\
  131 & tool & Execution Output & Objective Extracted & \texttt{parse\_gurobi\_terminal\_log\_file} \\
  132 & code & Objective Extracted & Type Checked & Validate if the extracted objective value strictly conforms to a numeric floating-point type \\
  133 & code & Objective Extracted & Type Checked & Convert the raw string output payload into a standardized floating-point tensor \\
  134 & prompt & Objective Extracted & Type Checked & \textit{``...ensure the numerical result is strictly positive as required by this specific problem...''} \\
  135 & tool & Objective Extracted & Type Checked & \texttt{handle\_nonetype\_extraction\_exceptions} \\
  136 & code & Type Checked & Status Checked & Check the solver terminal output for execution success or syntactical crash indicators \\
  137 & code & Type Checked & Status Checked & Validate the internal solution feasibility indicator variable within the Gurobi model \\
  138 & prompt & Type Checked & Status Checked & \textit{``...ensure the final result is not negative infinity or flagged as computationally unbounded...''} \\
  139 & tool & Type Checked & Status Checked & \texttt{evaluate\_gurobi\_status\_code\_array} \\
  140 & prompt & Status Checked & Reflection Synthesized & \textit{``...the solver yielded an INFEASIBLE status. Analyze the constraints logically...''} \\
  141 & prompt & Status Checked & Reflection Synthesized & \textit{``...identify the mathematically conflicting constraints causing the model infeasibility...''} \\
  142 & code & Status Checked & Reflection Synthesized & Parse the execution traceback to pinpoint the exact line of Python syntax failure \\
  143 & tool & Status Checked & Reflection Synthesized & \texttt{retrieve\_debugging\_heuristics\_from\_kb} \\
  144 & tool & Reflection Synthesized & Corrected Result & \texttt{query\_llm} \\
  145 & prompt & Reflection Synthesized & Corrected Result & \textit{``...rewrite the formulation logic to circumvent the identified strict infeasibility...''} \\
  146 & tool & Reflection Synthesized & Corrected Result & \texttt{query\_llm\_reasoner\_for\_debugging} \\
  147 & prompt & Reflection Synthesized & Corrected Result & \textit{``...adjust Big-M parameters dynamically based on the reflection analysis...''} \\
  148 & code & Corrected Result & Code Updated & Extract and integrate the newly generated corrected code block into the workflow \\
  149 & code & Corrected Result & Code Updated & Confirm abstract syntax tree changes differ fundamentally from the previous iteration \\
  150 & tool & Corrected Result & Code Updated & \texttt{regex\_replace\_faulty\_logic\_functions} \\
  151 & code & Corrected Result & Code Updated & Write the patched codebase back into the primary execution memory slot \\
  152 & code & Code Updated & Execution Output & [Loop] Re-trigger the sandbox execution environment with the updated code \\
  153 & code & Code Updated & Execution Output & [Loop] Bypass cache and force a hard re-run of the modified solver script \\
  154 & prompt & Code Updated & Execution Output & \textit{``...re-evaluate the execution efficiency of the patched Python logic...''} \\
  155 & tool & Code Updated & Execution Output & \texttt{increment\_debug\_attempt\_counter} \\
  156 & code & Status Checked & Benchmarking Done & Successful optimal execution detected, bypassing all repair and reflection loops entirely \\
  157 & tool & Status Checked & Benchmarking Done & \texttt{record\_successful\_evaluation\_metrics} \\
  158 & code & Process Terminated & Benchmarking Done & Maximum debugging iterations reached, mark evaluation as a terminal failure \\
  159 & tool & Process Terminated & Benchmarking Done & \texttt{aggregate\_final\_benchmark\_results} \\
  160 & code & Execution Output & Benchmarking Done & Catastrophic segmentation fault detected, abort process and log fatal framework error \\
  161 & code & Execution Output & Benchmarking Done & Hardware timeout exceeded consistently, forcefully terminate and log execution failure \\
  162 & code & Type Checked & Benchmarking Done & Objective value is fundamentally un-parsable NaN, force end of benchmarking cycle \\
\end{longtable}
\twocolumn

\section{Experimental Reproducibility Details}
\label{app:reproducibility}

This section provides the comprehensive experimental configurations required to reproduce the empirical results reported in the main paper. It summarizes the benchmark alignment, Large Language Model (LLM) API configurations, and the explicit software and hardware environments utilized in our comparative evaluations across heterogeneous OR tasks.

\subsection{Benchmark Specifications}
\label{app:benchmarks}

Consistent with the experimental setup delineated in the main text, we evaluate \textit{EvoOR-Agent} on a comprehensive suite encompassing seven core OR benchmark datasets. These datasets span a wide spectrum of operations research scenarios, ranging from theoretical academic word problems to practical industrial deployments. Crucially, as highlighted by a recent survey \cite{xiao2025survey}, prevailing open-source OR benchmarks suffer from non-negligible native error rates due to logical fallacies, ambiguous problem definitions, and incorrect ground-truth labels. To guarantee the absolute rigor of our empirical evaluation, this work builds upon their pioneering data-cleaning protocols by executing further manual cross-verification and deep cleaning on six public datasets. Flawed instances have been meticulously filtered out, and all remaining valid data points have been processed into a unified format. The final post-cleaned statistics and intrinsic characteristics of each benchmark are systematically summarized in Table~\ref{tab:dataset_stats}.

\begin{table}[htbp]
  \centering
  \caption{Statistics of the cleaned and self-constructed optimization problem benchmarks.}
  \label{tab:dataset_stats}
  \footnotesize
  \setlength{\tabcolsep}{4pt}
  \renewcommand{\arraystretch}{1.22}
  \begin{tabularx}{\columnwidth}{@{}>{\raggedright\arraybackslash}p{0.28\columnwidth} >{\centering\arraybackslash}p{0.14\columnwidth} >{\raggedright\arraybackslash}X@{}}
    \toprule
    \textbf{Dataset Name} & \textbf{Cleaned Size} & \textbf{Core Characteristics \& Mathematical Domain} \\
    \midrule
    NL4Opt \cite{ramamonjison2022augmenting} & 213 & Linear programming word problems across diverse domains. \\
    EasyLP \cite{huang2024mamo} & 545 & High school-level linear and mixed-integer programming problems. \\
    ComplexLP \cite{huang2024mamo} & 111 & Undergraduate-level complex LP and MILP challenges. \\
    NLP4LP \cite{OptiMUS-03} & 178 & Abstract, multi-step LP/MILP domain-specific formulations from classical human-authored settings. \\
    IndustryOR \cite{huang2025orlm} & 42 & Industrial-scale problems sourced from 8 different industries. \\
    ReSocratic \cite{Optibench} & 403 & Rich semantic formulations reverse-translated from structured document demonstrations. \\
    BWOR\cite{Or-llm-agent} & 82 & High-complexity problems collected from classic OR textbooks. \\
    \bottomrule
  \end{tabularx}
\end{table}

\textbf{NL4Opt} \cite{ramamonjison2022augmenting}. NL4Opt, short for \emph{natural language for optimization}, is a widely used benchmark for automated OR modeling. It contains annotated linear programming word problems drawn from domains such as sales, advertising, investment, and transportation. The dataset is designed to connect natural-language problem descriptions with formal optimization modeling elements, so it serves as a standard test bed for language-driven OR formulation. For our evaluation, we selected 213 validated problems from this benchmark.

\textbf{MAMO} \cite{huang2024mamo}. Mamo is a benchmark constructed to evaluate mathematical modeling ability across different levels of difficulty. It focuses specifically on whether an LLM can accurately translate a detailed natural language description into a correct mathematical formulation, excluding nonlinear or differential equation modeling. The benchmark is divided into EasyLP and ComplexLP. EasyLP contains mixed-integer linear programming problems and focuses on basic reasoning and formulation accuracy. ComplexLP includes instances with more advanced linear and mixed-integer programming structures, requiring stronger abstraction of constraints and more mature modeling skills. From this benchmark, we selected 545 easy and 111 complex instances for our study.

\textbf{NLP4LP} \cite{OptiMUS-03}. The NLP4LP benchmark evaluates LLM-based agents on translating natural language operations research problems into solver-ready code and mathematical models. It consists of human-authored LP and MILP problems, where each instance presents an optimization scenario from classical OR domains such as scheduling, knapsack allocation, and production planning. The benchmark provides a fine-grained testbed for assessing formulation accuracy. In our work, we chose to evaluate 178 verified instances from NLP4LP.

\textbf{IndustryOR} \cite{huang2025orlm}. IndustryOR is an industrial-scale benchmark built to reflect the complexity of practical OR deployment. It includes real-world optimization scenarios collected from eight industry sectors, including manufacturing, energy, logistics, retail, and finance. Because it spans multiple industrial settings and problem types, it is well suited for evaluating the robustness, adaptability, and deployment potential of generated solver code. For our experiments, we selected 42 validated problems from this benchmark.

\textbf{ReSocratic} \cite{Optibench}. The ReSocratic dataset is introduced alongside a data synthesis method that formats optimization demonstrations in a reverse manner first, and then back-translates them into a question. Through these intermediate reasoning steps, ReSocratic delivers higher quality and richer semantic configurations than prior pipeline methods. In this paper, we selected 403 validated problems from this benchmark.

\textbf{BWOR} \cite{Or-llm-agent}. Recent evaluations suggest that standard benchmarks do not always separate advanced reasoning LLMs clearly from non-reasoning models. To obtain a more discriminative test set, we construct BWOR, which contains 82 challenging OR modeling and solving problems collected from well-known OR textbooks. The problems are written in LaTeX-formatted natural language, often with accompanying tables, and remain closely connected to real OR scenarios. This level of difficulty gives a clearer view of the agent's mathematical reasoning and self-correction ability.

\subsection{LLM API Configuration}
\label{app:llm_configuration}

This subsection reports the configuration of the four foundation models used in EvoOR-Agent. All models were accessed through their official API endpoints and called through the unified interface implemented in \texttt{new\_utils.py}. This keeps model invocation consistent across initialization, semantic mutation, reflection, and code generation.

A uniform temperature of \textbf{1.0} was used in all calls so that each model could explore diverse reasoning trajectories under a common setting. The maximum output length (\texttt{max\_tokens}) was set to 8,192 to accommodate long mathematical formulations and executable solver code.

The model-specific settings are listed below.

\begin{itemize}
  \item \textbf{DeepSeek-V3.2} \cite{liu2025deepseek}: Invoked via the DeepSeek official API (OpenAI-compatible format) with the model identifier \texttt{deepseek-reasoner}. This model serves as the primary engine for logical abstraction of mixed-integer linear programming (MILP) structures. By explicitly leveraging its specialized reasoning mechanism (the ``thinking'' process), the agent effectively identifies implicit constraints and breaks down complex OR logic into manageable mathematical components.

  \item \textbf{GPT-5}\cite{singh2025openai}: Accessed through the OpenAI API using the model identifier \texttt{gpt-5}. GPT-5 is primarily employed for high-level semantic reflection and structural backtracking tasks. We utilized the \textit{Structured Outputs} mode to ensure that the generated AOE network JSON arrays strictly adhere to our predefined schema, guaranteeing flawless bidirectional transformation between graphs and code.

  \item \textbf{Gemini 3 Flash}\cite{deepmind2025gemini3}: Deployed via the Google Vertex AI platform with the model identifier \texttt{gemini-3-flash-preview}. Benefiting from its extensive context window, Gemini 3 Flash was used to integrate large-scale OR knowledge base priors and maintain long-range dependency tracking across multiple evolution generations without information truncation.

  \item \textbf{Qwen 3 Max}\cite{yang2025qwen3}: Invoked through the Alibaba Cloud DashScope platform with the model identifier \texttt{qwen3-max}. Qwen 3 Max demonstrated superior precision in generating algorithmic syntax, particularly for Gurobi and Pyomo solver code. It was selected as the designated specialist for the final code synthesis and programmatic repair stages of the pipeline.
\end{itemize}

Table \ref{tab:llm_api_params} summarizes the global and model-specific parameters utilized during the benchmarking process.

\begin{table}[htbp]
  \centering
  \caption{Detailed parameters for LLM API configurations.}
  \label{tab:llm_api_params}
  \scriptsize
  \setlength{\tabcolsep}{3pt}
  \renewcommand{\arraystretch}{1.18}
  \begin{tabularx}{\columnwidth}{@{}>{\centering\arraybackslash}p{0.18\columnwidth} >{\raggedright\arraybackslash}X >{\centering\arraybackslash}p{0.18\columnwidth} >{\centering\arraybackslash}p{0.13\columnwidth} >{\centering\arraybackslash}p{0.15\columnwidth}@{}}
    \toprule
    \rowcolor{gray!12}
    \textbf{Model} & \textbf{Model Identifier} & \textbf{API Platform} & \textbf{Temp.} & \shortstack[c]{\textbf{Max}\\\textbf{Tokens}} \\
    \midrule
    DeepSeek-V3.2 & \shortstack[l]{\texttt{deepseek-}\\\texttt{reasoner}} & DeepSeek & 1.0 & 8192 \\
    GPT-5 & \texttt{gpt-5} & OpenAI & 1.0 & 8192 \\
    Gemini 3 Flash & \shortstack[l]{\texttt{gemini-3-}\\\texttt{flash-preview}} & \shortstack[c]{Google\\Vertex AI} & 1.0 & 8192 \\
    Qwen 3 Max & \texttt{qwen3-max} & DashScope & 1.0 & 8192 \\
    \bottomrule
  \end{tabularx}
\end{table}

All API calls were executed over encrypted SSL connections. To maintain stable large-scale evaluation on subsets such as IndustryOR and BWOR, the \texttt{query\_llm} function used exponential backoff with at most five retry attempts per request.

\subsection{Software and Hardware Environment}
\label{app:env}

This subsection outlines the dual-track software and hardware configurations deployed across our evaluation, reproduction, and fine-tuning pipelines. To guarantee strict controlled variables and empirical integrity, the experimental infrastructure is bifurcated into: (i) an API-driven orchestrating framework for black-box models, and (ii) a dedicated GPU-accelerated environment for baseline reproduction and specialized fine-tuning.

\subsubsection{Hardware Configuration}
Depending on the diverse computational characteristics of the evaluation pipelines, two distinct high-performance cloud server configurations from the AutoDL platform were leveraged:

\begin{itemize}
  \item \textbf{Platform I (API Orchestration \& Local Evaluation)}: Designed primarily for coordinating massive concurrent API requests and executing local deterministic solvers. Local computation was driven by two CPU cluster alternatives to prevent bottlenecks:
    \begin{itemize}
      \item \textit{Configuration A (AMD Architecture)}: 32-core AMD EPYC\texttrademark~9654 CPU, 60 GB RAM, and 30 GB System Disk supplemented by 50 GB Data Disk.
        \textit{Configuration B (Intel Architecture)}: 32-core Intel\textregistered~Xeon\textregistered~Platinum 8352V CPU, 60 GB RAM, and a matching dual-disk storage topology.
    \end{itemize}

  \item \textbf{Platform II (GPU-Accelerated Reproduction \& Fine-Tuning)}: To faithfully replicate and fine-tune computationally intensive baselines (e.g., \textit{ORLM} and \textit{StepORLM}), a dedicated hardware stack was introduced to support parameterized model weights:
    \begin{itemize}
      \item \textit{GPU Cluster:} $1 \times$ NVIDIA\textregistered~vGPU-48GB-350W hosting 48 GB of dedicated VRAM.
      \item \textit{Host CPU:} 12 vCPUs allocated from an Intel\textregistered~Xeon\textregistered~Platinum 8260 CPU clocked at 2.40 GHz.
      \item \textit{System Memory \& Storage:} 62 GB RAM, 30 GB System Disk, and 50 GB high-speed local Data Disk.
    \end{itemize}
\end{itemize}

\subsubsection{Software Environment and Managed Dependencies}
The software architecture was systematically isolated into standard virtual runtime containers tailored to their respective platform duties.

\begin{itemize}
  \item \textbf{Runtime Container for Platform I (API Sandbox)}: Managed within a native Python 3.13 pipeline. The primary engine utilized for mathematical programming execution was the \textbf{Gurobi Optimizer} (v11.0 or higher). Auxiliary ML metrics and dataset formatting were handled via \texttt{scikit-learn}. API communications were funneled through secured official \texttt{openai} and \texttt{anthropic} SDK endpoints. Isolated sandboxes for executing generated code blocks were instantiated dynamically via Python's standard \texttt{subprocess}, \texttt{re}, and \texttt{ast} modules.

  \item \textbf{Runtime Container for Platform II (Deep Learning Virtualization)}: Implemented within a Linux container running \textbf{Ubuntu 22.04 LTS} managed via \textbf{Miniconda3}.
    \begin{itemize}
      \item \textit{Deep Learning Core:} Python 3.10 coupled with \textbf{CUDA Compute Unified Device Architecture v11.8} and \textbf{PyTorch} to exploit the underlying hardware acceleration. Local inference and fine-tuning pipelines utilized \texttt{transformers} for floating-point 16-bit (\texttt{FP16}) model loading and greedy decoding configurations.
      \item \textit{Weight and Hub Management:} Model checkpoints were programmatically retrieved and managed via the \texttt{huggingface\_hub} SDK using automated snapshot isolation tracking (\texttt{snapshot\_download}) to enforce offline reproducibility.
      \item \textit{Service Configuration:} Custom HTTP network interfaces were bridged via ports 6006 and 6008 to facilitate real-time monitoring of model convergence, loss distribution, and weight scaling.
    \end{itemize}
\end{itemize}

\subsubsection{API-Based Baseline Adaptations and Task Customization}
To establish an unbiased and rigorous benchmarking ecosystem, we cloned three advanced iterative prompt optimization and agentic methods from their official open-source GitHub repositories: \textit{OR-LLM-Agent} \cite{Or-llm-agent}, \textit{EvoPrompt} \cite{guo2023evoprompt}, and \textit{EvoAgent} \cite{yuan2025evoagent}. However, because the original implementation frameworks of \textit{EvoPrompt} and \textit{EvoAgent} were inherently agnostic to the structural specifications of optimization modeling and agent workflow synthesis, they were technically incompatible with tasks involving optimization agent generation. To address this, we executed task-specific adaptations and rigorous reimplementations of their prompt structures and evaluation interfaces, tailoring them to our optimization context while strictly preserving the baseline authors' core evolutionary framework and meta-evolutionary loop logic.
\begin{itemize}
  \item \textbf{Token-Constrained Training Budget:} To alleviate potential token explosion during programmatic evolutionary loops, we enforced a strict token computational budget, capping the total optimization loop feedback data at exactly 400,000 tokens per model generation.
  \item \textbf{Expanded Evaluation Endpoint:} For black-box model benchmarks, we expanded the base model support beyond standard generalist LLMs, scaling the remote API evaluation uniformly to encompass four diverse foundation architectures: DeepSeek-v3.2, GPT-5, Gemini 3 Flash, and Qwen 3 Max. All local and remote inference protocols were aligned under a uniform \texttt{pass@1} evaluation metric to ensure a fair comparison.
\end{itemize}

\subsubsection{Fine-Tuning and Inference for Specialized OR-LLMs}
Local parametric replication and standardized inference were executed for specialized OR foundation models, specifically \textit{ORLM} \cite{Or-llm-agent} and \textit{StepORLM} \cite{zhou2025steporlm}.
\begin{itemize}
  \item \textbf{Weight Loading \& Precision:} Model weights (\texttt{ORLM-LLaMA-3-8B} and \texttt{StepORLM-Qwen3-8B}) were programmatically fetched via \texttt{snapshot\_download} and loaded into local host memory in full 16-bit floating-point (\texttt{FP16}) precision without quantization, ensuring the integrity of their mathematical reasoning properties.
  \item \textbf{Decoding \& Execution Controls:} Localized model inferences were managed using the \texttt{transformers} library under a deterministic greedy decoding strategy (\texttt{decoding\_method='greedy'}) with a constraint of \texttt{max\_new\_tokens=2048}. Prompts and formatting structures matched their official code repositories, with minor calibrations to fit our expert-cleaned evaluation datasets. Generations were subsequently validated using a multi-threaded solver sandbox to extract rigid \texttt{pass@1} accuracy.
\end{itemize}

\subsubsection{Experimental Timeline and Inter-Model Integrity}
The complete grid evaluation, agent optimization tracking, and local fine-tuning steps were carried out from \textbf{March 2026 to April 2026}. Model API response distributions were monitored periodically across the endpoints throughout this time frame. No silent version updates, sudden performance decay, or prompt interface migrations were detected, guaranteeing the absolute determinism, stability, and reproducibility of all reported SOTA benchmarks.

\section{Knowledge Base Construction and Anti-Contamination Pipeline}
\label{app:knowledge_base}

EvoOR-Agent is supported by a dynamically constructed KB that stores reusable OR reasoning patterns, formulation heuristics, solver interaction strategies, and prompt templates distilled from recent literature and verified open-source implementations. The role of this KB is not to provide benchmark-specific solutions. Instead, it supplies abstract methodological priors that improve initialization, semantic mutation, and recovery during agent evolution.

At the same time, KB construction introduces a serious threat to experimental validity, namely \textbf{data leakage}. Recent studies at the intersection of OR and LLMs often report results on public benchmarks such as NL4Opt, Mamo, IndustryOR, and BWOR. Their papers and repositories may contain problem statements, concrete parameter settings, partial formulations, or near-complete solutions. If such instance-level artifacts are retrieved during evaluation, any downstream gain can no longer be attributed to genuine generalization. For this reason, KB construction must be treated as a controlled curation process rather than a simple retrieval step.

The pipeline is therefore designed around two objectives, namely \emph{methodological usefulness} and \emph{contamination safety}. It first retrieves candidate sources, then extracts high-value algorithmic components, and finally sanitizes the retained artifacts before inserting them into the KB.

\subsection{Source Retrieval and Scope Control}
The pipeline begins with automated literature retrieval over Google Scholar and arXiv-accessible metadata sources. Retrieval is restricted by Boolean queries that enforce the intersection of the two target themes, namely large language models and operations research. The primary query string is shown below.

\begin{quote}
  \texttt{("Large Language Model" OR "LLM" OR "Agent") AND ("Operations Research" OR "MILP" OR "Combinatorial Optimization" OR "Mathematical Modeling")}
\end{quote}

Only papers published between January 2023 and March 2026 are retained at this stage. This produces an initial corpus of candidate manuscripts together with their bibliographic metadata. The time restriction keeps the KB aligned with the recent OR and LLM literature while excluding earlier work that predates current agentic and reasoning-oriented paradigms.

\subsection{Three-Tier Extraction and Sanitization Pipeline}
The retrieved corpus still contains substantial noise. It includes papers that use OR to optimize LLM infrastructure, repositories that expose benchmark-specific demonstrations, and implementations whose reusable contribution is obscured by engineering detail. We therefore use a cascading LLM-assisted curation procedure that progressively narrows the corpus into a compact, high-value, and contamination-safe collection of reusable methodological components.

\subsubsection{Tier 1 Semantic Relevance and Thematic Alignment}
Tier 1 serves as the primary relevance filter. It evaluates the title, abstract, and introduction of each candidate paper to determine whether the work genuinely contributes LLM-driven mechanisms for OR modeling, solving, or agent orchestration, rather than merely sharing adjacent terminology.

\promptboxlabelnum{9}{box:tier1_filter}
\begin{promptbox}[Prompt Box~\thepromptbox: Tier 1 -- Thematic Alignment Classifier]
  You are an Academic Relevance Reviewer specializing in OR and artificial intelligence.

  [Task Description]
  Evaluate the provided paper abstract and introduction and determine whether the paper satisfies the strict inclusion criteria below.

  [Inclusion Criteria]
  The paper must satisfy both of the following conditions:
  \begin{enumerate}
    \item LLM application: The core methodology uses large language models (LLMs) or multi-agent reasoning frameworks.
    \item OR problem solving: The target domain is the mathematical modeling or solution of Operations Research problems (e.g., MILP, LP, combinatorial optimization, vehicle routing).
  \end{enumerate}

  [Exclusion Criteria]
  Reject the paper if any of the following is true:
  \begin{itemize}
    \item It uses OR methods to optimize LLM infrastructure (e.g., GPU memory routing optimized by MILP).
    \item It is a purely theoretical mathematics paper without substantive LLM usage.
    \item It focuses on general code generation without explicit formulation or solver integration (e.g., Gurobi or Pyomo).
  \end{itemize}

  [Output Format]
  Output a strict JSON object in the following form:
  \begin{quote}
    {\ttfamily\scriptsize
      \{"relevance": "YES or NO",\\
        "reasoning": "Two-sentence justification.",\\
    "contains\_github": "YES or NO (extract URL if YES)"\}}
  \end{quote}

  [Input Paper Text]
  \begin{quote}
    \verb|{paper_abstract_and_intro}|
  \end{quote}
\end{promptbox}

\subsubsection{Tier 2 Algorithmic Component and Source Code Extraction}
Papers that pass Tier 1 are then analyzed for the methodological elements most likely to transfer across OR tasks. These elements include formulation-oriented prompts, solver-feedback interpreters, debugging loops, decomposition templates, and relaxation or symmetry-handling strategies. When a GitHub repository URL is available, we use an automated crawler through the GitHub REST API so that extraction is grounded in executable artifacts rather than paper-level description alone.

The crawler first retrieves the repository \texttt{README.md} and directory tree. The LLM then uses the README to infer repository organization, locate the \texttt{.py} files containing the principal OR-agent logic, and extract the most informative code fragments or prompt templates. To preserve KB precision and avoid unnecessary context inflation, extraction is limited to at most three high-value components per paper.

\promptboxlabelnum{10}{box:tier2_extract}
\begin{promptbox}[Prompt Box~\thepromptbox: Tier 2 -- Component and Code Extraction]
  You are a Principal OR-Agent Architect. You are provided with the methodology section of an accepted paper together with source code fetched from its official GitHub repository after README-guided navigation.

  [Task Description]
  Extract the components proposed by this paper that most substantially improve LLM performance on OR tasks.

  [Examples of Valuable Components]
  \begin{itemize}
    \item Advanced prompt templates for mathematical modeling.
    \item Programmatic debugging loops or execution-feedback parsers.
    \item Algorithmic wrappers such as decomposition-based or relaxation-based structures.
  \end{itemize}

  [Extraction Constraints]
  \begin{enumerate}
    \item Quantity limit: Extract at most three critical components. Do not return trivial helper utilities.
    \item Summarization: For each component, provide a high-level theoretical summary of no more than 100 words.
    \item Code/prompt mapping: Extract the exact Python code snippet or prompt template corresponding to the component, and ensure that the extracted content is self-contained.
  \end{enumerate}

  [Output Format]
  Output a JSON list in the following schema:
  \begin{quote}
    {\ttfamily\scriptsize
      [\{"component\_name": "...",\\
          "theoretical\_summary": "...",\\
    "source\_code\_or\_prompt": "..."\}]}
  \end{quote}

  [Input Paper Methodology \& Fetched GitHub Code]
  \begin{quote}
    \verb|{paper_methodology_and_code}|
  \end{quote}
\end{promptbox}

\subsubsection{Tier 3 Data Sanitization and Leakage Prevention}
The components extracted in Tier 2 are potentially valuable, but they also represent the point at which contamination risk becomes most acute. Papers and repositories often embed concrete benchmark instances, real-world parameter values, or near-complete formulations directly in prompts, examples, and code comments.

If such artifacts enter the EvoOR-Agent context during evaluation, they create a direct leakage channel from external benchmarks into the reasoning process. Tier 3 therefore applies a strict sanitization protocol that removes all instance-specific content while preserving only the abstract procedural logic needed for methodological reuse.

\promptboxlabelnum{11}{box:tier3_sanitize}
\begin{promptbox}[Prompt Box~\thepromptbox: Tier 3 -- Anti-Contamination and Data Sanitization]
  You are a Strict Data Sanitization Inspector. Your highest priority is to prevent test-set contamination (data leakage) in an LLM evaluation pipeline.

  [Contamination Definition]
  Treat the component as contaminated if it contains any of the following:
  \begin{enumerate}
    \item Specific natural-language OR problem descriptions (e.g., ``A factory produces 300 units ...'', ``A supply chain with five nodes ...'').
    \item Hard-coded numerical parameters associated with benchmark datasets such as NL4Opt, Mamo, IndustryOR, or BWOR.
    \item Semantic variable names copied directly from problem statements (e.g., truck\_capacity, factory\_A\_production).
  \end{enumerate}

  [Sanitization Protocol]
  Systematically scrub all contamination while preserving the abstract algorithmic logic.
  \begin{enumerate}
    \item Parameter abstraction: Replace all specific numbers with abstract placeholders (e.g., 300 $\rightarrow$ [PARAM\_1]).
    \item Semantic stripping: Replace entity-specific names in code and prompts with generic mathematical notation (e.g., truck\_capacity $\rightarrow$ capacity\_i, production $\rightarrow$ variable\_x).
    \item Few-shot deletion: If a prompt contains a complete example for a specific benchmark-style problem, delete the example block and replace it with [Provide abstract mathematical formulation example here].
  \end{enumerate}

  [Zero-Tolerance Check]
  After sanitization, perform a final scan. If the component still contains any recognizable trace of a specific real-world problem scenario, discard it entirely and output "STATUS": "COMPROMISED".

  [Output Format]
  Output the sanitized JSON component list. If a component cannot be safely sanitized, remove it from the list.

  [Input Extracted Components]
  \begin{quote}
    \verb|{tier_2_extracted_components}|
  \end{quote}
\end{promptbox}

\subsection{Case Study: Processing the OR-LLM-Agent Framework}
To illustrate the behavior of the proposed pipeline on a realistic external source, we analyze how it processes the recent \textbf{OR-LLM-Agent} \cite{Or-llm-agent} framework. OR-LLM-Agent is a reasoning-oriented multi-agent method for OR problem solving. It coordinates a Math Agent, a Code Agent, and a Debugging Agent, and it is evaluated extensively on the BWOR benchmark. This framework is representative because it contains genuinely reusable methodological structure while also presenting a clear contamination risk.

\begin{itemize}
  \item \textbf{Tier 1 Processing:} The system identified the paper's title and abstract as a perfect thematic match, strictly aligning with both the LLM multi-agent framework criteria and the OR mathematical modeling requirements.
  \item \textbf{Tier 2 Extraction:} Triggered by the \texttt{contains\_github} flag, the crawler navigated to the official repository (\texttt{or\_llm\_agent}). Guided by the README, the LLM successfully extracted the core \texttt{or\_llm\_agent} Python function. It isolated three high-value structural components: the explicit instruction separating mathematical formulation from Gurobi code generation, the structural backtracking prompt for infeasible solutions (e.g., \textit{``The current model resulted in *no feasible solution*...''}), and the multi-attempt error tracing loop.
  \item \textbf{Tier 3 Sanitization:} The OR-LLM-Agent paper prominently features a specific benchmark example regarding a ``Candy factory'' producing grades J, K, and L with specific raw material percentages to maximize a profit of 5450. The Tier 3 inspector detected these concrete entities and numerical values within the extracted context payload. The sanitization protocol successfully scrubbed all tabular data and specific entity names (e.g., replacing ``candy factory'' with generic \texttt{[facility]} placeholders), ensuring that the abstract multi-agent debugging logic was integrated into the Knowledge Base without contaminating our evaluation against the BWOR dataset.
\end{itemize}

To make the post-sanitization representation explicit, we convert the extracted OR-LLM-Agent prompt and execution logic into a plain-text case-analysis artifact that follows the same schema as the JSON KB. Rather than preserving the original prompt box or raw Python listing verbatim, the artifact below records only the reusable methodological core in a compact, contamination-safe form suitable for KB insertion.

\begin{lstlisting}[language={}, caption={Case-analysis TXT document for a sanitized OR-LLM-Agent knowledge-base entry.}, label={lst:kb_case_or_llm_agent}]
  [KB CASE ANALYSIS TXT]
  component_name: OR-LLM-Agent three-stage reasoning and self-repair workflow

  theoretical_summary:
  This component encodes a staged OR-agent architecture in which problem solving
  is decomposed into mathematical modeling, solver-code generation, and iterative
  self-repair. Its main contribution is not a benchmark-specific formulation, but
  an executable control pattern that separates symbolic modeling from code
  synthesis and escalates to deeper reflection when code-level repair is
  insufficient. The workflow is reusable as a generic orchestration template for
  LLM-based optimization agents.

  abstract_template:
  Stage 1 - Mathematical modeling:
  - Parse a natural-language OR problem.
  - Identify sets, parameters, decision variables, objective, and constraints.
  - Produce a formal mathematical model before any code is generated.
  - Prioritize correctness, dimensional consistency, and logical completeness.

  Stage 2 - Code generation:
  - Translate the mathematical model into executable Python solver code.
  - Include imports, model initialization, variable definitions, objective,
  constraints, optimization call, and result extraction.
  - Require the generated solver program to be returned as a complete code block.

  Stage 3 - Automated debugging and self-repair:
  - If execution fails, inspect runtime errors and regenerate corrected code.
  - If execution succeeds but yields no valid numeric solution, trigger a deeper
  reflection step over the mathematical model itself.
  - Rebuild the formulation when infeasibility, contradiction, or repeated repair
  failure suggests a modeling-level defect.

  orchestration_logic:
  1. Initialize the message history with a modeling-oriented system prompt.
  2. Query the LLM to obtain a mathematical model.
  3. Append the model to the conversation state.
  4. Query the LLM again to obtain solver code derived from that model.
  5. Execute the generated code through the external solve-and-repair routine.
  6. If the returned result is non-numeric, request formulation-aware revision.
  7. If repeated code debugging fails, escalate to model-level reconstruction.
  8. Return final execution success status and the parsed optimization result.

  sanitization_notes:
  - Removed all benchmark-specific examples, entities, and numeric instances.
  - Replaced implementation details with reusable procedural descriptions.
  - Preserved only the abstract multi-stage control logic needed for KB reuse.
  - Stored the artifact as a plain-text case-analysis document aligned with the
  JSON KB fields: component_name, theoretical_summary, and abstract_template.
\end{lstlisting}

\subsection{Knowledge Base Representation and Evolutionary Use}
After passing Tier 3, the retained components are serialized into a structured JSON knowledge base. Each entry stores a generalized \texttt{component\_name}, a concise \texttt{theoretical\_summary}, and a sanitized \texttt{abstract\_template} that is intentionally stripped of benchmark-specific entities, values, and formulations. This representation is designed to preserve transferable reasoning structure while preventing the KB from degenerating into a cache of hidden exemplars.

During evolution, whenever the EvoOR-Agent encounters difficult routing structures, fragile formulations, or solver-facing failure modes, it can query this database for abstract templates, as exemplified by Prompt Box~\ref{box:kb_mutation}. The retrieved knowledge serves as structural guidance, such as multi-agent decomposition patterns, reflection prompts, or calibrated Big-M heuristics, without exposing the model to the original benchmark instances from which those ideas were abstracted. Consequently, any performance gain attributable to KB access reflects improved reasoning organization and methodological reuse rather than memorization of evaluation data.

\end{document}